\documentclass[10pt,twocolumn,letterpaper]{article}

\usepackage{iccv}
\usepackage{xcolor}
\usepackage{dsfont}
\usepackage{pifont}
\usepackage{amsmath}
\usepackage{amssymb}
\usepackage{booktabs}
\usepackage[utf8]{inputenc}
\usepackage[english]{babel}
\usepackage{csquotes}
\usepackage{subcaption}
\usepackage{tikz}
\usepackage{graphicx}
\usepackage{multirow}
\usepackage{adjustbox}
\newcommand{\cmark}{\ding{51}}%
\newcommand{\xmark}{\ding{55}}%
\usetikzlibrary{backgrounds}

\newcommand{\myparagraph}[1]{\vspace{0.1cm}\noindent {\bf #1.}}
\usepackage{etoc}

\usepackage{amsmath,amsfonts,bm}

\def\Ncal{{\mathcal N}}
\def\Scal{{\mathcal S}}

\def\eqref#1{equation~\ref{#1}}

\def\1{\bm{1}}

\DeclareMathAlphabet{\mathsfit}{\encodingdefault}{\sfdefault}{m}{sl}
\SetMathAlphabet{\mathsfit}{bold}{\encodingdefault}{\sfdefault}{bx}{n}

\definecolor{iccvblue}{rgb}{0.21,0.49,0.74}
\usepackage[pagebackref,breaklinks,colorlinks,allcolors=iccvblue]{hyperref}

\title{LUDVIG: Learning-Free Uplifting of 2D \\ Visual Features to Gaussian Splatting Scenes}

\author{
    Juliette~Marrie$^{1,2}$ 
    \hspace{0.1cm} Romain~Menegaux$^1$
    \hspace{0.1cm} Michael~Arbel$^1$
    \hspace{0.1cm} Diane~Larlus$^2$
    \hspace{0.1cm} Julien~Mairal$^1$\\
    $^1$~{Univ.\ Grenoble Alpes, Inria, CNRS, Grenoble INP, LJK} \hspace{1cm} $^2$~{NAVER LABS Europe}
}

\begin{document}
\maketitle
\begin{abstract}
We address the problem of extending the capabilities of vision foundation models such as DINO, SAM, and CLIP, to 3D tasks. Specifically, we introduce a novel method to uplift 2D image features into Gaussian Splatting representations of 3D scenes. 
Unlike traditional approaches that rely on minimizing a reconstruction loss, our method employs a simpler and more efficient feature aggregation technique, augmented by a graph diffusion mechanism. 
Graph diffusion refines 3D features, such as coarse segmentation masks, by leveraging 3D geometry and pairwise similarities induced by DINOv2.
Our approach achieves performance comparable to the state of the art on multiple downstream tasks while delivering significant speed-ups.
Notably, we obtain competitive segmentation results using only generic DINOv2 features, despite DINOv2 not being trained on millions of annotated segmentation masks like SAM.  When applied to CLIP features, our method demonstrates strong performance in open-vocabulary object segmentation tasks, highlighting the versatility of our approach.\footnote{\scriptsize{Project page: \url{https://juliettemarrie.github.io/ludvig}}}
\vspace{-.2cm}
\end{abstract}

\section{Introduction}
\label{sec:intro}

\begin{figure}
\vspace{-0.3cm}
\includegraphics[width=\linewidth]{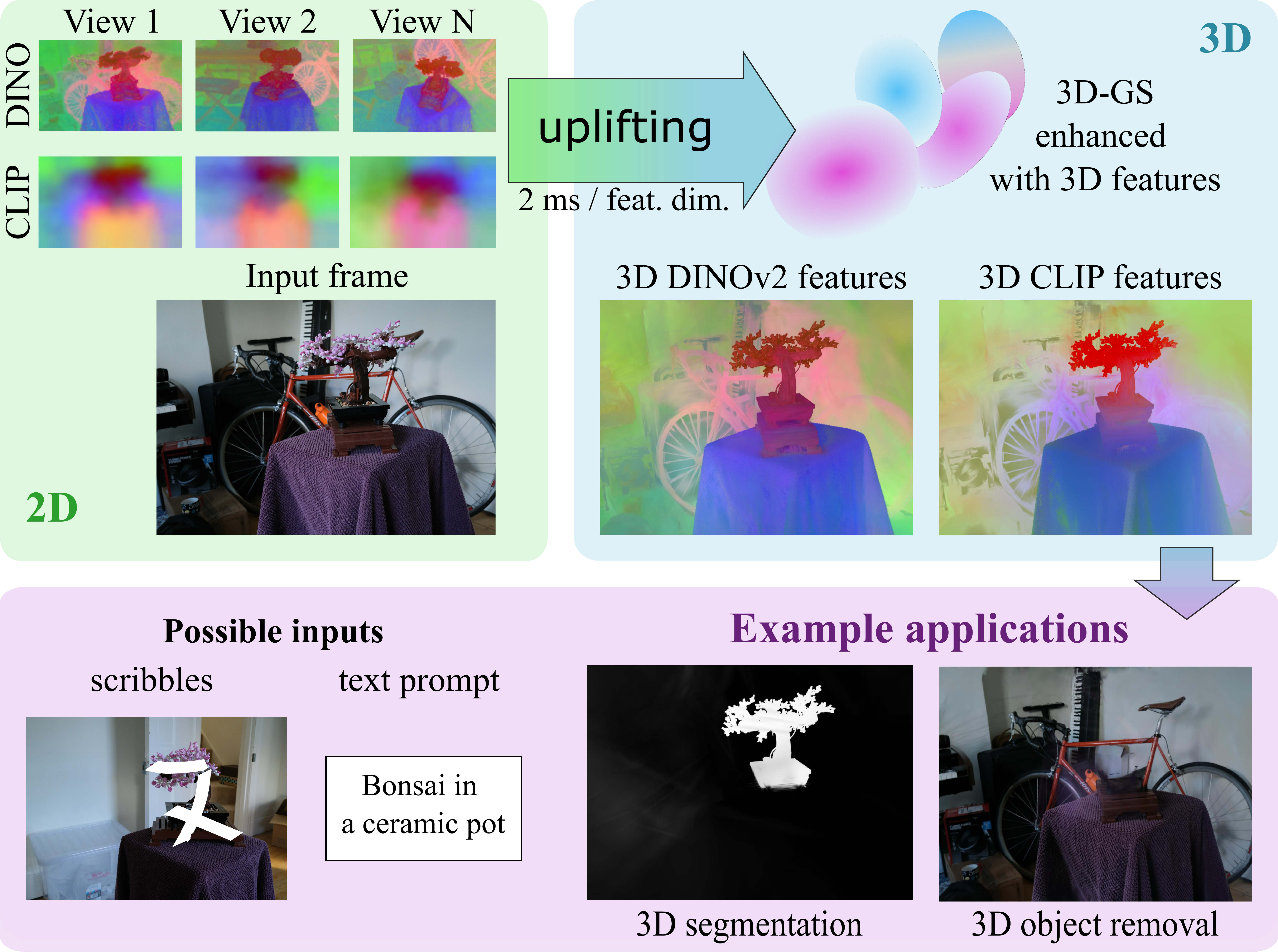}
\vspace{-0.3cm}
\caption{
In this paper, we propose a simple, parameter-free method to uplift any 2D visual features (\emph{e.g.}, CLIP or DINO) into a 3D Gaussian Splatting (3D-GS) 
representation. Uplifting is directly implemented into the rendering process, and takes about 2ms per image and feature dimension. The uplifted features can be leveraged for various 3D tasks, such as segmentation or object removal, based on various inputs, such as scribbles or text prompts.
}
\vspace{-0.2cm}
\label{fig:teaser}
\end{figure}

Image understanding has made remarkable progress, driven by large pretrained models such as CLIP~\citep{radford2021clip}, DINO~\citep{caron2021dino,oquab2024dinov2}, or SAM~\citep{kirillov2023sam}, also called foundation models. A key factor behind their exceptional generalization capabilities lies in the size of their training datasets, often composed of millions or even billions of samples.

Meanwhile, 3D scene representation has also advanced through machine learning approaches like NeRF~\citep{mildenhall2021nerf} and model fitting techniques such as Gaussian Splatting~\citep{kerbl2023gaussiansplatting}. 
These methods typically reconstruct scenes from a few dozen views captured from different angles, effectively preserving both appearance and geometric details. However, they are not suited for semantic tasks.

The complementarity of these two families of approaches has been exploited to integrate geometry and image-level semantics extracted by large 2D pretrained models into NeRF or Gaussian Splatting 3D representations. 
This has led to a surge in methods for language-guided object retrieval~\citep{kerr2023lerf, liu2023weakly, zuo2024fmgs, wu2024opengaussian}, scene editing~\citep{kobayashi2022decomposing, chen2024gaussianeditor, fan2023lightgaussian}, or semantic segmentation~\citep{cen2023segmentNeRF, ye2024gaussiangrouping, ying2023omniseg3d} in 3D scenes.

The main limitation of these approaches lies in their dependence on optimization, which requires an iterative process to learn a scene-specific 3D representation by minimizing a reprojection error across all training views. While this loss function is intuitive, a faster and more straightforward method for transferring 2D generic visual features to \emph{already trained} Gaussian Splatting 3D models would be preferable, which is one of the purposes of this work.

In this paper, we demonstrate that a simple, learning-free process is highly effective for uplifting 2D features or semantic masks into 3D Gaussian Splatting scenes. This process, which can be viewed as an `inverse rendering' operation, is both computationally efficient and adaptable to any feature type. We showcase its effectiveness by uplifting visual features from DINOv2~\citep{oquab2024dinov2, darcet2024registers}, semantic masks from SAM~\citep{kirillov2023sam} and SAM2~\citep{ravi2024sam2}, and visual features from CLIP~\citep{ilharco2021openclip}, enabling language-driven applications.
Then, we show that a graph diffusion mechanism~\citep{kondor2002diffusion,smola2003kernels} is helpful for feature refinement in 3D scenes. This mechanism is rooted in spectral graph theory and spectral clustering~\citep{belkin2001laplacian,shi2000normalized,meila2000learning}.
In the context of our work, it transforms coarse segmentation inputs, such as scribbles or alignment scores between visual features and a text query, into accurate 3D segmentation masks without the need for segmentation models such as SAM.
When evaluated on segmentation and open-vocabulary object localization tasks, our method matches state-of-the-art performance while being significantly faster than previous optimization-based approaches.

To summarize, our contributions are threefold: (i)~We introduce a simple, learning-free uplifting approach that can be directly integrated into the rendering process (Sec.~\ref{sec:uplifting}), achieving state-of-the-art segmentation results when applied to SAM-generated semantic masks (Sec.~\ref{sec:method_seg}); 
(ii) We introduce a graph diffusion process for 3D segmentation based on uplifted features (Sec.~\ref{sec:diffusion}) which yields competitive results in both foreground/background segmentation with DINOv2 alone and open-vocabulary segmentation when combined with CLIP (Secs.~\ref{sec:method_seg}, \ref{sec:method_clip});
(iii) We show that our simple uplifting approach can replace more complex and time-consuming feature learning methods and yield strong accuracy gains (Sec.~\ref{sec:method_semantic}).
\section{Related work}
\label{sec:rw}

\myparagraph{Learning 3D {semantic} scene representations with NeRF}
NeRF~\citep{mildenhall2021nerf} uses a multilayer perceptron to predict the volume density and radiance for any given 3D position and viewing direction.
Such a representation can naturally be extended to semantic features.
The early works N3F~\citep{tschernezki2022n3f} and DFF~\citep{kobayashi2022decomposing} distill DINO 2D (\emph{i.e.}, image-level) features~\citep{caron2021dino} in scene-specific NeRF representations. \citet{kobayashi2022decomposing} also distill LSeg~\citep{li2022lseg}, a  language-driven model for semantic segmentation. Building on this, LERF~\citep{kerr2023lerf} and 3D-OVS~\citep{liu2023weakly} learn 3D CLIP~\citep{radford2021clip} and DINO~\citep{caron2021dino} features jointly for open-vocabulary segmentation. These works were extended to other pretrained models such as latent diffusion models~\citep{ye2023featurenerf} or  SAM~\citep{kirillov2023sam} for segmentation~\citep{cen2023segmentNeRF, ying2023omniseg3d}.

\myparagraph{Learning 3D {semantic} scene representations with GS}
Subsequent work has relied on the Gaussian Splatting method~\citep{kerbl2023gaussiansplatting}, enabling rendering speeds orders of magnitude faster than NeRF-based models. 
Several tasks have been addressed such as semantic segmentation using SAM~\citep{cen2023segmentGS, ye2024gaussiangrouping, kim2024garfield, wu2024opengaussian}, language-driven retrieval or editing using CLIP combined with DINO~\citep{zuo2024fmgs} or SAM~\citep{ye2023featurenerf, wu2024opengaussian}, scene editing using diffusion models~\citep{chen2024gaussianeditor, wang2024gaussianeditor}, and 3D-aware finetuning~\citep{yue2024fit3d}.
These works learn 3D semantic representations by minimizing a reprojection loss.
As a single scene can be represented by over a million Gaussians, such optimization-based techniques have strong memory and computational limitations. 
To handle these, FMGS~\citep{zuo2024fmgs} employs a multi-resolution hash embedding (MHE) of the scene for uplifting DINO and CLIP representations, Feature 3DGS~\citep{zhou2024feature3dgs} learns a $1\times 1$ convolutional upsampler of Gaussians' features distilled from LSeg and SAM's encoder, and LangSplat~\citep{qin2023langsplat} learns an autoencoder to reduce CLIP feature dimension from 512 to 3. In contrast, our approach requires no learning, which significantly speeds up the uplifting process and reduces the memory requirements. 

\myparagraph{Direct uplifting of 2D features into 3D} Direct uplifting from 2D to 3D has been explored in prior works.
GaussianEditor~\citep{chen2024gaussianeditor} uplifts 2D SAM masks into 3D to selectively optimize Gaussians for editing tasks, see details in the supplementary (Sec.~\ref{sec:appendx-gaussianeditor}). Dr. Splat~\citep{jun2025drsplat} also lifts 2D SAM masks, to create an assignment from object CLIP features to Gaussians. Both approaches are specific to segmentation and are ill-suited for uplifting generic representations. Semantic Gaussians \citep{guo2024semantic} pairs 2D pixels with 3D Gaussians along each pixel’s ray based on depth information but relies on a learned 3D convolutional network. OpenGaussian \citep{wu2024opengaussian} also pairs pixels with  Gaussians, but relies on pseudo-features learned with a contrastive loss on SAM masks. In contrast, our approach is simple to implement, applicable to any 2D representation and entirely parameter-free.

\myparagraph{Leveraging 3D information to better segment in 2D}
Most prior works focusing on semantic segmentation leverage 2D models specialized for this task. The early work of \cite{yen2022nerfsupervision} uplifts learned 2D image inpainters by optimizing view consistency over depth and appearance. 
Subsequent works have mostly relied on uplifting either features from SAM's encoder~\citep{zhou2024feature3dgs}, binary SAM masks~\citep{cen2023segmentNeRF, cen2023segmentGS}, or SAM masks automatically generated for all objects in the image~\citep{ye2024gaussiangrouping, ying2023omniseg3d, kim2024garfield}.
The latter approach is computationally expensive, as it requires querying SAM on a grid of points over the image. It also requires matching inconsistent mask predictions across views, with \eg a temporal propagation model~\citep{ye2024gaussiangrouping} or a hierarchical learning approach~\citep{kim2024garfield}, which introduces additional computational overhead. In this work, we focus on single instance segmentation and show that our uplifted features are on par with the state of the art~\citep{cen2023segmentNeRF,cen2023segmentGS,ying2023omniseg3d}.
Standing out from prior work uplifting DINO features~\citep{tschernezki2022n3f,kobayashi2022decomposing,isrfgoel2023interactive,kerr2023lerf,liu2023weakly, ye2023featurenerf, zuo2024fmgs}, we show that DINOv2 features can be used on their own for semantic segmentation and rival SAM-based models through a simple graph diffusion process that leverages 3D geometry.

\myparagraph{3D CLIP features for open-vocabulary localization} 
For learning 3D CLIP features, prior works also leverage vision models such as DINO or SAM. DINO is used to regularize and refine CLIP features~\citep{kerr2023lerf, liu2023weakly, zuo2024fmgs, shi2024legaussians}, while SAM is employed for generating instance-level CLIP representations~\citep{qin2023langsplat, wu2024opengaussian, jun2025drsplat}. These approaches suffer from high computational costs, resorting to dimensionality reduction or efficient multi-resolution embedding representations, and usually run for a total of one to two hours for feature map generation and 3D feature optimization. 
In contrast, our approach bypasses the high computational cost of gradient-based optimization and, combined with graph diffusion, is an order of magnitude faster than these prior works.
\section{Uplifting 2D visual representations into 3D}
\label{sec:method_uplifting}

We now present a simple yet effective method for lifting 2D visual features into 3D using Gaussian Splatting, and discuss its relation with optimization-based techniques.

\subsection{Background on Gaussian Splatting}
\label{sec:background}

\myparagraph{Scene representation}
The Gaussian Splatting method consists in modeling a 3D scene as a set of~$n$ Gaussians densities $\Ncal_i$, each defined by a mean $\mu_i$ in $\mathbb{R}^3$, a covariance $\Sigma_i$ in $\mathbb{R}^{3\times 3}$, an opacity $\sigma_i$ in $(0,1)$, and a color function $c_i({d})$ that depends on the viewing direction $d$.
$d$ refers to the full camera pose, \ie its extrinsics and intrinsics.

A 2D frame at a given view is an image $\hat{I}_d$ rendered by projecting the 3D Gaussians onto a 2D plane, parametrized by the viewing direction $d$. This projection accounts for the opacity of the Gaussians and the order in which rays associated with each pixel pass through the densities.
More precisely, a pixel $p$ for a view $d$ is associated to an ordered set $\Scal_{{d},p}$ of Gaussians and its value is obtained by summing their contributions:
\vspace{-.07cm}
\begin{equation}\label{eq:render}
\hat{I}_{{d}}(p) = \sum_{i \in \Scal_{d,p}} c_i(d) w_{i}(d, p). 
\end{equation}
\vspace{-.07cm}
The above weights are obtained by $\alpha$-blending, \ie $w_{i}(d, p) = \alpha_{i}(d,p) \prod_{j\in \mathcal{S}_{d,p}, j<i} \left(1-\alpha_{j}(d,p)\right)$, 
where the Gaussian contributions $\alpha_{i}(d,p)$ are given by the product of the opacity $\sigma_i$ and the Gaussian density $\mathcal{N}_i$ projected onto the 2D plane at pixel position $p$.

\myparagraph{Scene optimization}
Let $I_1,\dots,I_m$ be a set of 2D frames from a 3D scene and $d_1,\dots,d_m$ the corresponding viewing directions. Gaussian Splatting optimizes the parameters involved in the scene rendering
function described in the previous section. This includes the means and covariances of the Gaussian densities, their opacities, and the color function parametrized by spherical harmonics. Denoting these parameters by $\theta$, the following reconstruction loss is used
\vspace{-.07cm}
 \begin{equation}
 \label{eq:gs-problem}
 \min_{\theta} \frac 1m \sum_{k=1}^m \mathcal{L}(I_k, \hat{I}_{d_k,\theta}),
 \end{equation}
\vspace{-.07cm}
where $\hat{I}_{d_k,\theta}$ is the frame of the scene rendered in the direction $d_k$ as in Eq.~(\ref{eq:render}), using the parameters $\theta$, and $\mathcal L$ is a combination of $\ell_1$ and SSIM loss functions~\citep{kerbl2023gaussiansplatting}.

\begin{figure*}[h!]
    \centering
    \includegraphics[width=.84\linewidth]{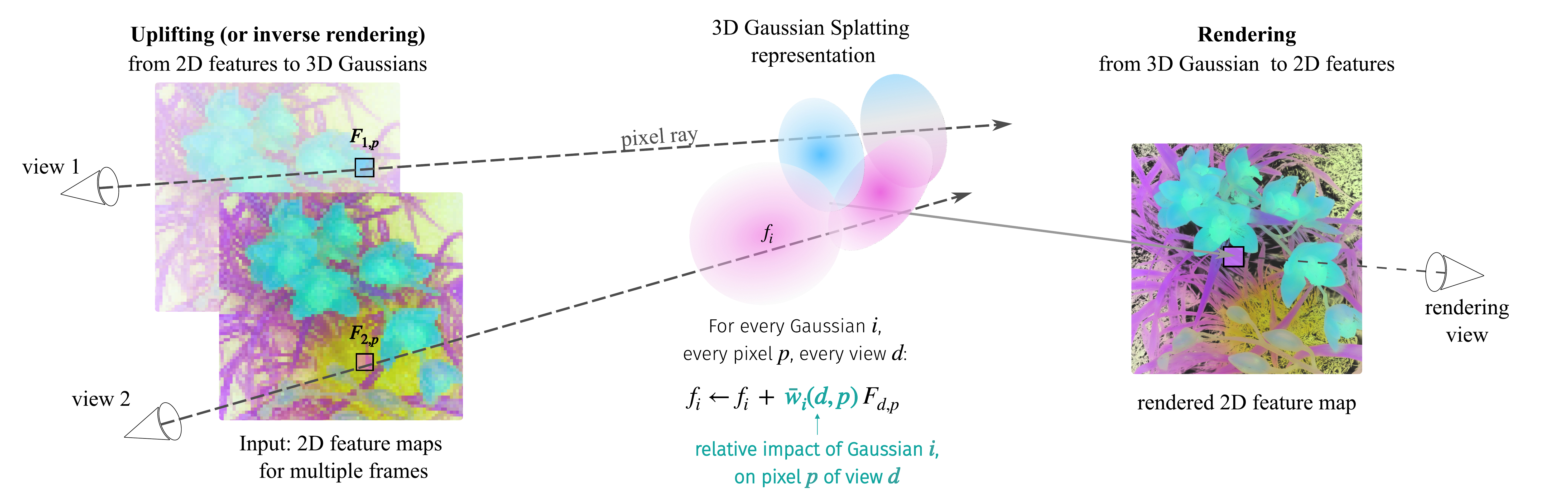}
    \vspace{-0.2cm}
    \caption{\textbf{Illustration of uplifting and rendering.} In the uplifting phase, features ${\bf{f}}$ are created for each 3D Gaussian by aggregating coarse 2D features ${\bf F}$ over all viewing directions. For rendering, the 3D features ${\bf{f}}$ are projected on any given viewing direction as in regular Gaussian Splatting. The rendering weight $\bar{w}_{i}(d, p)$ represents the relative influence of the Gaussian $i$ on pixel $p$ (Eq.~(\ref{eq:uplifting})).} 
    \label{fig:uplift_render}
    \vspace{-.1cm}
    \vspace{-0.3cm}
\end{figure*}

\subsection{Uplifting 2D feature maps into 3D}
\label{sec:uplifting}

Given a 3D Gaussian Splatting scene representation, we propose a method to uplift any set of 2D features maps $\bf{F} = \{F_1,...,F_m\}$ associated to $m$ views into 3D. These features may be obtained from pretrained models~\cite{radford2021clip,oquab2024dinov2} or segmentation masks. The uplifted features ${\bf f}$ should allow for downstream 3D tasks, such as 3D segmentation, while being `faithful' when reprojected back in 2D.

\myparagraph{Uplifting with simple aggregation}
We construct uplifted features for each 3D Gaussian of the Gaussian Splatting scene as a weighted average of 2D features from all frames. 
Each 2D feature $F_{d,p}$ from a frame at a given viewing direction $d$ and pixel $p$ contributes to the feature $f_i$ by a factor proportional to the rendering weight $w_{i}(d, p)$, if the Gaussian $i$ belongs to the ordered set $\mathcal{S}_{d,p}$ associated to the view/pixel pair $(d,p)$. Denoting $\mathcal{S}_i = \{(d,p), i\in \mathcal{S}_{d,p}\}$ as the set of view/pixel pairs contributing to the feature $f_i$, the resulting features are defined as follows: 
\begin{equation}
    \label{eq:uplifting}
    f_i =  \!\! \sum_{(d,p) \in \mathcal{S}_i}  \! \bar{w}_{i}(d, p) F_{d,p},  ~~\text{}~~  \bar{w}_i(d,p) = \frac{w_{i}(d, p)}{\sum_{(d,p) \in \mathcal{S}_i} \! w_{i}(d, p)}.
\end{equation}
We can interpret this equation as a normalized version of the transposed rendering operation over the $m$ viewing directions. 
More precisely, the rendering of any view-independent collection of features ${\bf{f}} = (f_i)$ attached to the~$n$ Gaussians into the $m$ training frames can be represented as a linear operator $W$ acting on the collection $\bf{f}$ and returning a collection of 2D feature maps ${\bf{\hat{F}}} = (\hat{F}_{d,p})$, see Eq.~(\ref{eq:rendering}) below.
Here, non-zero entries of the matrix $W$ consists of all rendering weights $w_{i}(d, p)$ when $(d,p) \in \mathcal{S}_i$ is placed at row $(d,p)$ and column $i$, and $\mathbf{\hat{F}}$ is a 2D matrix containing all (flattened) 2D feature maps generated for all cameras poses, with $\mathbf{\hat{F}}_{d,p}$ the feature of pixel $p$ from view at direction $d$.
Similarly, the uplifting expression introduced in Eq.~(\ref{eq:uplifting}) can be expressed in terms of the transpose of $W$ and a diagonal matrix $D$ of size $m$ representing the normalization factor and whose diagonal elements are obtained by summing over the rows of $W$ as in Eq.~(\ref{eq:uplifting_matrix}) below:

\vspace{-.2cm}
\begin{minipage}{0.2\textwidth}
\[
\text{Rendering to $m$ frames}
\]
\begin{equation}
    {\bf \hat{F}} = W{\bf{f}},
\label{eq:rendering}
\end{equation}
\end{minipage}
\hfill
\begin{minipage}{0.2\textwidth}
\[
\text{Uplifting from $m$ frames}
\]
\begin{equation}
    {\bf f} = D^{-1}W^{\top}{\bf{F}}.
\label{eq:uplifting_matrix}
\end{equation}
\end{minipage}
\vspace{.3cm}

Note that $W$ and $D$ are never explicitly constructed. Instead, they are computed by calling the forward rendering function for Gaussian Splatting and replacing the color vectors by the feature vectors (see Fig.~\ref{fig:uplift_render}). All these operations are performed within the CUDA rendering process.

\myparagraph{Connection with optimization-based inverse rendering}  
An alternative approach to uplifting 2D features ${\bf F}$ is to minimize a reconstruction objective $\mathcal{L}({\bf f})$, where the goal is to find uplifted features~${\bf f}$ whose rendering closely matches the original 2D features ${\bf F}$ \citep{tschernezki2022n3f, kerr2023lerf, zuo2024fmgs}. A natural choice is to minimize the mean squared error between the 2D features~${\bf F}$ and the rendered ones ${\bf \hat{F}}$ as defined by Eq.~(\ref{eq:rendering}):
\begin{align}
\min_{{\bf f}} \mathcal{L}({\bf f}) := \frac{1}{2}\Vert {\bf F} - W{\bf f} \Vert^2.
\label{eq:reconstruction_loss}
\end{align}
Such an approach requires an optimization procedure which is costly compared to our proposed uplifting method. Nevertheless, it is possible to interpret the proposed uplifting scheme in Eq.~(\ref{eq:uplifting_matrix}) as a single pre-conditioned gradient descent step on the reconstruction objective, starting from a~${\bf 0}$~feature, \emph{i.e.}, ${\bf f}= -D^{-1}\nabla \mathcal{L}({\bf 0})$. In practice, we found that performing more iterations on the objective $\mathcal{L}({\bf f})$ did not improve the quality of the features, thus suggesting that the computationally cheaper scheme in Eq.~(\ref{eq:uplifting_matrix}) is already an effective approach to uplifting.

\myparagraph{Gaussian filtering}
The normalization term $\beta_i = \sum_{d,p\in \mathcal{S}_i} w_{i}(d, p)$ serves as an estimator of the relative importance of each Gaussian in the scene. Therefore, it can be used as a criterion to prune the set of Gaussians for memory efficiency. In our experiments, we filter out half of the Gaussians based on $\beta_i$ and observe no qualitative nor quantitative degradation of the results. This approach is inspired by prior work on efficient Gaussian Splatting representation such as proposed by~\cite{fan2023lightgaussian} that also prunes Gaussians based on their contribution to each pixel in the training frames.  

\subsection{Enriching features by graph diffusion}
\label{sec:diffusion}
DINOv2 features have shown remarkable performance on semantic segmentation with simple linear probing~\citep{oquab2024dinov2}, making them 
a good candidate to enrich features that lack such a property, such as those from CLIP~\citep{wysoczanska2023clipdino,zuo2024fmgs,liu2023weakly}. 
Inspired by spectral clustering techniques~\citep{shi2000normalized,kondor2002diffusion,belkin2001laplacian} and manifold denoising~\citep{hein2006manifold}, 
we propose to \textit{diffuse} features that have been uplifted to 3D. This process aims to align semantic features with the scene layout and object boundaries implied by DINOv2.
In contrast to prior work, we perform this landscaping with DINOv2 directly in the 3D scene, thereby taking 3D geometry into account as well.

\myparagraph{Graph construction}
We construct a graph whose nodes are the $n$ 3D Gaussians and edges, represented by a matrix $A$ of size $n\times n$, are based on 3D Euclidean geometry between the nodes and the similarity between their DINOv2 features. More precisely, we extract the $k$ nearest neighbors $\mathcal{N}(i)$ for each node $i$, as measured by the Euclidean distance between the centers of the 3D Gaussians. 
Two nodes $i$ and $j$ in the graph are linked by an edge if $i \in \mathcal{N}(j)$ or $j \in \mathcal{N}(i)$, and the edge is assigned the following weight:
\vspace{-.07cm}
\begin{equation}
\label{eq:edges}
A_{ij} = S_f(f_i, f_j) \, P(f_i)^{\frac 12}P(f_j)^{\frac 12}, 
\vspace{-.07cm}
\end{equation}
with $S_f(f_i,f_j)$ a local similarity between features $f_i$ and $f_j$, typically defined as a RBF kernel. For tasks requiring diffusion to be confined to a specific object instance, we prevent leakage into the background by introducing a node-wise unary regularization term $P(f_i)$ which quantifies the similarity between the node feature $f_i$ and the features of the object of interest. Details on $S_f$ and $P$ are provided in the supplementary (Sec.~\ref{sec:appendix-diffusion}).

\myparagraph{Diffusion on the graph}  
Given initial 3D features $g_0$ in $\mathbb{R}^n$, which we aim to improve by using information encoded in $A$ (3D geometry and DINOv2 similarities), we perform $T$ diffusion steps to construct a sequence of diffused features $(g_t)_{1\leq t \leq T}$  defined as follows:
\vspace{-.07cm}
\begin{equation}
g_{t+1} = A\tilde{g}_{t}, \quad \tilde{g}_{t} = {g_t}/{\|g_t\|_2},
\label{eq:diffusion}
\vspace{-.07cm}
\end{equation}
This can be seen as performing a few steps of the power method, projecting $g_0$ into the dominant eigenspace of~$A$. Depending on the task, $g_0$ may represent generic features or task-specific features such as 3D segmentation masks.
\vspace{-.05cm}
\section{From 3D uplifting to downstream tasks}

In this section, we describe our pipeline around the uplifting process from Sec.~\ref{sec:method_uplifting}, covering preceding and following stages for three tasks: multi-view segmentation, open-vocabulary object segmentation, and open-vocabulary semantic segmentation. 
As in Sec.~\ref{sec:method_uplifting}, we are given a set of 2D frames $I_1, \dots, I_m$, with viewing directions $d_1, \dots, d_m$, and a 3D Gaussian Splatting representation of the scene.

\subsection{Multi-view segmentation}\label{sec:muti_seg}
\label{sec:method_seg}
We assume that a foreground mask of the object to be segmented is provided on the \emph{reference frame}~$I_1$.
This mask can either be a full segmentation or a sparse set of \emph{scribbles}, both defining a set of foreground pixels $\mathcal{P}$.
The task is to generate a 2D segmentation mask on one or more \emph{target frames} based on the foreground from the \emph{reference frame}. It is evaluated with the intersection over union (IoU). 

\myparagraph{Segmentation with SAM}
SAM~\citep{kirillov2023sam, ravi2024sam2} is a supervised pretrained model that, given an input image and point prompts, generates segmentation masks. 
We uplift SAM masks from multiple views using Eq.~(\ref{eq:uplifting}); this aggregation improves cross-view consistency and leads to better single-view segmentation. On each view, SAM is run multiple times with different prompt sets obtained by uplifting and reprojecting the reference mask, and the resulting predictions are averaged (see Sec.~A.1).
The final prediction is obtained by rendering the 3D mask into the target frame and thresholding.
Results obtained with this strategy are reported for SAM~\citep{kirillov2023sam} and SAM2~\citep{ravi2024sam2} in Sec.~\ref{sec:exp-seg}. 

\myparagraph{Segmentation with DINOv2}
We construct 2D feature maps using DINOv2 with registers~\citep{darcet2024registers}, applying a sliding window over the image followed by dimensionality reduction of the DINOv2 features.
To favor the first principal components, known to focus on the foreground objects~\citep{oquab2024dinov2}, the features are re-weighted by the eigenvalues of the PCA decomposition.
The 2D feature maps from the $m$ training views are uplifted using Eq.~(\ref{eq:uplifting}) and the resulting 3D features are re-projected into any direction using Eq.~(\ref{eq:rendering}).
2D segmentation is obtained by thresholding the similarity between projected features and those from the foreground pixels $\mathcal{P}$ in the reference frame (see Sec.~A.2).
This approach corresponds to an ablation discussed in Sec.~\ref{sec:exp-seg}.

\myparagraph{Improving DINOv2 segmentation with graph diffusion}
Segmentation based on DINOv2 alone may struggle to distinguish objects with similar features, as it lacks 3D spatial information.
To address this, we leverage the graph diffusion process introduced in Sec.~\ref{sec:diffusion} and illustrated in Fig.~\ref{fig:diffusion}. 
The foreground mask from the reference frame is first uplifted into 3D, defining a set of anchor Gaussians $\mathcal{M}$. This set is used to initialize the weight vector $g_0$ and to define the regularization term $P$, based on the similarity between each 3D feature and those of Gaussians in $\mathcal{M}$ (see supplementary Sec.~\ref{sec:appendix-diffusion}).
For this task, we binarize $A$ with a fixed threshold (set to $10^{-5}$). After $T$ diffusion steps, we recover the  nodes $\mathcal{S}$ in $g_T$ with strictly positive values (\emph{i.e.}, those reachable after $T$ steps). The final weight is defined as $h_i = P(f_i)$ if $i\in \mathcal{S}$ and $0$ otherwise. Segmentation is then performed by projecting $\mathbf{h} = (h_i)$ into 2D and thresholding.
This approach achieves results comparable to SAM, as reported and discussed in Sec.~\ref{sec:exp-seg}.

\subsection{Open-vocabulary object segmentation}
\label{sec:method_clip}

Following \cite{kerr2023lerf}, we tackle the task of open-vocabulary object localization by uplifting CLIP features~\citep{ilharco2021openclip}.
CLIP effectively aligns images and text in a shared representation space. As a measure of alignment, we use the relevancy score introduced by LERF~\citep{kerr2023lerf}, which measures the similarity between a CLIP visual feature and a text query.

\myparagraph{Construction of CLIP feature maps}
We follow common practice \citep{kerr2023lerf, zuo2024fmgs} and construct multi-resolution CLIP 2D feature maps by querying CLIP on a grid of overlapping patches at different scales. 
As in \cite{zuo2024fmgs}, rather than keeping the resulting representations separate, we aggregate them via average pooling. These multi-resolution CLIP features are uplifted into 3D using Eq.~(\ref{eq:uplifting}).

\myparagraph{Relevancy scores} 
After uplifting CLIP features, we compute relevancy scores for each Gaussian's feature to text queries embedded by CLIP. These relevancy scores can then be projected into 2D and used for both localization and segmentation. For localization, we choose the pixel with the highest relevancy score.
For segmentation, we render the relevancy scores and either (i) directly threshold the rendered scores, or (ii) predict a SAM mask by selecting point prompts among pixels with the highest score. Specifics on the computation of relevancy scores and segmentation masks are provided in the supplementary (Sec.~\ref{sec:appendix-clip}).

\myparagraph{Refining relevancy scores with DINOv2 graph diffusion}
We refine 3D relevancy scores with the diffusion process described in Sec.~\ref{sec:diffusion}. To this end, DINOv2 features are also uplifted, and the similarity matrix is built as in Eq.~(\ref{eq:edges}), with the unary term $P$ constructed using a logistic regression over thresholded relevancies, see details in the supplementary (Sec.~\ref{sec:appendix-clip}). The diffusion process propagates CLIP relevancies to Gaussians with similar DINOv2 features. The resulting 3D relevancy scores span the object of interest without covering other objects with similar features and show a strong decay at the object's borders, as defined by DINOv2's feature landscape, resulting in improved segmentation results. 
This approach is evaluated and compared to direct segmentation without diffusion in Sec.~\ref{sec:exp-clip}.

\subsection{Open-vocabulary semantic segmentation}
\label{sec:method_semantic}

In semantic segmentation, one needs to label each Gaussian with the appropriate text label from a predefined set.
For this task, we directly replace OpenGaussian’s 3D feature learning with our uplifting approach, keeping the rest of their protocol unchanged. This enables a fair comparison under the same evaluation setup.
In particular, 2D feature map generation is performed using SAM in automatic mask generation mode and assigning a CLIP feature to each mask, as in~\citep{qin2023langsplat, shi2024legaussians, wu2024opengaussian}.
After uplifting, each Gaussian is assigned 
the text label with the highest CLIP similarity.
Results are reported in Sec.~\ref{sec:exp-semantic}.
\section{Experiments}
\label{sec:exp}

In the following experiments, we train all scenes using the original Gaussian Splatting implementation~\citep{kerbl2023gaussiansplatting} with default hyperparameters. To reduce memory usage, half of the Gaussians are filtered out based on their importance, as described in Sec.~\ref{sec:uplifting}.

We uplift features from DINOv2 ViT-g with registers~\citep{darcet2024registers}, SAM~\citep{kirillov2023sam}, SAM 2~\citep{ravi2024sam2}, and OpenCLIP ViT-B/16~\citep{ilharco2021openclip}.
Key parameters (\eg the segmentation threshold and the definitions of $S_f$ and $P$ in graph diffusion) are either fixed across all scenes of a task or automatically selected, as detailed in the supplementary material (Secs.~\ref{sec:appendix-sam_seg},  \ref{sec:appendix-diffusion}, \ref{sec:appendix-clip}).
Results are averaged over three independent runs.

\subsection{Qualitative results}

\begin{figure}[t!]
    \centering
    \begin{subfigure}{.32\linewidth}
        \includegraphics[width=\linewidth]{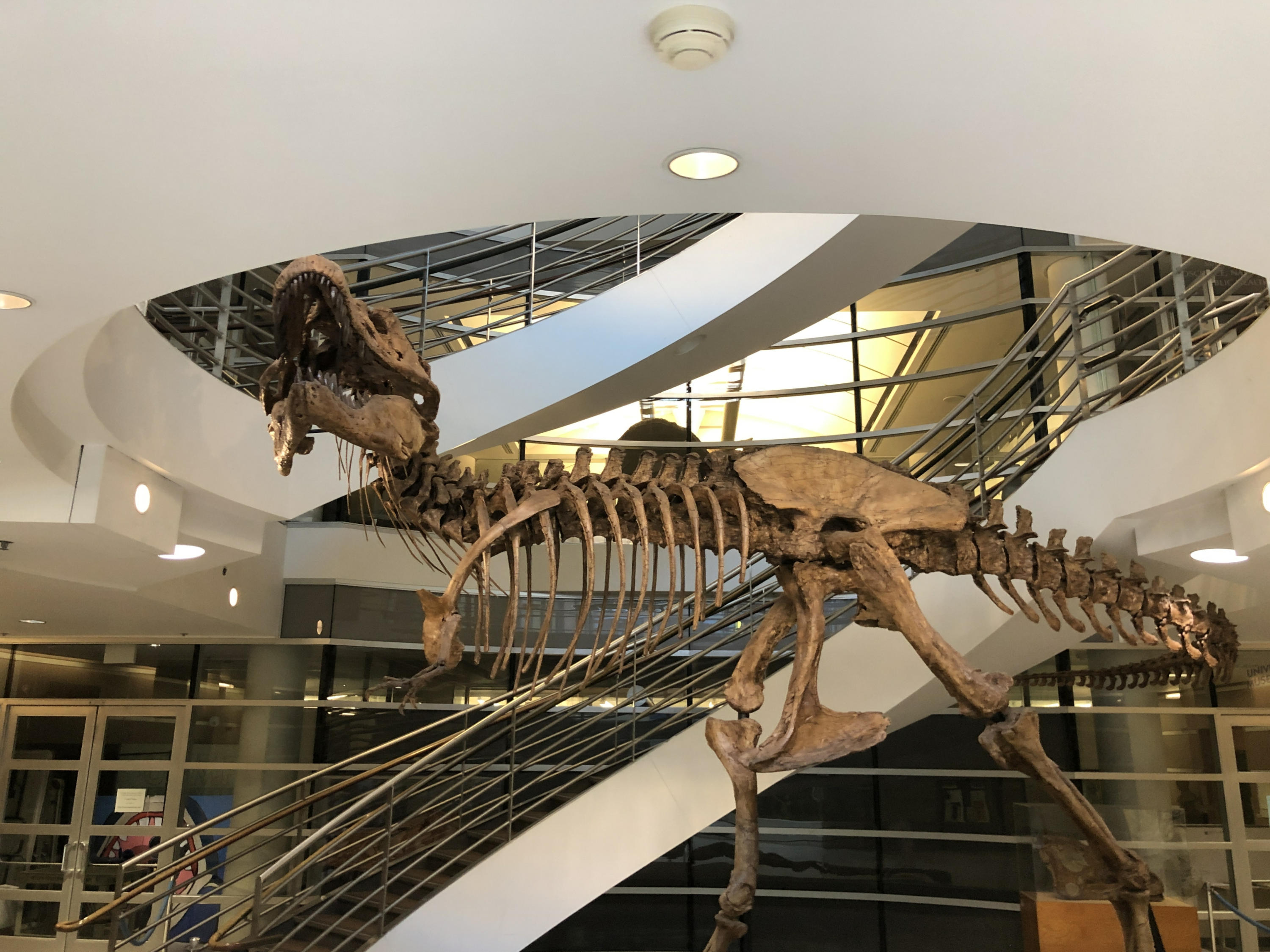}
        \includegraphics[width=\linewidth]{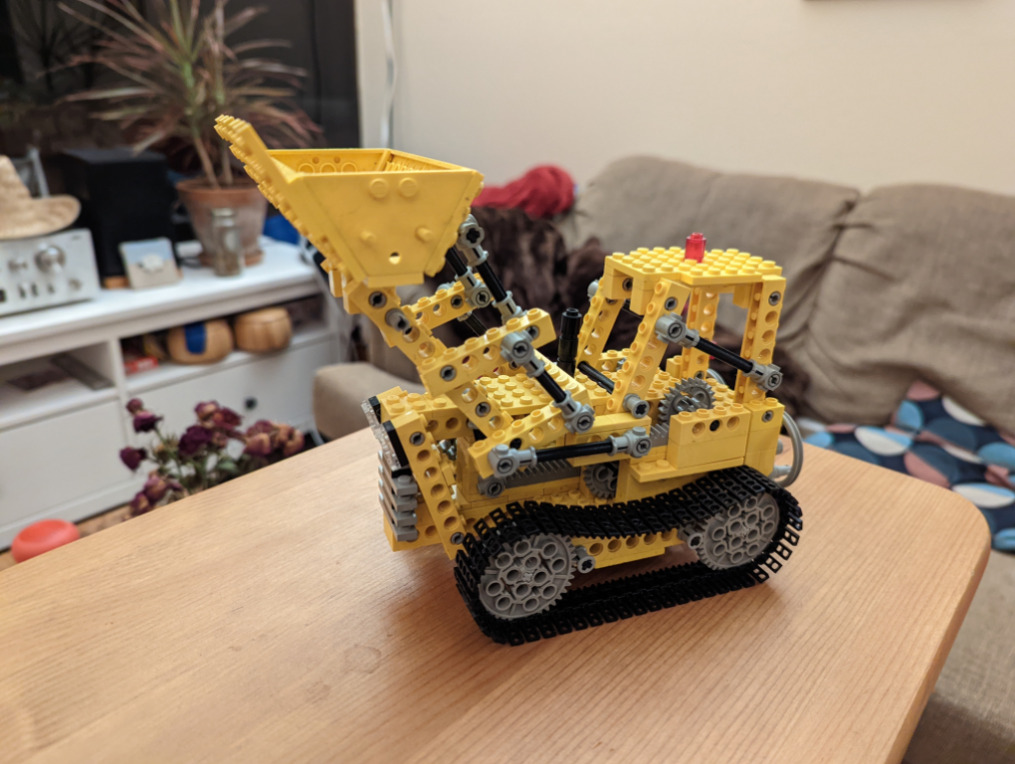} 
        \caption{RGB image}
    \end{subfigure}
    \begin{subfigure}{.32\linewidth}
        \includegraphics[width=\linewidth]{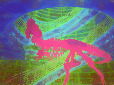}
        \includegraphics[width=\linewidth]{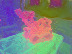} 
        \caption{Single-view PCA}
    \end{subfigure}
    \begin{subfigure}{.32\linewidth}
        \includegraphics[width=\linewidth]{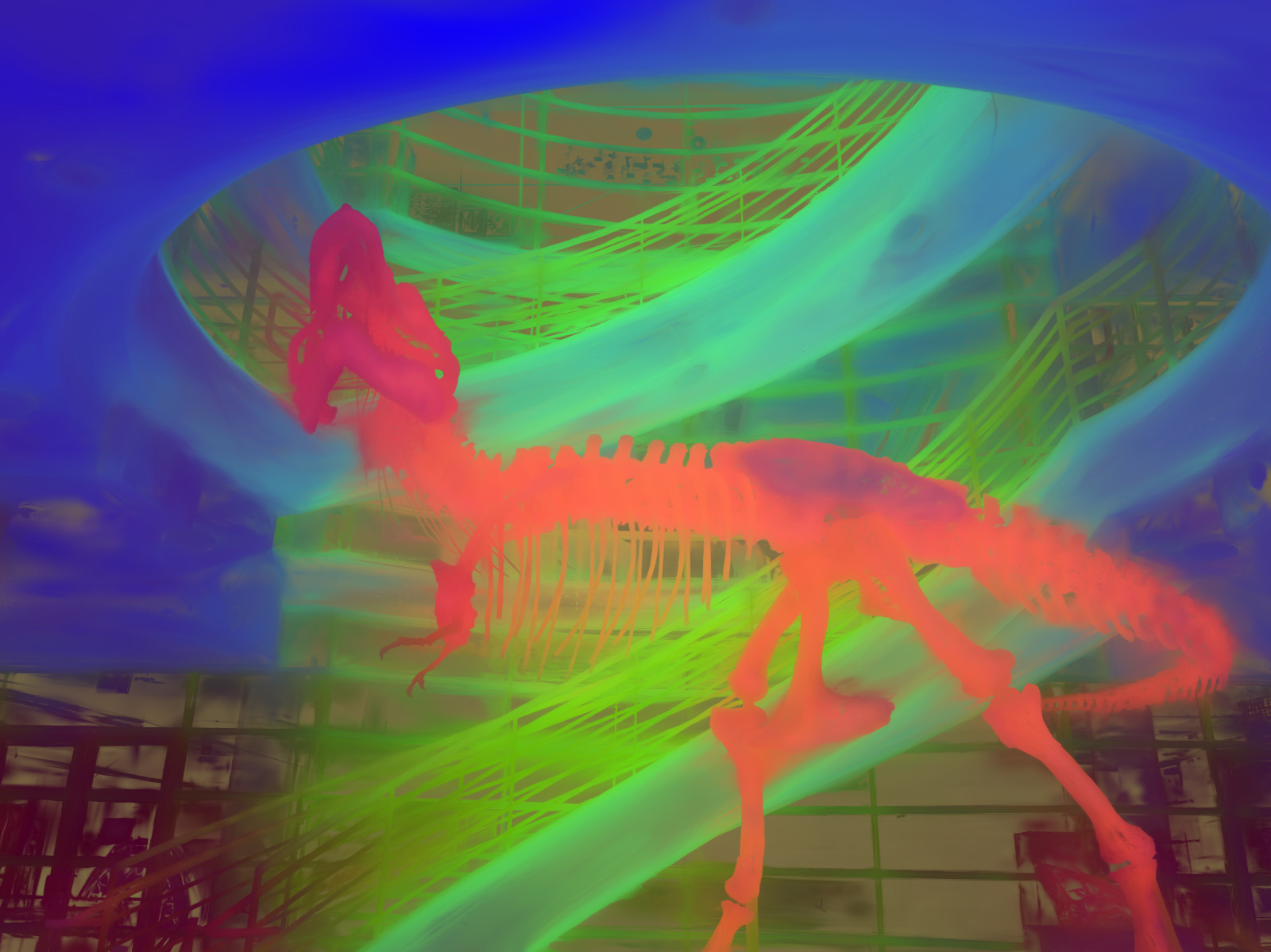}
        \includegraphics[width=\linewidth]{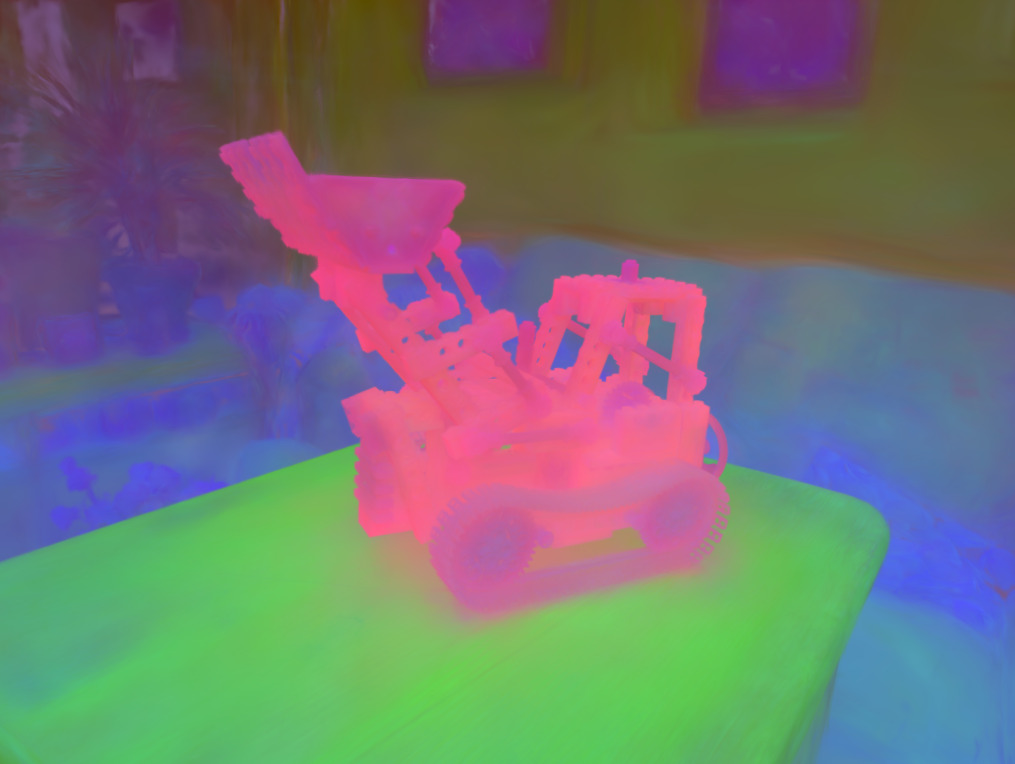} 
        \caption{Multi-view PCA}
    \end{subfigure}
    \caption{\textbf{PCA visualizations.} The DINOv2 patch-level representations (middle) predicted from the RGB images (left) are aggregated into highly detailed 3D representations (right) using Eq.~(\ref{eq:uplifting}).}
    \label{fig:pca}
\end{figure}

\paragraph{DINOv2 feature uplifting.}
First, we illustrate the effectiveness of our simple uplifting approach. Fig.~\ref{fig:pca} shows the first three PCA components (one channel per component) over DINOv2's patch embeddings. The coarse patch-level representations from every view (middle) are aggregated using Eq.~(\ref{eq:uplifting_matrix}) to form a highly detailed 3D semantic representation, and reprojected into 2D (right) using Eq.~(\ref{eq:rendering}). The aggregation is very fast, being directly implemented in the Gaussian Splatting CUDA-based rendering process, and takes about 2ms per view and feature dimension. The first principal component (encoded in the red channel) mostly captures the foreground object, and the subsequent ones allow refining the foreground representations and delivering a detailed background.
In the supplementary, we provide additional comparative visualizations of our learned 3D features 
and of 3D segmentation for object removal. 

\begin{figure}[t!]
    \centering
    \begin{subfigure}{.24\linewidth}
    \includegraphics[width=\linewidth]{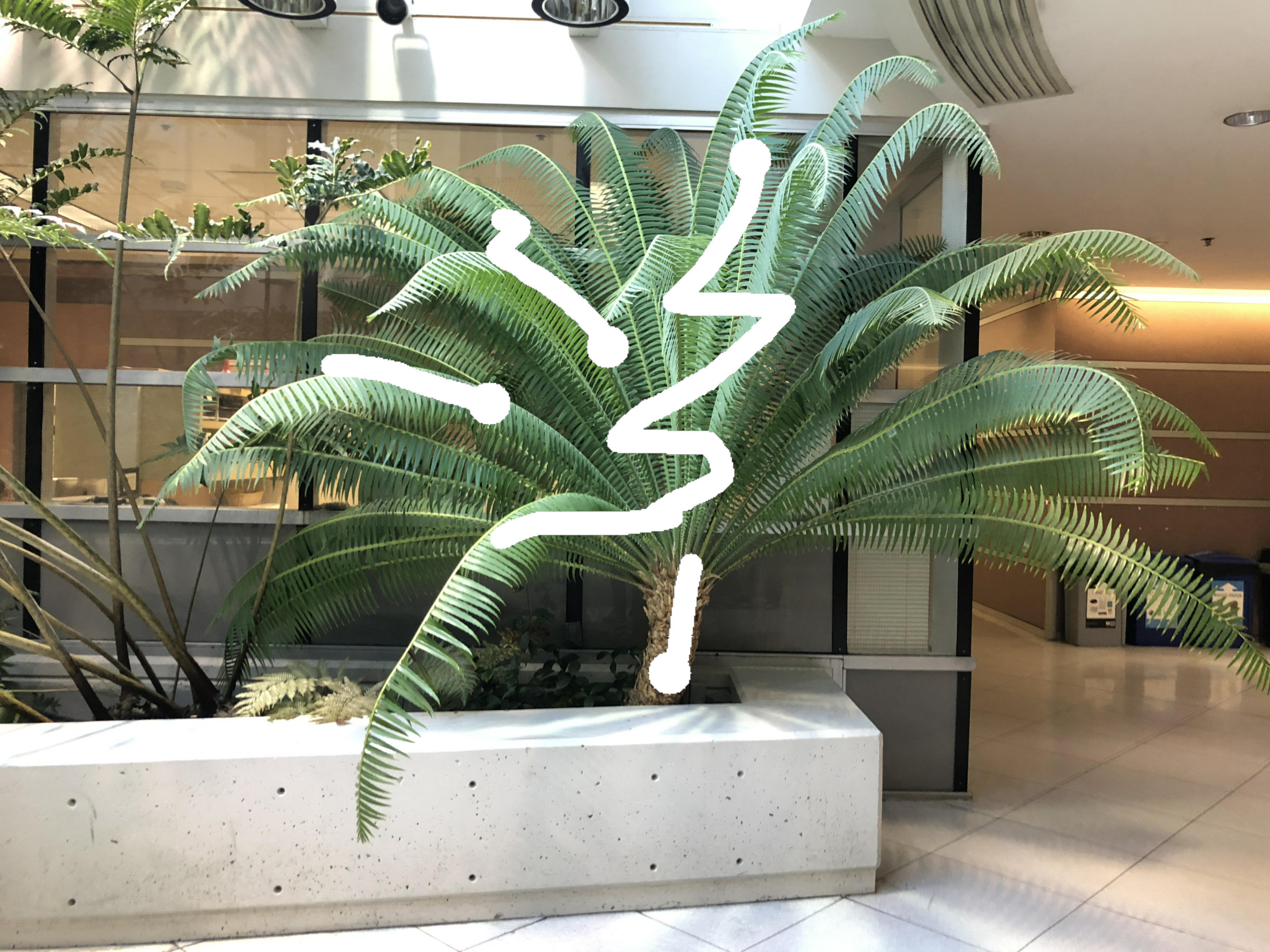}
    \includegraphics[width=\linewidth]{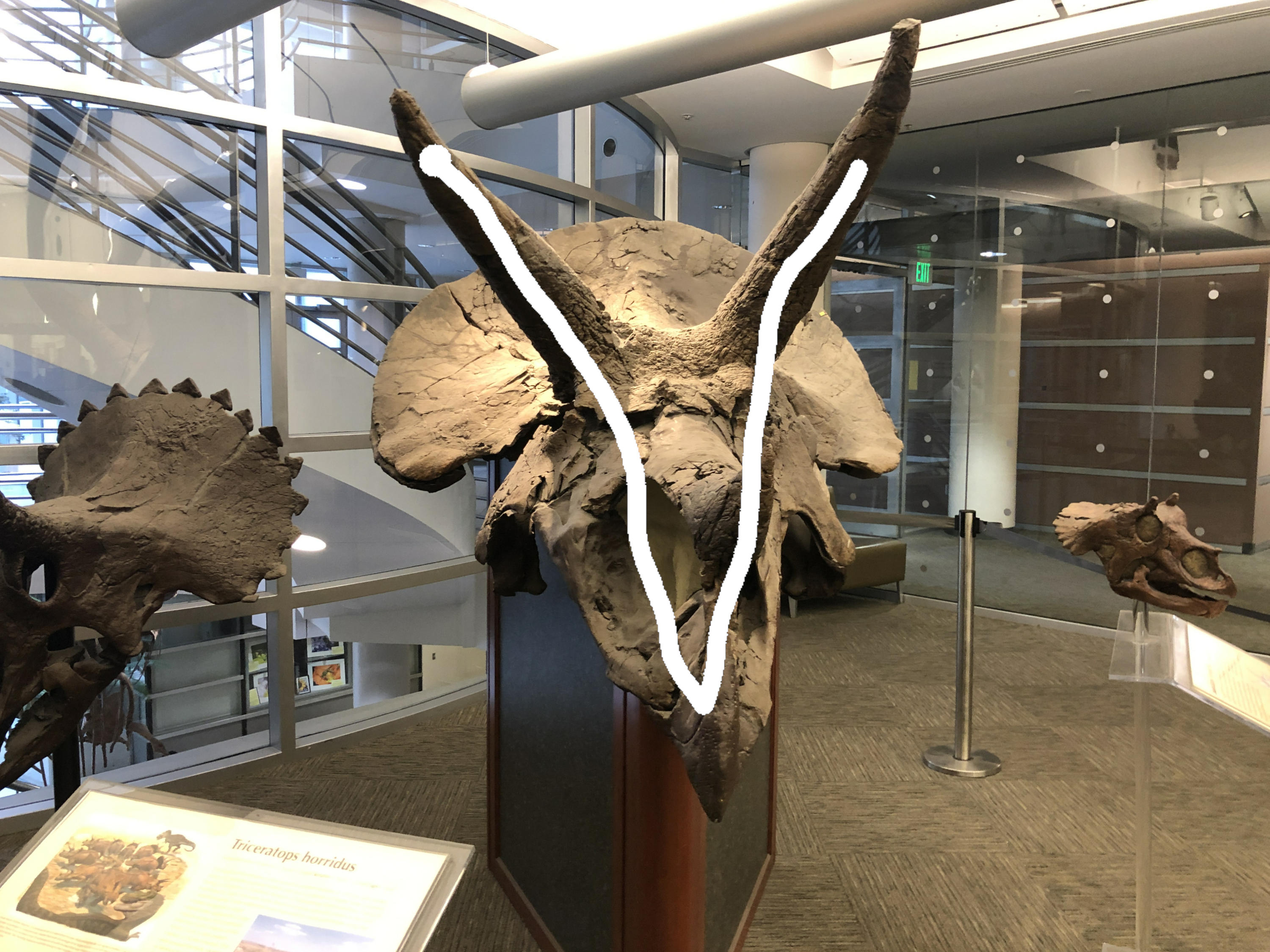}
    \subcaption{RGB image}
    \end{subfigure}
    \begin{subfigure}{.24\linewidth}
    \includegraphics[width=\linewidth]{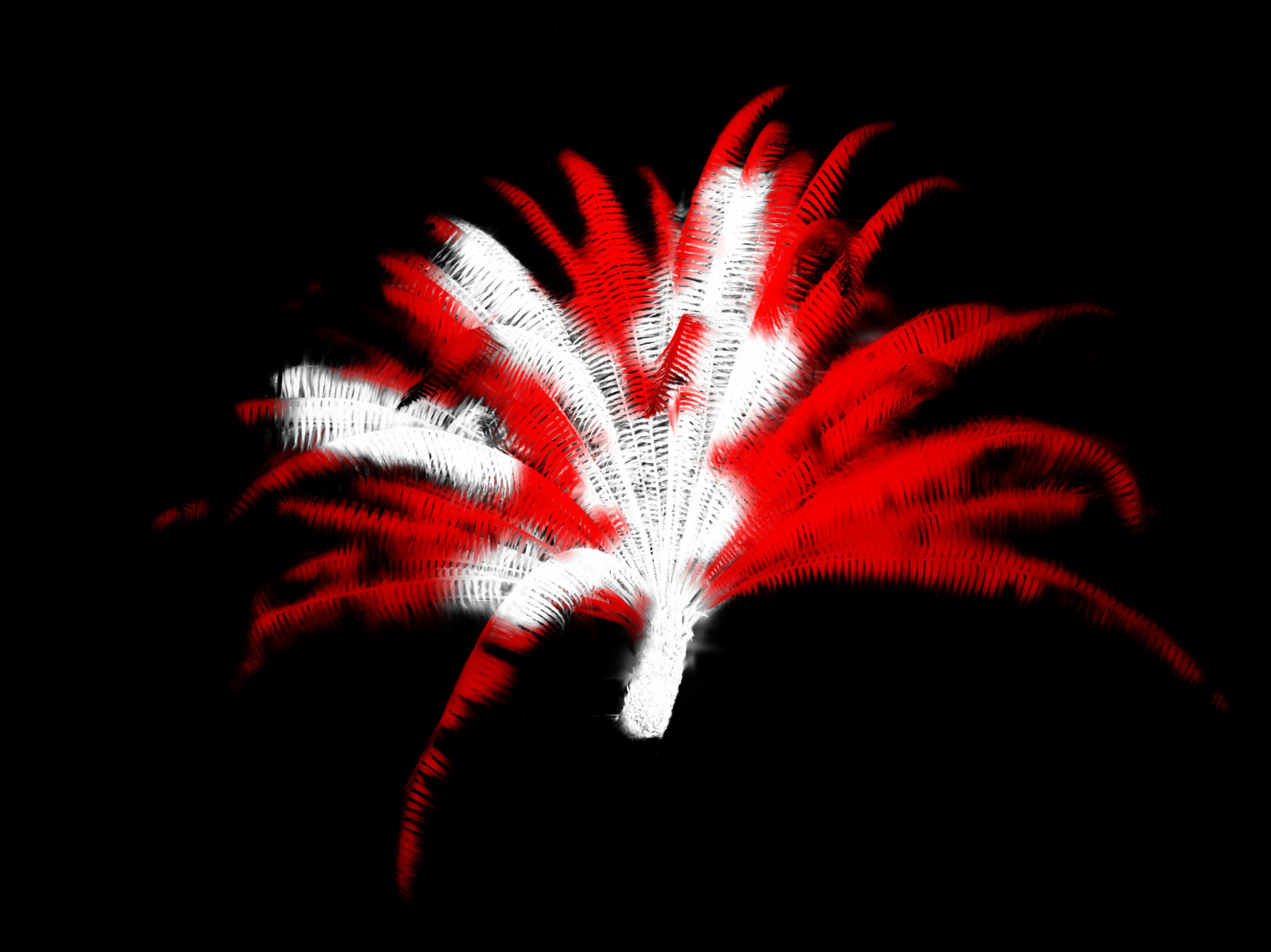} 
    \includegraphics[width=\linewidth]{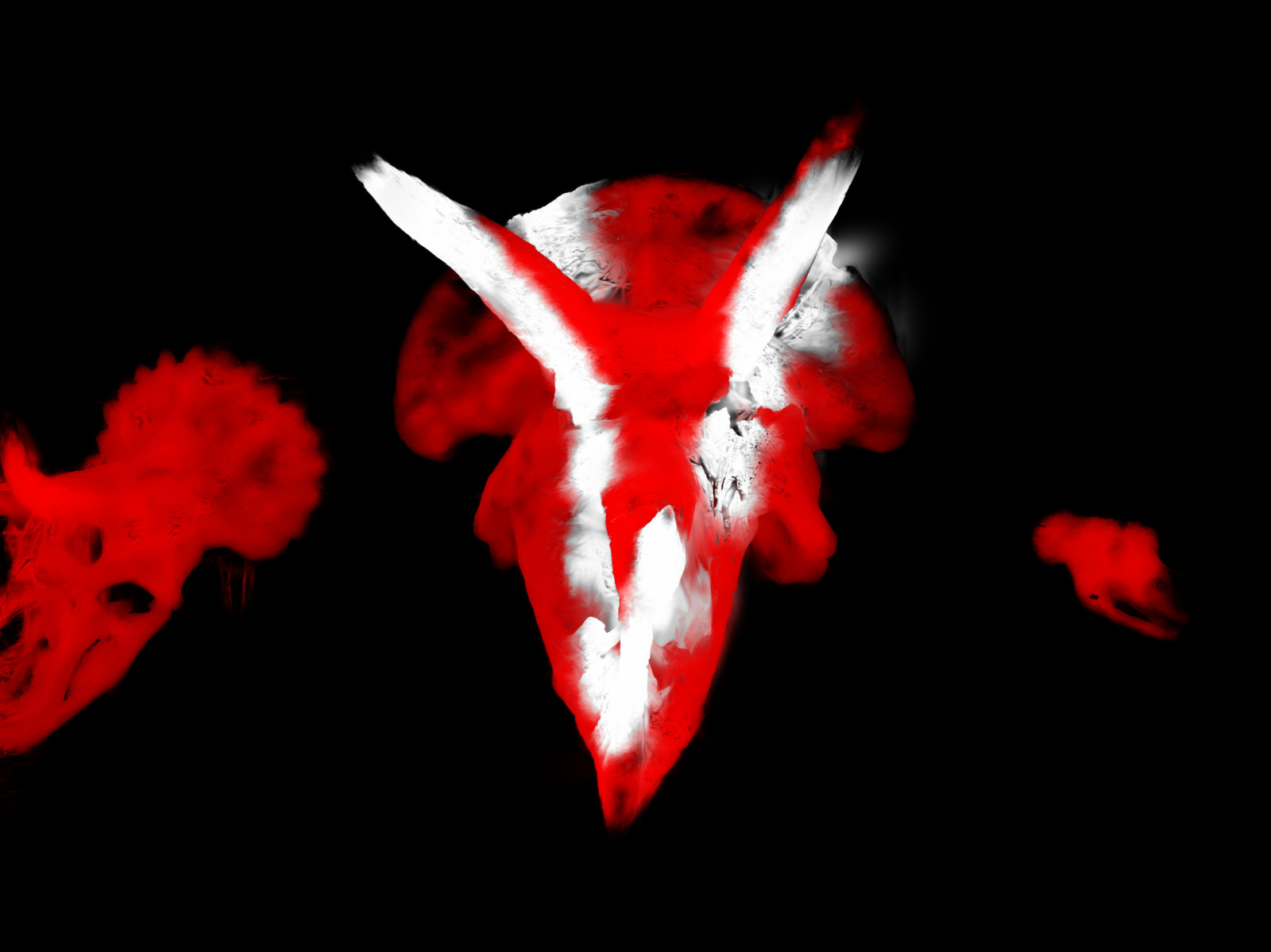}
    \caption{3 steps}
    \end{subfigure}
    \begin{subfigure}{.24\linewidth}
    \includegraphics[width=\linewidth]{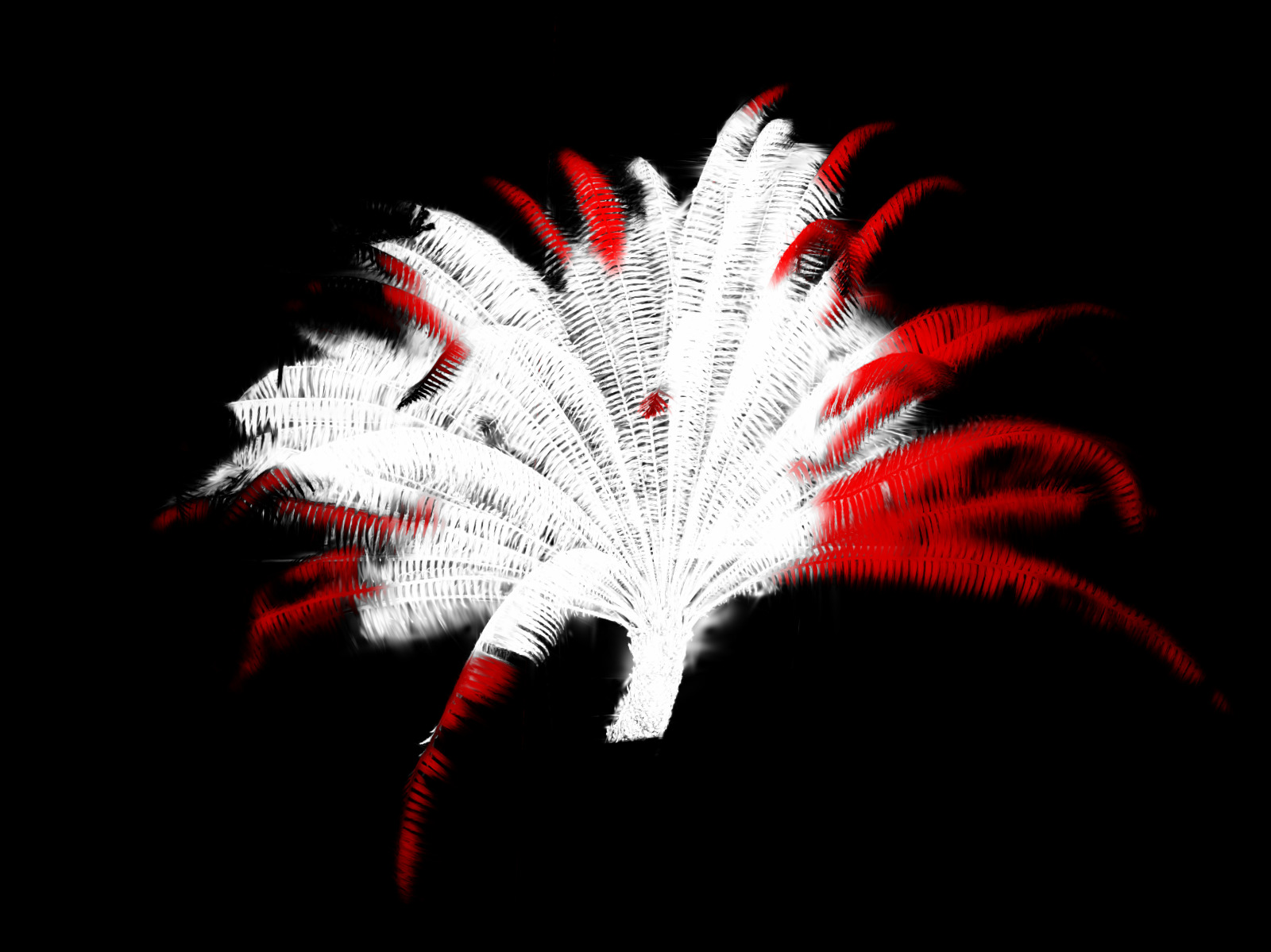}
    \includegraphics[width=\linewidth]{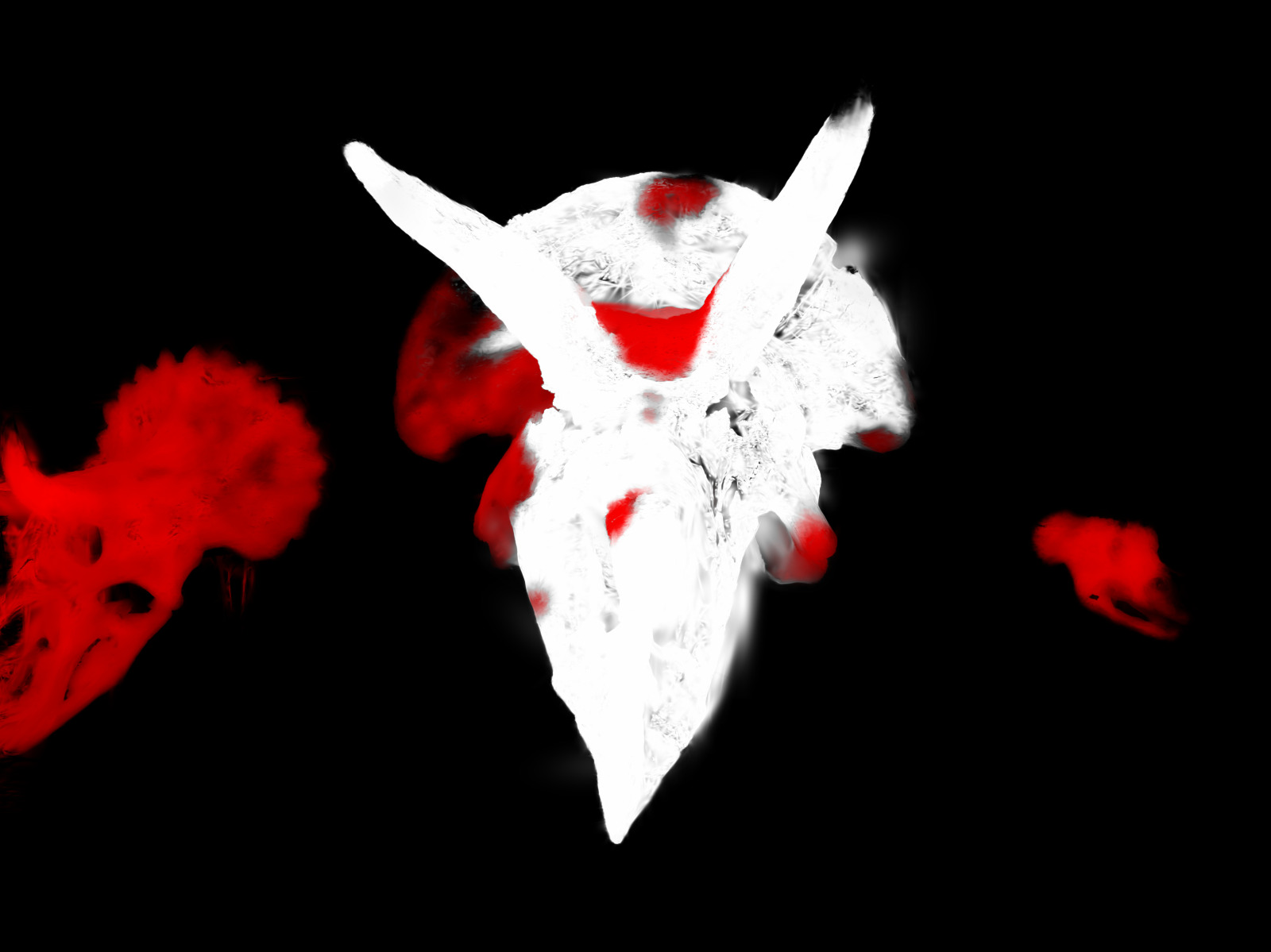}
    \subcaption{7 steps}
    \end{subfigure}
    \begin{subfigure}{.24\linewidth}
    \includegraphics[width=\linewidth]{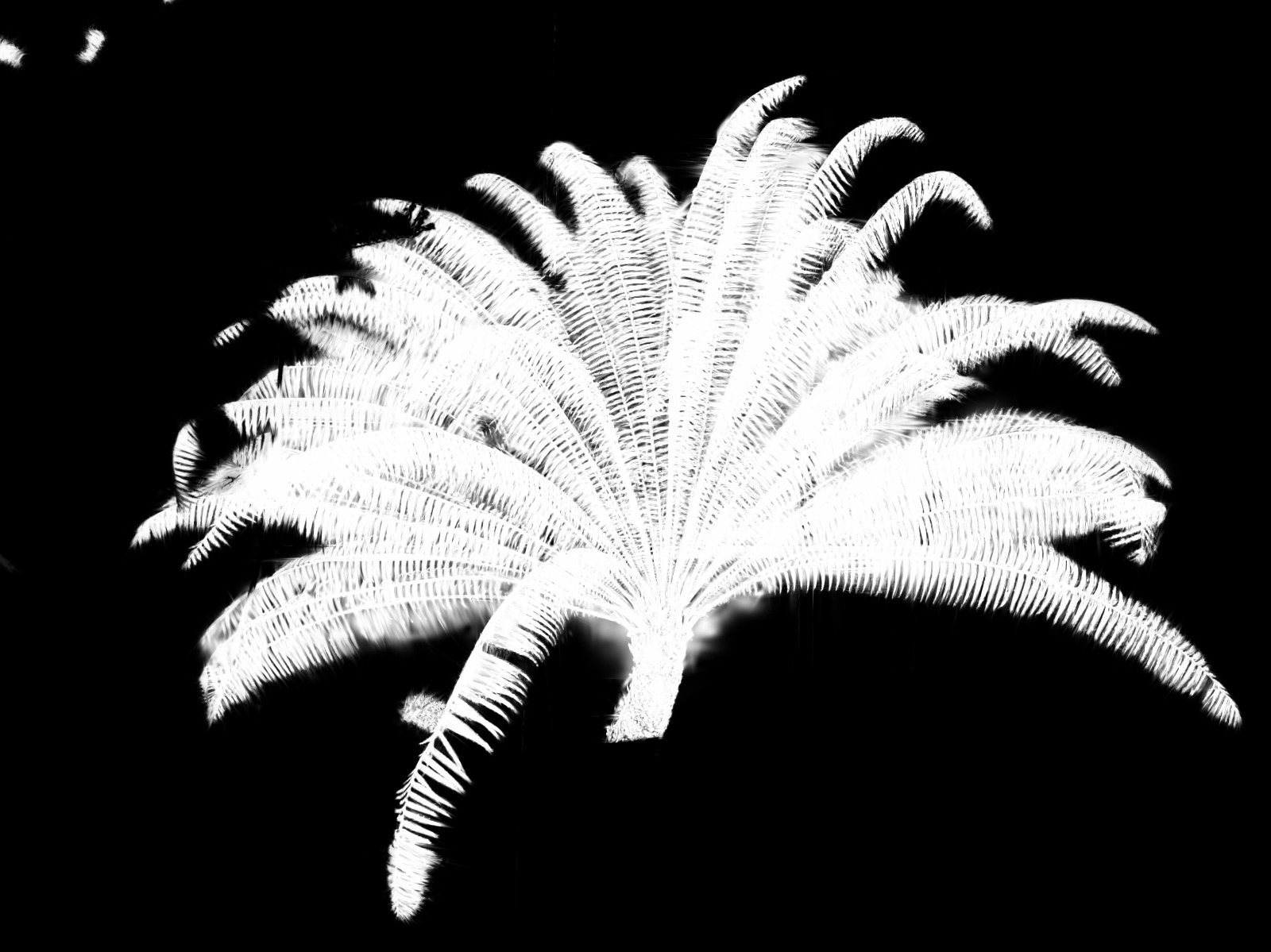} 
    \includegraphics[width=\linewidth]{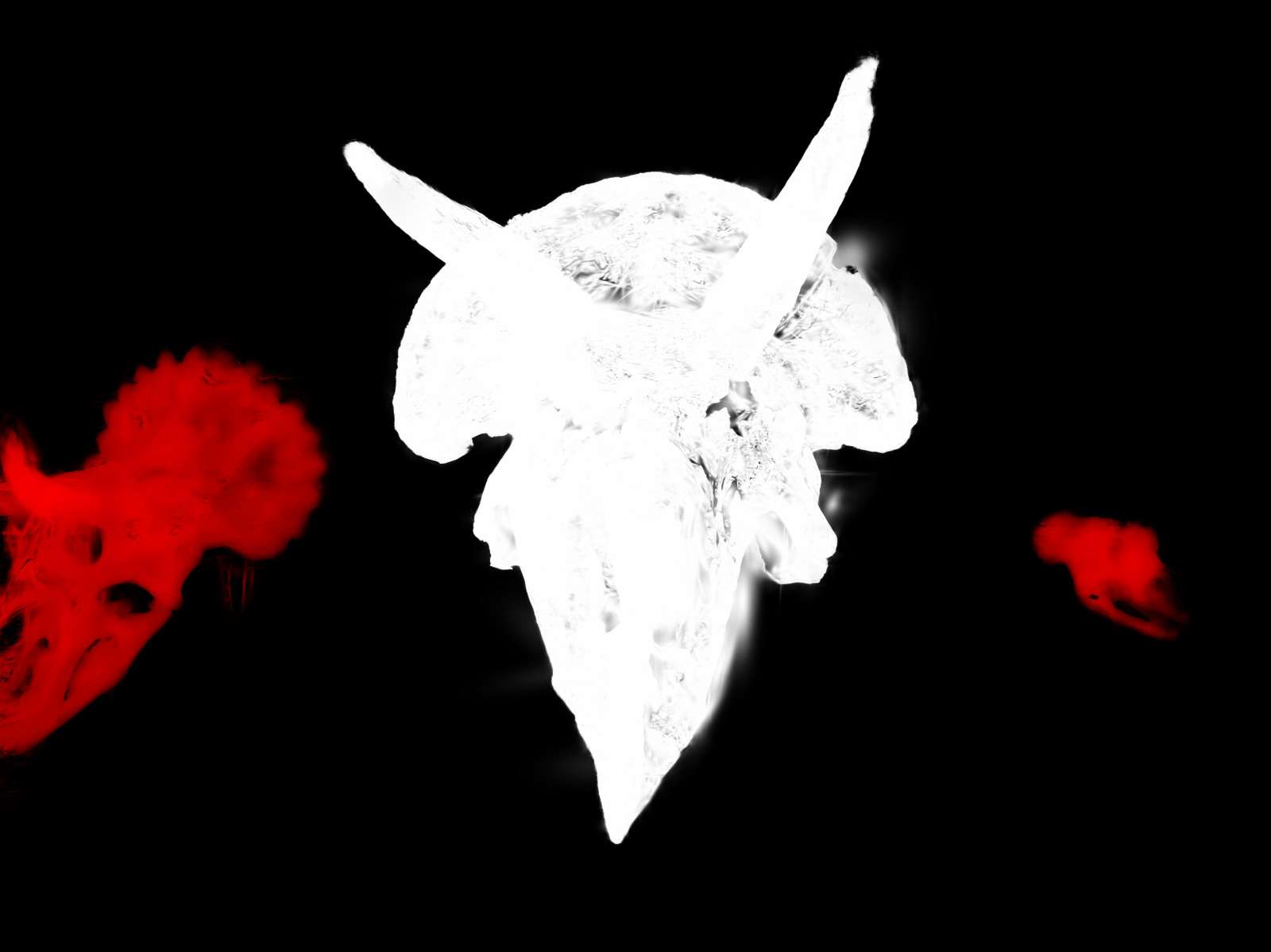}
    \caption{100 steps}
    \end{subfigure}
    \caption{\textbf{Diffusion process.} 2D projection of the weight vector $g_t$ (white) and unary regularization term (red) at diffusion steps $t$. The diffusion process filters out unwanted objects that have similar features to the object of interest (such as the two smaller skulls on \emph{horns}, bottom-row), but are disconnected in space. The regularization term (red background) prevents leakage from the object to the rest of the scene (such as through the \emph{fern}'s trunk, top-row).}
    \vspace{-.1cm}
    \label{fig:diffusion}
\end{figure}

\myparagraph{Graph diffusion}
Fig.~\ref{fig:diffusion} illustrates the diffusion process. The graph nodes are initialized with the reference scribbles, and the diffusion spreads through the object of interest.
As illustrated with the case of Horns, diffusion filters out unwanted objects that are similar to the object of interest (here, the two skulls on the side). In the Fern scene, diffusion progressively spreads through the branches to their extremities, with the regularization (red background) constraining it to the trunk and preventing leakage, even after a large number of iterations. Fig.~\ref{fig:appendix-diffusion} in the supplementary also illustrates this for the Flower and Trex scenes: diffusion rapidly spreads, achieving near-full coverage after only 5 steps before reaching all the much smaller Gaussians on the border, allowing for a refined segmentation.

\begin{figure}[t]
    \centering
    \includegraphics[width=0.32\linewidth, trim=0 500 0 200, clip]{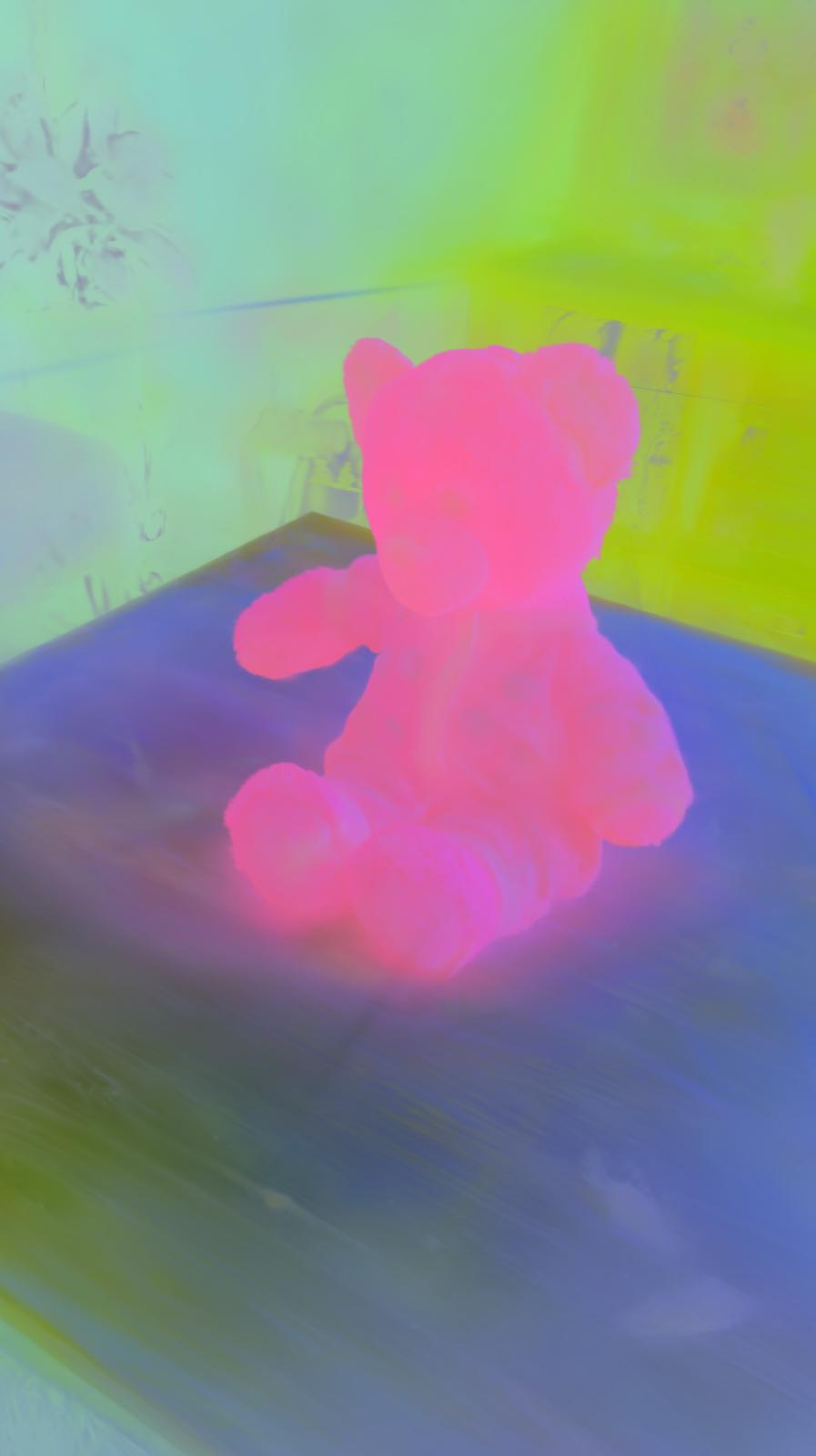}
    \includegraphics[width=0.32\linewidth, trim=0 610 0 200, clip]{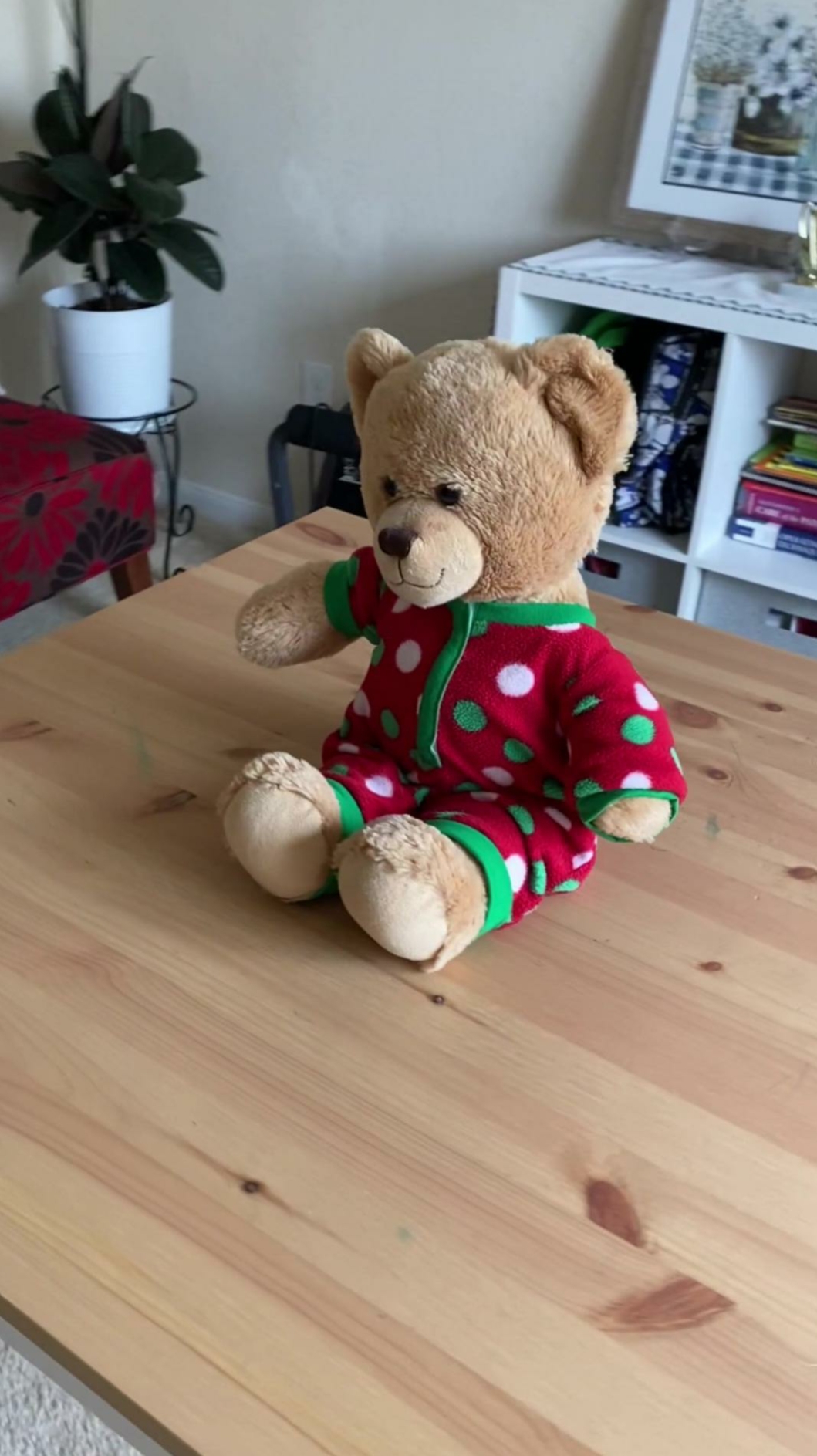}
    \includegraphics[width=0.32\linewidth, trim=0 500 0 200, clip]{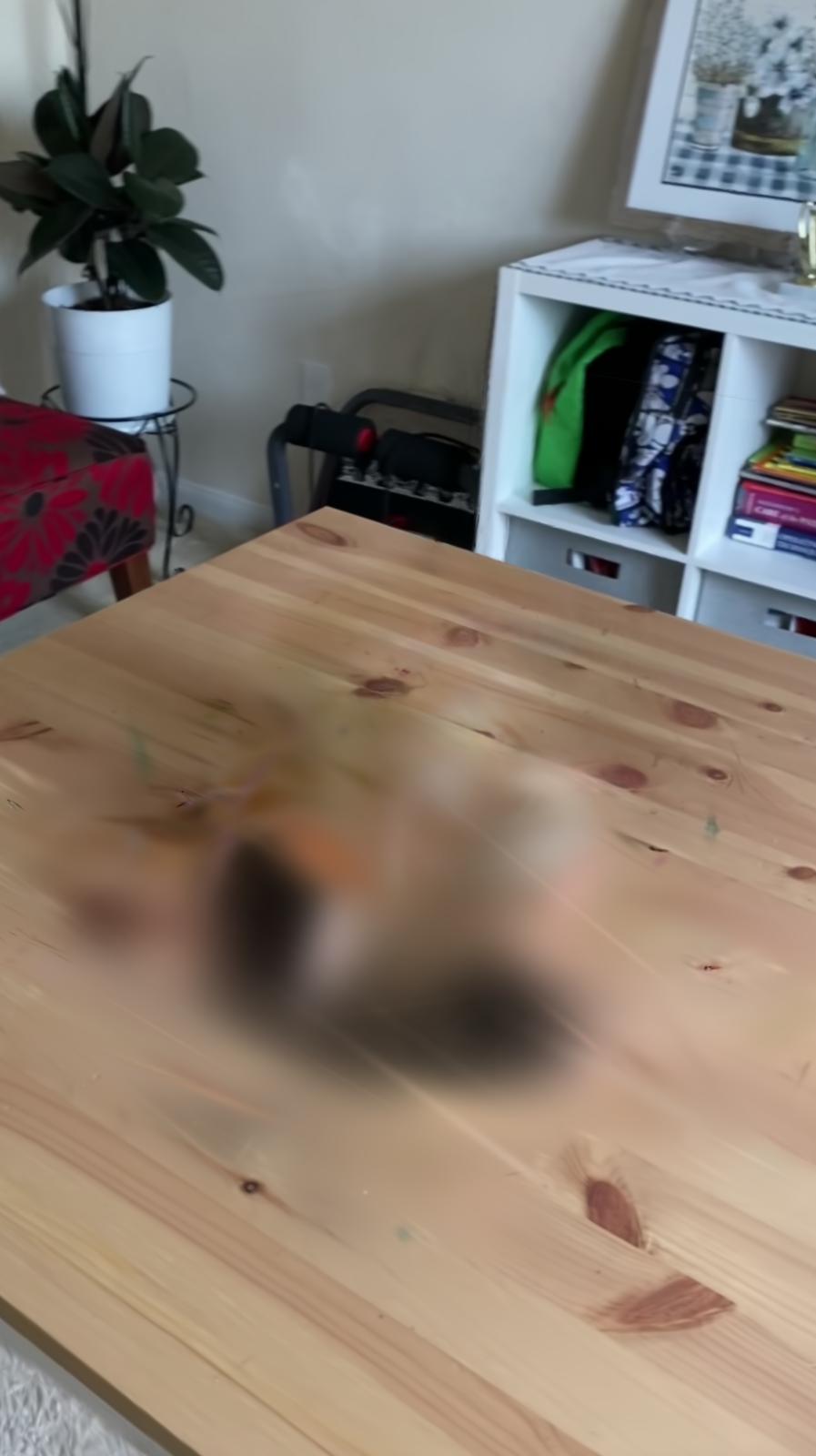}
    \caption{\textbf{Open-vocabulary object removal.} Removing the teddy bear from the CO3D dataset~\citep{reizenstein2021co3d}, using DINOv2-guided graph diffusion based on 3D CLIP relevancy scores.}
    \vspace{-.3cm}
    \label{fig:removal-bonsai}
\end{figure}

\myparagraph{Open-vocabulary object removal} Fig.~\ref{fig:teaser} and~\ref{fig:removal-bonsai} provide qualitative results for the object removal task, performed by discarding all Gaussians corresponding to a binary 3D segmentation mask. 
This mask is obtained by computing relevancy between 3D CLIP features and the scene-specific textual prompt ``bonsai in a ceramic pot''  or ``teddy bear'', followed by DINOv2-guided graph diffusion.

\subsection{Multi-view segmentation results}
\label{sec:exp-seg}

\myparagraph{Task} In this section, we consider two segmentation tasks: (i)~Neural Volumetric Object Selection (NVOS)~\cite{ren2022nvos}, which is derived from the LLFF dataset~\citep{mildenhall2019llff}, and (ii)~SPIn-NeRF, which contains a subsets of scenes from NeRF-related datasets~\citep{knapitsch2017tanks, mildenhall2019llff, mildenhall2021nerf, yen2022nerfsupervision, fridovich2022plenoxels}.
NVOS consists of forward-facing sequences where the task is to predict the segmentation mask of a labeled frame based on reference scribbles from another frame. 
SPIn-NeRF contains both forward-facing and 360-degree scenes, in which all frames are labeled with segmentation masks, and the standard evaluation protocol uses the segmentation mask from the first frame as reference to label the subsequent frames.

\myparagraph{Setup}
We evaluate segmentation based on SAM and SAM2 mask uplifting and on DINOv2 feature uplifting combined with graph diffusion (see Sec.~\ref{sec:method_seg}). 
We compare our segmentation results (IoU) to the current state of the art: SA3D~\citep{cen2023segmentNeRF}, SA3D-GS~\citep{cen2023segmentGS}, SAGA~\citep{cen2023saga}, OmniSeg3D~\citep{ying2023omniseg3d}. All these methods are specifically designed for uplifting the 2D segmentation masks produced by SAM into 3D using gradient-based optimization of a projection loss. We also report results from NVOS \citep{ren2022nvos} and MVSeg \citep{yen2022nerfsupervision}, who respectively introduced the NVOS and SPIn-NeRF datasets.

\myparagraph{Results} Table~\ref{tab:nvos-spin} reports the average IoU across all scenes for NVOS and SPIn-NeRF. Per-scene results can be found in supplementary Tables \ref{tab:spin} and \ref{tab:nvos}.
Our results are comparable to the state of the art, while not relying on gradient-based optimization. 
Surprisingly, our segmentation with DINOv2 using graph diffusion also gives results on par with models leveraging SAM masks. Compared to SAM, DINOv2 better captures complex objects but sometimes also captures background noise. This can be seen in supplementary Fig.~\ref{fig:appendix-nvos}, T-Rex example: while SAM misses the end of the tail and ribs, DINOv2 captures the whole T-Rex but also includes part of the stairs behind.
Our lower segmentation results compared to OmniSeg's can partly be attributed to the poor Gaussian Splatting reconstruction of highly specular scenes, such as Fork. As also noted by \cite{cen2023saga}, the reconstruction includes semi-transparent Gaussians floating over the object, attempting to represent reflections or surface effects, which are challenging to capture using standard rasterization techniques~\citep{jiang2024gaussianshader}.

\myparagraph{Ablation study}
We compare our segmentation protocol using DINOv2 and SAM2 to multiple simpler variants. More precisely, we evaluate (i) a purely geometrical variant that reprojects the reference mask on the other views, without using SAM2 or DINOv2, (ii) single-view segmentation in 2D based on SAM2 or DINOv2 2D predictions, (iii) uplifting DINOv2 features or SAM2 masks into 3D then rendering them for segmentation, and (iv) segmenting using graph diffusion over DINOv2 3D feature similarities. Results are reported in Table~\ref{tab:ablation-spin-small}, and per-scene IoU as well as a detailed analysis can be found in supplementary Table~\ref{tab:ablation-spin-dino-sam} and Sec.~\ref{sec:appendix-seg_scene}. We observe that the purely geometrical approach works well on the forward-facing scenes and fails on 360-degree scenes. The single-view variant performs reasonably well on average, but the low resolution of patch-level representations (illustrated in  Fig.~\ref{fig:pca}) lead to a coarser segmentation. 3D uplifting considerably boosts results compared to single-view approaches, and introducing 3D spatial information through 3D graph diffusion further enhances results on the more challenging 360-degree scenes.

\begin{table*}[t!]
  \centering
  \footnotesize
\scalebox{0.9}{
\begin{tabular}{lccccccccc}
\toprule
& NVOS~\citep{ren2022nvos} & MVSeg~\citep{mirzaei2023spin} & SA3D-TRF~\citep{cen2023segmentNeRF} & SA3D-GS~\citep{cen2023segmentGS} & SAGA~\citep{cen2023saga} & OmniSeg3D~\citep{ying2023omniseg3d} & \multicolumn{3}{c}{\textbf{LUDVIG (Ours)}} \\
\cmidrule(lr){8-10}
3D representation &  &  & TensoRF & 3D-GS & 3D-GS & NeRF &  \multicolumn{3}{c}{3D-GS} \\
Uplifting & & & SAM & SAM & SAM & SAM &  DINOv2 & SAM & SAM2 \\
\midrule
NVOS & 70.1 & - & 90.3 & 92.2 & \textbf{92.6} & 91.7 & 92.4 & 91.3 & 91.3 \\
SPIn-NeRF & - & 90.9 & 93.7 & 93.2 & 93.4 & \textbf{94.3} & 93.8 & 93.8 & 93.8 \\
\bottomrule
\end{tabular}
}
  \caption{\textbf{Multi-view segmentation (IoU) on NVOS~\citep{ren2022nvos} and SPIn-NeRF~\citep{mirzaei2023spin}.} 
  }
  \label{tab:nvos-spin}
\end{table*}

\begin{table}[t!] 
\centering \footnotesize 
\scalebox{.85}{
\begin{tabular}{cccccc} 
\toprule
Geometry only & \multicolumn{2}{c}{Single view} & \multicolumn{2}{c}{Uplifting} & \multicolumn{1}{c}{Uplifting + Dif.} \\
\cmidrule(lr){2-3} \cmidrule(lr){4-6}
Reference mask & DINOv2 & SAM2 & DINOv2 & SAM2 & DINOv2 \\ 
\midrule
73.1 & 88.5 & 88.6 & 91.6 & 93.8 & 93.8 \\
\bottomrule 
\end{tabular} 
}
\caption{\textbf{Ablation on SPIn-NeRF segmentation}. We compare purely geometrical reference mask reprojection and single-view prediction with our feature/mask uplifting and graph diffusion.} 
\label{tab:ablation-spin-small} 
\end{table}
\begin{table}[t!]
\centering
\scalebox{0.87}{
\footnotesize
\begin{tabular}{lccccc}
\toprule
& ramen & figurines & teatime & waldo & \textbf{overall} \\ 
\midrule
LERF~\citep{kerr2023lerf}      & 62.0 & 75.0 & 84.8 & 72.7 & 73.6 \\ 
LangSplat~\citep{qin2023langsplat} & 73.2 & 80.4 & 88.1 & \textbf{95.5} & 84.3 \\ 
\textbf{LUDVIG (ours)}    &  \textbf{78.9} & \textbf{80.4} & \textbf{94.9} & 90.9 & \textbf{86.3} \\ 
\bottomrule
\end{tabular}
}
\vspace{-.1cm}
\caption{\textbf{LERF object localization.} Comparison with the state of the art on the dataset introduced by LangSplat~\citep{qin2023langsplat}.}
\label{tab:lerf-loc}

\end{table}

\begin{table}[t!]
\centering
\scalebox{0.85}{
\footnotesize
\begin{tabular}{lcccccc}
\toprule
& \multirow{2}{*}{ramen} & \multirow{2}{*}{figurines} & \multirow{2}{*}{teatime} & \multirow{2}{*}{waldo} & \multirow{2}{*}{\textbf{overall}} & time \\
& &&&&& (min) \\ \midrule
\multicolumn{7}{l}{\emph{Initial evaluation protocol from~\cite{kerr2023lerf,qin2023langsplat}: 2D object segmentation}} \\
\midrule
LERF~\citep{kerr2023lerf}      & 28.2 & 38.6 & 45.0 & 37.9 & 37.4 & 45  \\ 
LangSplat~\citep{qin2023langsplat} & 51.2 & 44.7 & 65.1 & 44.5 & 51.4 & 105 \\ 
LUDVIG (ours)    & \textbf{58.1} & \textbf{63.3} & \textbf{77.1} & \textbf{58.5} & \textbf{64.3} & 10 \\ 
\midrule 
\multicolumn{7}{l}{\emph{New evaluation protocol introduced in OpenGaussian~\cite{wu2024opengaussian}: 3D object selection}} \\
\midrule
OpenGaussian~\citep{wu2024opengaussian} & 23.9 & 59.3 & 54.4 & 34.6 & 43.1 & 50 \\ 
Dr. Splat~\citep{jun2025drsplat} & 24.7 & 53.4 & 57.2 & 39.1 & 43.6 & - \\ 
LUDVIG (ours)   & \textbf{42.3} & \textbf{58.0} & \textbf{58.6} & \textbf{42.8} & \textbf{50.4} & 10 \\ 
\bottomrule
\end{tabular}
}
\caption{\textbf{LERF object segmentation.} 
We evaluate on the LERF dataset introduced by LangSplat~\citep{qin2023langsplat} with two different evaluation protocols: i) segmentation based on the 2D reprojected features, ii) 3D segmentation (or \emph{object selection} and reprojection of the \emph{binary 3D masks}, proposed by \cite{kerr2023lerf, qin2023langsplat} and \cite{wu2024opengaussian} respectively.}
\label{tab:lerf-seg}

\end{table}

\subsection{Open-vocabulary object localization}
\label{sec:exp-clip}

\myparagraph{Task}
We evaluate on the LERF dataset~\citep{kerr2023lerf}, which includes localization and segmentation tasks on complex in-the-wild scenes. We report our results on the extended evaluation task introduced by LangSplat~\citep{qin2023langsplat} containing additional challenging localization samples.

\myparagraph{Setup}
For localization, we simply reproject the 3D CLIP relevancy scores and follow the evaluation protocol from LangSplat~\citep{qin2023langsplat}. 
For segmentation, we consider two evaluation protocols:
(i) \textit{standard 2D segmentation}, as introduced by~\citep{kerr2023lerf} (top part of the table), and (ii) \emph{3D object selection}, which consists of binarizing the 3D masks before rendering (bottom part). The latter protocol was introduced in~\citep{jun2025drsplat} as a way to better assess the quality of 3D semantic features, such as 3D CLIP features, compared to prior existing protocols which focus on the performance on 2D downstream tasks.
For both approaches, we refine raw 3D CLIP relevancy scores using graph diffusion and use them for segmentation as follows. For
(i) we perform 2D segmentation from rendered features using SAM, as described in Sec.~\ref{sec:method_clip}, which we compare to 2D segmentation by LERF~\citep{kerr2023lerf} and LangSplat~\citep{qin2023langsplat} (top of table).
For (ii) we directly segment in 3D by binarizing the diffused features before rendering them, then computing 2D segmentation masks from these rendered features. We use automatic thresholding for binarization before and after rendering. 

\myparagraph{Results} 
Table~\ref{tab:lerf-loc} reports object localization results, where our method outperforms LangSplat despite not relying on SAM for this task.
Table~\ref{tab:lerf-seg} presents segmentation results along with average runtimes, which account for feature map generation and 3D feature training when applicable.
Across both evaluation settings, our method consistently surpasses prior approaches while achieving a 5 to 10$\times$ speedup in feature generation and uplifting.

\myparagraph{Runtime analysis}
Our approach based on graph diffusion offers fast feature map generation compared to leveraging SAM’s automatic mask generation in 2D as in \citep{qin2023langsplat,wu2024opengaussian}, but it comes with additional inference-time overhead. This tradeoff is further discussed in supplementary Sec.~\ref{sec:appendix-runtime-comp}. When inference time is a critical constraint, LUDVIG’s uplifting can also be applied directly to full segmentation masks, following the same setup as LangSplat~\citep{qin2023langsplat} and OpenGaussian~\citep{wu2024opengaussian}, as shown in the next section.

\begin{table}[t!]
\centering
\scalebox{0.88}{
\footnotesize
\begin{tabular}{cccccccc}
\toprule
SAM & Graph diffusion & ramen & figurines & teatime & waldo & overall \\ 
\midrule
\xmark & \xmark  & 27.8 & 37.8 & 38.2 & 30.4 & 33.6 \\ 
\xmark & \cmark   & 42.3 & 58.0 & 58.6 & 42.9 & 50.4 \\ 
\cmark & \xmark  & 52.2 & 51.8 & 68.9 & 56.4 & 57.3 \\ 
\cmark & \cmark  & \textbf{58.1} & \textbf{63.3} & \textbf{77.1} & \textbf{58.5} & \textbf{64.3} \\ 
\bottomrule
\end{tabular}
}
\caption{\textbf{Ablation on LERF object segmentation.} Results (IoU) with and without using 3D graph diffusion and/or 2D SAM segmentation, evaluated on the dataset introduced by LangSplat~\citep{qin2023langsplat}.}
\label{tab:ablation-lerf}
\end{table}

\myparagraph{Ablation study}
Table~\ref{tab:ablation-lerf} presents an ablation study on the effect of graph diffusion and SAM-based segmentation following the first evaluation protocol (top part) in Table~\ref{tab:lerf-seg}. 

\subsection{Open-vocabulary semantic segmentation}
\label{sec:exp-semantic}

\myparagraph{Task}
We evaluate on ScanNetv2~\citep{dai2017scannet}, which provides posed RGB images from video scans, reconstructed point clouds, and ground-truth 3D point-level semantic labels.

\myparagraph{Setup}
We follow the evaluation protocol of OpenGaussian~\citep{wu2024opengaussian}.  
In particular, we train Gaussian Splatting by initializing from the ground-truth point cloud, keeping the 3D positions fixed and disabling densification.
For this task, we replace OpenGaussian’s quantized feature learning with our uplifting process, leaving the rest of the protocol unaltered. Specifically, we uplift 2D segmentation masks generated by SAM based on textual queries, as in the prior works we compare to \citep{qin2023langsplat,shi2024legaussians, wu2024opengaussian}.

\myparagraph{Results} 
As shown in Table~\ref{tab:scannet},
our approach yields significant accuracy gains over OpenGaussian~\citep{wu2024opengaussian}, while being considerably simpler and faster than their quantization-based learning process.
\begin{table}[t]
\centering
\footnotesize
\scalebox{.95}{
\begin{tabular}{lcccccc}
\toprule
\multirow{2}{*}{Methods} & \multicolumn{2}{c}{19 classes} & \multicolumn{2}{c}{15 classes} & \multicolumn{2}{c}{10 classes} \\
\cmidrule(lr){2-3} \cmidrule(lr){4-5} \cmidrule(lr){6-7}
 & mIoU & mAcc & mIoU & mAcc & mIoU & mAcc \\
\midrule
LangSplat$^*$      & 3.8  & 9.1  & 5.4  & 13.2 & 8.4  & 22.1 \\
LEGaussians$^*$ & 3.8  & 10.9 & 9.0  & 22.2 & 12.8 & 28.6 \\
OpenGaussian                   & 24.7 & 41.5 & 30.1 & 48.3 & 38.3 & 55.2 \\
LUDVIG  &  \textbf{33.9}  & \textbf{51.4}   &  \textbf{37.4}  & \textbf{57.2}   &  \textbf{46.4}  & \textbf{66.2}   \\
\bottomrule
\end{tabular}
}
\caption{\textbf{ScanNet semantic segmentation.} For LUDVIG, we replace OpenGaussian's quantization-based feature training stage ($\sim40$min) with our direct uplifting process ($\sim3$min), 
starting from the same 2D feature maps ($\sim50$min) 
and GS pre-training stage ($\sim9$min). \footnotesize{$^*$Results reported by OpenGaussian.}}
\label{tab:scannet}
\end{table}
\section{Concluding remarks and limitations}

In this work, we introduce a simple yet effective aggregation mechanism for transferring 2D visual representations into 3D, bypassing traditional optimization-based approaches. The aggregation builds upon already trained Gaussian Splatting representations and is implemented within the CUDA rendering process, making 2D-to-3D uplifting as fast as 3D-to-2D rendering. 
Our approach can serve as an off-the-shelf module for transferring 2D features into 3D across a wide range of applications, as demonstrated in our experiments (Sec.~\ref{sec:exp-semantic}) and its recent integration into Panst3R~\cite{zust2025panst3r} for novel-view panoptic segmentation.

We also propose a graph diffusion process that combines spatial structure with DINOv2 similarity to generate accurate 3D segmentation masks, starting from coarse inputs such as scribbles or CLIP relevancy maps. This leads to strong gains over SAM-based open-vocabulary segmentation baselines while remaining computationally efficient.

However, the quality of the resulting 3D features depends on the underlying 3D scene reconstruction, which remains challenging in cases of high specularity~\citep{jiang2024gaussianshader,yang2024spec} or motion blur~\citep{zhao2024badgaussians,lee2024deblurring}. 
In such scenarios, learning 3D features \emph{along with} 3D Gaussian Splatting reconstruction may serve as a regularization and improve scene geometry.

\subsubsection*{Acknowledgments}
This project was supported by ANR 3IA MIAI@Grenoble Alpes (ANR-19-P3IA-0003) and by ERC grant number 101087696 (APHELEIA project). This work was granted access to the HPC resources of IDRIS under the allocation [AD011013343R2] made by GENCI.

{
    \small
    \bibliographystyle{ieeenat_fullname}
    \bibliography{main}
}

\renewcommand{\thefigure}{\Alph{figure}}
\renewcommand{\thetable}{\Alph{table}}

\clearpage
\appendix

\setcounter{figure}{0}    
\setcounter{table}{0}    

{\Large{Supplementary Material}}
\etocdepthtag.toc{mtappendix}
\etocsettagdepth{mtappendix}{subsection}
\etocsettagdepth{mtchapter}{none}

{
  \hypersetup{linkcolor=ForestGreen}
  \tableofcontents
}

\section{Using LUDVIG for downstream tasks}
\label{sec:appendix-method_seg}

In this section, we describe our approach for uplifting DINOv2, SAM and CLIP models and evaluating the 3D features on two downstream tasks: segmentation and open-vocabulary object detection.
As in Sec.~\ref{sec:method_uplifting}, we are given a set of 2D  frames $I_1,\dots, I_m$, with viewing directions $d_1,\dots, d_m$ and a corresponding 3D scene obtained using the Gaussian Splatting method.  

\myparagraph{Multi-view segmentation} For this task, we assume that a \textit{foreground mask} of the object to be segmented is provided on a \emph{reference frame} taken to be the first frame $I_1$.
We consider two types of foreground masks: either \emph{scribbles} or a whole \emph{reference mask} of the object, both of which define a set of \textit{foreground pixels} $\mathcal{P}$. 
In the following subsections, we present the proposed approaches for segmentation using SAM and DINOv2 features, based on both types of foreground masks.

\subsection{Multiple-view segmentation with SAM}\label{sec:appendix-sam_seg}

SAM~\citep{kirillov2023sam, ravi2024sam2} is a powerful image segmentation model, that can generate object segmentation masks from point prompts on a single 2D image.
Aggregating SAM 2D segmentation masks in 3D allows for cross-view consistency and improves single-view segmentation results.
In order to leverage SAM, we propose a simple mechanism for generating SAM 2D features for each frame from a \textit{foreground mask} in the \textit{reference frame}.

\myparagraph{Generating 2D query points for SAM}
The key idea is to generate point prompts on each frame from the  \textit{foreground mask} provided on the \textit{reference frame}. To this end, we perform an uplifting of the \textit{foreground mask}  (Eq.~(\ref{eq:uplifting_matrix}) from the main paper) and re-project it on all frames (Eq.~(\ref{eq:rendering}) from the main paper).
From the reprojected mask for viewing direction $d$, further normalized by its average value, we retain a subset $\mathcal{P}_d$ of pixels with values higher than a threshold $\tau$ fixed for all scenes ($\tau=0.4$ for SPIn-NeRF and $\tau=1$ for NVOS).  
We then predict a SAM mask based on these point prompts as described below.

\myparagraph{Predicting SAM masks from sets of query points}
Given a set of pixels $\mathcal{P}_d$ pertaining to the foreground, we compute 2D segmentation masks using SAM by randomly selecting $3$ points prompts from $\mathcal{P}_d$, repeating the operation $10$ times and averaging the resulting masks for each view to obtain the final 2D SAM feature maps. 

\myparagraph{Segmentation with uplifted SAM masks} The 2D segmentation masks generated by SAM are uplifted using the aggregation scheme described in Sec.~\ref{sec:uplifting}. Our final prediction is obtained by rendering the uplifted feature maps into the target frame and thresholding.

\myparagraph{Evaluation and hyperparameter tuning}
Segmentation with 3D SAM masks requires setting a threshold for foreground/background pixel assignment, and optionally choosing one of the three masks proposed by SAM (representing different possible segmentations of the object of interest).
For SPIn-NeRF, the threshold and mask prediction are chosen based on the IoU for the available reference mask. For NVOS, only reference scribbles are provided; hence, a single mask is predicted, and the segmentation threshold is fixed across all scenes for SAM, and automatically chosen using Li's iterative Minimum Cross Entropy method~\citep{li1993minimum} for SAM 2.

\subsection{{Multiple-view} segmentation with DINOv2}\label{sec:appendix-dino_seg}

DINOv2~\citep{oquab2024dinov2} is a self-supervised vision model recognized for its generalization capabilities. 
In this work, we aggregate the patch-level representations produced by DINOv2 with registers~\citep{darcet2024registers} into a high resolution and fine-grained 3D semantic representation.  

\begin{figure*}[t!]
    \centering
    \includegraphics[width=\linewidth]{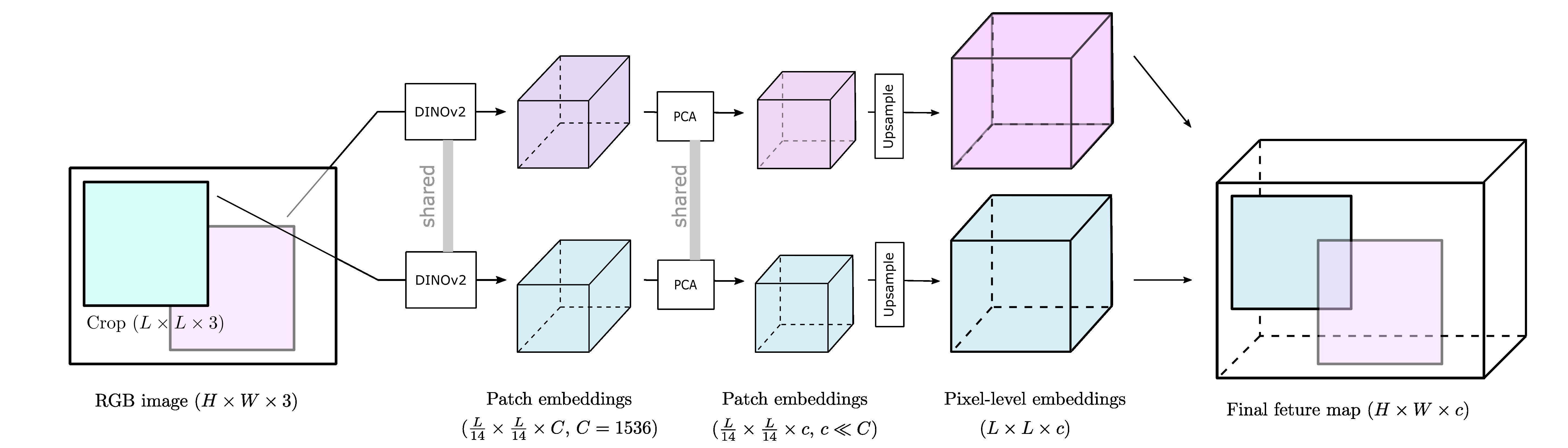}
    \caption{Sliding windows for construction of DINOv2 feature maps.}
    \label{fig:appendix-sliding-windows}
\end{figure*}

\myparagraph{Construction of 2D feature maps}
\label{sec:slidingwindows}
We construct the 2D feature maps using a combination of a sliding windows mechanism and dimensionality reduction of the original DINOv2 features. 
Specifically, we i) extract DINOv2 patch-level representations across multiple overlapping crops of each image, ii) apply dimensionality reduction over the set of all patch embeddings, ii) upsample and aggregate the dimensionality-reduced patch embeddings to obtain pixel-level features for each image. The process is illustrated in Fig.~\ref{fig:appendix-sliding-windows}. 
This approach enhances the granularity of spatial representations by aggregating patch-level representations to form pixel-level features. To favor the first principal components, known to focus on the foreground objects~\citep{oquab2024dinov2}, the features are re-weighted by the eigenvalues of the PCA decomposition.

\myparagraph{Segmentation with uplifted DINOv2 features}
The 2D feature maps from the $m$ views are uplifted using Eq.~(\ref{eq:uplifting_matrix}) from the main paper, and the resulting 3D features are re-projected into any viewing direction $d$ using Eq.~(\ref{eq:rendering}) from the main paper to compute rendered 2D features $(\hat{F}_{d,p})$.
To obtain segmentation masks, we construct a score $P(\hat{F}_{d,p})$ for a 2D pixel $p$ to belong to the foreground, based on its corresponding rendered feature. More precisely, $P$ relies on the rendered \textit{foreground features} $\mathcal{F}_{\text{ref}} := (\hat{F}_{d_1,p})_{p\in \mathcal{P}}$  corresponding to the \textit{foreground mask} computed on the \textit{reference frame} $I_1$. 
We propose two approaches for constructing $P$. 
The first one is a simple approach that sets $P(\hat{F}_{d,p}) = \mathcal{S}_F(\hat{F}_{d,p}, \bar{F})$ where $\bar{F}$ is the average over foreground features $\mathcal{F}_{\text{ref}}$, and $\mathcal{S}_F$ is defined based on the cosine similarity. 
The second approach is more discriminative and first trains a logistic regression model $P$ on all rendered 2D features of the reference frame, so that the foreground features $\mathcal{F}_{\text{ref}}$  are assigned a positive label. Then $P(\hat{F}_{d,p})$ gives the probability that a pixel $p$ belongs to the foreground. The final mask is then obtained by thresholding.

Experimentally, the second approach is extremely efficient when the set of \textit{foreground pixels}  $\mathcal{P}$ covers the whole object to segment, so that $P$ captures all relevant features. This is the case when a whole \textit{reference mask} of the object is provided. When the \textit{foreground pixels}  $\mathcal{P}$ does not cover the whole object, as with scribbles, $P$ can be discriminative to parts of the object that are not covered by $\mathcal{P}$. Therefore, we rely on the second approach for tasks where a reference mask is provided, and use the simpler first approach when only scribbles serve as reference. 

\subsection{Enhancing segmentation with DINOv2 using 3D graph diffusion}
\label{sec:appendix-diffusion}

DINOv2 provides generic visual features that do not explicitly include information for segmentation, unlike models such as SAM that were specifically trained for such a task. 
Consequently, using the 2D projections of uplifted DINOv2 features, as proposed in Sec. \ref{sec:appendix-dino_seg}, might fail to separate different objects that happen to have similar features while still being distinct entities. 
This challenge can be mitigated by incorporating 3D spatial information in which the objects are more likely to be well-separated. 
To this end, we propose to leverage the graph diffusion process introduced in Sec.~\ref{sec:diffusion}. 
Below, we describe the initialization of the weight vector $g_0$ and the construction of the adjacency matrix $A$.  

\myparagraph{Initialization of the weight vector} 
The initial vector of weights $g_0\in \mathbb{R}^n$ representing a coarse estimation of the contribution of each Gaussian to the segmentation mask. It is computed by uplifting the 2D \textit{foreground mask} (either scribbles or a reference mask) from the \textit{reference frame} using Eq.~(\ref{eq:uplifting}) from the main paper, and retaining only the top 10\% of Gaussians with positive mask values, setting the rest to zero.
The nodes for which $g_0$ has a positive value define a set of anchor nodes $\mathcal{M}$ that are more likely to contribute to the foreground. The resulting weight vector is a coarse estimation of how much each Gaussian contributes to a rendered 2D segmentation mask.

\myparagraph{Construction of the graph edges}
We define the pairwise similarity function $S_f$ as:
\begin{equation}
    \label{eq:appendix-rbf}
    S_f(f_i, f_j) = \exp\left ( - \frac{||f_i - f_j||_2^2}{b s_f^2} \right)
\end{equation}
 where $f_i, f_j$ are the $l_2$-normalized DINOv2 features, $s_f$ is the median of pairwise $l_2$ distances and $b$ is a bandwidth parameter.
 We choose a global unary regularization term $P(f_i)$ on each node $i$ to contain diffusion to nodes with features similar to those of the foreground. More precisely, $P$ is defined using a similar approach as in Sec.~\ref{sec:appendix-dino_seg}: \begin{enumerate}
     \item When scribblers are provided, $P(f_i) = \mathcal{S}_{f}(f_i, \bar{f})$ with the averaged feature $\bar{f}$ over the anchor nodes $\mathcal{M}$, and a different value for the bandwidth $b$. 
     \item When a full foreground mask is available, we train a logistic regression model on the uplifted features with positive labels for the anchor nodes' features. The unary term is then defined as $P(f_i) = \mathcal{L}(f_i)^{1/b}$, with $b$ a bandwidth parameter and $\mathcal{L}(f_i)$ is the model's predicted probability for $f_i$.
 \end{enumerate}

The local term $S_f$ allows diffusing to neighbors that have similar features while the unary term prevents leakage to background nodes and allows using an arbitrary number of diffusion steps. 

The matrix $A$ of graph edges is then defined based on $S_f$ and $P$ as in Eq.~(\ref{eq:edges}) from the main paper.
For this task, we binarize $A$ with a fixed threshold (set to $10^{-5}$). After the $T$ diffusion steps, we recover the  nodes $\mathcal{S}$ in $g_T$ with strictly positive values (\emph{i.e.}, those reachable after $T$ iterations). The final weight is defined as $h_i = P(f_i)$ if $i\in \mathcal{S}$ and $0$ otherwise. Segmentation is then performed by projecting $\mathbf{h} = (h_i)$ into 2D and thresholding. The selection process of the threshold and bandwidth parameters is detailed below.

\myparagraph{Evaluation and hyperparameter tuning}
Segmentation relies on three hyperparameters: the bandwidths for $S_f$ and $P$, and the threshold for foreground/background pixel assignment. 
For SPIn-NeRF, all hyperparameters are chosen based on the IoU for the available reference mask. For NVOS, only reference scribbles are provided, hence we predict a SAM mask based on the scribbles of the reference frame, and choose the hyperparameters maximizing the IoU with this SAM mask. This is consistent with a scenario where the user, here SAM, would choose hyperparameters based on visual inspection on one of the frames.

\subsection{Open-vocabulary object localization}
\label{sec:appendix-clip}

For the open-vocabulary object localization task, multi-resolution CLIP feature maps are constructed as described in Sec.~\ref{sec:method_clip} 
and uplifted along with DINOv2 features using Eq.~(\ref{eq:uplifting}) from the main paper.
The refined 3D CLIP features are then evaluated on the LERF localization and segmentation tasks as described below.

\subsubsection{Relevancy scores and object localization.}

We consider uplifted 3D CLIP features $f$.
We follow LERF~\citep{kerr2023lerf} and LangSplat~\citep{qin2023langsplat} to compute alignment scores between CLIP visual features and a text query, denoted as \emph{relevancy score}, and for object localization based on these relevancy scores.

\myparagraph{Relevancy scores}
The relevancy $r_{i,q}$ between a feature $f_i$ and text query $\phi_{\text{q}}$ is defined as follows:
\begin{equation}
r_{i,q} = \min_k \frac{\exp(T \cdot f_i \cdot \phi_{q})}{\exp(T \cdot f_i \cdot \phi_{q}) + \exp(T \cdot f_i \cdot \phi_{cano}^k)},
\end{equation}
where $T$ is a temperature parameter set to 10 by~\cite{kerr2023lerf} and $\phi_{\text{cano}}^k$ is the CLIP embedding of a predefined canonical phrase chosen from “object," “things," “stuff," and “texture."
Note that \cite{qin2023langsplat} compute the relevancy scores for 2D pixels, while we directly compute them for 3D Gaussians, allowing their manipulation in 3D.

\myparagraph{Localization}
The 3D CLIP relevancies can be projected into 2D for a given camera pose, and used for localization and segmentation for each text query. For localization, we follow \cite{qin2023langsplat} and choose the pixel with the highest relevancy score, following a 2D smoothing using a mean filter with kernel size $K=5$ in our work. 

\subsubsection{Object segmentation}

Segmentation based on raw CLIP relevancies is challenging, as fully covering the object of interest without capturing other objects of a similar nature is challenging.

We first describe our process for segmenting directly based 3D relevancies. We then present two complementary approaches that allow for a more targeted segmentation: predicting 2D SAM masks by retrieving query points with high relevancies, and refining 3D relevancy scores using graph diffusion based on 3D DINOv2 features.

\myparagraph{Segmentation from 3D relevancies}
Given a camera pose and a text query, a segmentation mask is obtained by first applying a rough thresholding over projected relevancies rescaled by their maximum value, with a fixed threshold value $\tau=0.8$, followed by automatic thresholding with Otsu's method \citep{otsu1975threshold}.

\myparagraph{Improving 2D segmentation with SAM} 
For segmentation with SAM, we use the pixels with the highest relevancy scores as query points for a given camera pose and text query. 
More precisely, we first obtain a mask $\mathcal{M}$ by projecting and thresholding the relevancies as described above, using $\tau=0.93$. We then use the approach described in Sec.~\ref{sec:appendix-sam_seg}, paragraph \emph{Predicting a SAM segmentation mask from a set of query points} and average 20 mask predictions.
We choose the top-$q$ percent of pixels as the subset of query points for SAM, where $q$ is the proportion of positive pixels in $\mathcal{M}$, hence extracting a larger subset of point prompts for larger objects. The scalar map obtained by averaging the 20 predicted masks is then automatically thresholded again using Otsu's method~\citep{otsu1975threshold}.

\myparagraph{Refining 3D CLIP relevancies with graph diffusion based on DINOv2 features}
We refine 3D CLIP relevancy scores using graph diffusion based on 3D DINOv2 features ($f$), as in Sec.~\ref{sec:appendix-diffusion}. The diffusion process runs in parallel for all text queries. For initialization, we keep positive a very restricted set of nodes with the highest relevancy, whose weight propagate to neighboring nodes with similar DINOv2 features, progressively spanning the object of interest. The diffusion process results in a better segmentation both with and without leveraging SAM. When using SAM, the set of query pixels can have a larger coverage of the object of interest without extending to other objects.

\myparagraph{Details on graph construction and node initialization for refining 3D CLIP relevancies}
The pairwise similarity function $S_f$ is defined as in Eq.~\ref{eq:appendix-rbf} with a bandwidth value $b=0.5$. For each text query $\phi_q$, we define a unary regularization term $P_q$ using a logistic regression model $\mathcal{L}_q$ that predicts the probability that a DINOv2 feature $f_i$ belongs to the object corresponding to query $\phi_q$. The set of nodes $\mathcal{P}$ with positive labels is defined based on 3D CLIP relevancy scores for prompt $\phi_q$. More precisely, we rescale 3D relevancies to $[0,1]$ and apply Otsu's method~\citep{otsu1975threshold} over relevancies above $0.5$. We use a regularization $C=0.001$ and equal class weighting for training the model.
The unary regularization term $P_q$ is then defined as $P_q(f_i) = \mathcal{L}_q(f_i)^{1/b}$, with $b=0.025$ for segmentation with SAM, and $b=2$ otherwise.
The initial weight vector $g_0$ is defined by applying two more iterations of Otsu's method among nodes in $\mathcal{P}$ and setting to zero all relevancies below the given threshold. Restricting the set of initial points ensures diffusion only happens within the object of interest. Segmentation based on the resulting 3D relevancy scores is then performed as described in the previous paragraphs, using $\tau=0.01$ for segmentation with SAM and $\tau=0$ otherwise. 

\begin{table*}[t!]
\centering
\footnotesize
\scalebox{1.}{
\begin{tabular}{lccccccccccccc}
\toprule
& \multicolumn{2}{c}{\textbf{Images (\#)}} &  \multicolumn{2}{c}{\textbf{Text queries (\#)}} & \multicolumn{2}{c}{\textbf{DINOv2 (s)}} & \textbf{CLIP (s)} & \multicolumn{2}{c}{\textbf{Graph diffusion (s)}} & \multicolumn{2}{c}{\textbf{2D segmentation (s)}} & Total \\
\cmidrule(lr){2-3} \cmidrule(lr){4-5} \cmidrule(lr){6-7} \cmidrule(lr){8-8} \cmidrule(lr){9-10} \cmidrule(lr){11-12}
Scene & Train & Test & Unique & Total & Gen. & Up. & Gen.+Up. &  Scene & Prompt & w/ SAM & w/o SAM & (mins) \\
\midrule
Teatime   & 177 & 6 & 14 & 59  & 45 & 14 & 363 & 42 & 15 & 9 & 0.9 & 8 \\
Waldo     & 187 & 5 & 18 & 22  & 44 & 18 & 371 & 39 & 19 & 4 & 0.5 & 8 \\
Ramen     & 131 & 7 & 14 & 71 & 40 & 9 & 227 & 37 & 14 & 11 & 1 & 6  \\
Figurines & 299 & 4 & 21 & 56  & 58 & 38 & 811 & 45 & 22  & 8 & 0.8 & 16  \\
\bottomrule
\end{tabular}
}
\caption{\textbf{Runtimes for evaluation on LERF Segmentation~\citep{kerr2023lerf, qin2023langsplat}.} 
The last column (Total) reports total time, which breaks down between i) feature map generation (Gen.) and uplifting (Up.) for all training images; ii) graph diffusion, divided between scene-specific (querying neighbors, defining $S_f$) and prompt-specific (defining $P$, running diffusion) operations for all text queries; iii) 2D segmentation with/without SAM for all text queries across test images. We also report the number of training and test images and the number of text queries across test images.}
\label{tab:runtimes}
\end{table*}
\begin{table*}[t!]
\centering
\footnotesize

\begin{subtable}{\linewidth}
\centering
\scalebox{.99}{
\begin{tabular}{lcccccc}
\toprule
 & \multirow{2}{*}{SA-TensoRF} & \multirow{2}{*}{SA-GS} & \multirow{2}{*}{OmniSeg3D} & \multirow{2}{*}{SAGA} & \multicolumn{2}{c}{LUDVIG} \\
 & & & & & SAM & DINOv2 \\
\midrule
2D feature extraction (sec.) & 30 & 30 & 900\footnotemark & 900$\footnotemark[1]$ & 20 & 30 \\
3D feature training (sec.) & \emph{15} & \emph{4} & 900 & 1500 & \emph{4} & 40+\emph{2} \\
\bottomrule
\end{tabular}
}
\caption{\textbf{Multi-view segmentation (NVOS)}. \emph{Italic} denote inference-time operations (after 2D scribbles are given). OmniSeg3D and SAGA avoid inference overhead but require longer feature extraction and training\vspace{.1cm}.}
\end{subtable}
\begin{subtable}{\linewidth}
\centering
\scalebox{.99}{
\begin{tabular}{lcccccc}
\toprule
 & \multirow{2}{*}{LERF} & \multirow{2}{*}{LangSplat} & \multirow{2}{*}{OpenGaussian} & \multicolumn{2}{c}{LUDVIG} \\
& & & & w/o diff. & w/ diff. \\
\midrule
2D feature extraction (min) & - & ~$50$\footnotemark[1] & ~$50$\footnotemark[1] & 4.5 & 4.5 \\
3D feature training (min) & 25 & 25 & 40 & 2 & 2.5 \\
Inference (s/query) & 30.9 & 0.26 & $\sim0.2$~~~ & 0.3 & 1.3 \\
\bottomrule
\end{tabular}
}
\caption{\textbf{Open-vocabulary segmentation (Ramen scene) \vspace{-.2cm}.} 
}
\end{subtable}
\vspace{-.15cm}
\caption{\textbf{Training and inference times.} All methods build on top of a GS pretraining stage ($\sim 10$min). Reported values are paper-sourced and indicative, as hardware setups may differ.
}
\vspace{-.2cm}
\label{tab:rebuttal-times}
\end{table*}

\section{Runtime analysis}

\subsection{Runtimes for LUDVIG}
\label{sec:appendix-runtime}

In this section, we detail our running times for feature uplifting and evaluation, conducted on a GPU RTX 6000 ADA. Table~\ref{tab:runtimes} shows a breakdown of running times between feature uplifting (Up.) and generation (Gen.), graph diffusion and 2D segmentation for evaluation on LERF segmentation. 
The total reported times can be divided between pre-uplifting, uplifting and post-uplifting. In our experiments, the pre-processing and uplifting steps are independent from the downstream tasks (except for our foreground/background segmentation with SAM), and part of the graph-diffusion process is task-dependent. Below we detail our runtimes for every step and compare them to the literature.

\myparagraph{Pre-uplifting: feature map generation} The time this step takes (Gen. in Table~\ref{tab:runtimes}) depends on the backbone model, on the number of images and on the number of calls to the model per image. The total time ranges from a few seconds up to an hour for approaches such as LangSplat \citep{qin2023langsplat}, that queries SAM over a grid of points on the image at various resolutions to generate full image segmentation masks.\footnote{This process takes 24s/image on a GPU 6000 ADA and amounts to an average of 80 minutes for the evaluated scenes.} In our experiments, the feature generation step takes from 1 to 5 minutes, except for Sec.~\ref{sec:exp-semantic} where we uplift semantic maps generated with LangSplat's approach.

\myparagraph{Uplifting} For LUDVIG, uplifting time is linear in the number of images (given a 3D scene representation). Experimentally, it is also linear in the number of feature dimensions, taking 2ms per dimension for an image of size $724 \times 986$. As reported in Table~\ref{tab:runtimes} (Up.), uplifting 100 images of dimension $40$ takes 9s on average. By contrast, gradient-based optimization requires approximately $n_{\text{steps}}$ times this duration, where the number of gradient steps $n_{\text{steps}}$ typically ranges from 3,000 to 30,000 for 3D feature distillation~\citep{kerr2023lerf, qin2023langsplat, zuo2024fmgs}. Gradient-based optimization can still be very fast for low-dimensional features such as SAM masks (can take as little as a few seconds, as reported by SA3D-GS~\citep{cen2023segmentGS}) or dimensionality-reduced features (LangSplat~\citep{qin2023langsplat} trains an autoencoder to reduce the CLIP feature dimension from 512 to 3 and runs for 25 minutes). However, optimization becomes intractable for high-dimensional features such as CLIP and DINO; FMGS~\citep{zuo2024fmgs} relies on an efficient multi-resolution hash embedding of the scene; however, their total training time still amounts to 1.4 hours.

\myparagraph{Post-uplifting: graph diffusion} After uplifting, LUDVIG performs graph diffusion using pairwise DINOv2 feature similarities for the segmentation tasks in Sec.~\ref{sec:exp-seg} and \ref{sec:exp-clip}. In Table~\ref{tab:runtimes}, we divide runtimes in two categories: 
\begin{itemize}
    \item \textbf{Scene}: operations performed once for the whole scene. This includes querying the Euclidean neighbors for each node, which is log-linear in the number of Gaussians. With 600,000 Gaussians as in our experiments, the step takes about 30s, and can be further optimized by using approximate nearest neighbor search algorithms~\citep{wang2021comprehensive}. Defining $S_f$ based on DINOv2 features is also independent from the downstream task.
    \item \textbf{Prompt}: operations that are specific to the downstream task. This includes defining the regularization $P$ (\emph{e.g.} training logistic regression model(s)) and running the diffusion. The time taken depends on dimension of the diffused features (\emph{e.g.} number of text queries): 1 to 2 seconds for foreground/background segmentation (a single mask) and 18 seconds on average for LERF segmentation (14 to 21 text queries).
\end{itemize}

\myparagraph{Post-uplifting: segmentation} Our evaluation on LERF involves 2D segmentation with SAM based on 3D relevancy scores. The runtime depends on the number of test images and on the total number of 2D text queries, as it involves one call to the SAM backbone per test image, and multiple calls to the SAM prediction head per text query, as detailed in Appendix~\ref{sec:appendix-clip}. Our total inference time per scene is of $8 s$ on average, against $0.8 s$ when not using SAM. In contrast, Langsplat does not rely on SAM at inference time, but relies on a computationally expensive feature map generation process, with more than 1 hour runtime.

\subsection{Runtime comparisons}
\label{sec:appendix-runtime-comp}

Table~\ref{tab:rebuttal-times} reports runtime comparisons for the multi-view~(Sec.~\ref{sec:exp-seg}) and open-vocabulary (Sec.~\ref{sec:exp-clip}) segmentation tasks.
Compared to approaches relying on SAM's automatic mask generation, such as OmniSeg3D~\citep{ying2023omniseg3d}, SAGA~\citep{cen2023saga}, LangSplat~\citep{qin2023langsplat}, and OpenGaussian~\citep{wu2024opengaussian}, our method offers significantly faster feature generation and uplifting, at the cost of a slightly higher inference time.
In particular, the graph diffusion step adds approximately 1~second per query at inference (reported as 14~seconds for Ramen’s 14 queries in supplementary Table~\ref{tab:runtimes}).

Our experiments in Sec.~\ref{sec:exp-semantic} show that our uplifting process can serve as a drop-in replacement for the training phase in methods optimized for fast inference, such as OpenGaussian~\citep{wu2024opengaussian}, which learns from 2D feature maps generated using LangSplat’s approach. In these experiments, we adopt the same 2D feature map generation and evaluation protocols as OpenGaussian, replacing only their quantization-based training phase with our fast feature aggregation. This results in significantly faster uplifting while achieving notable accuracy gains.

\begin{table*}[t!]
  \centering
  \footnotesize
\begin{tabular}{lccccccc}
\toprule
& \multirow{1}{*}{MVSeg} & \multirow{1}{*}{SA3D-GS} & \multirow{1}{*}{SAGA} & \multirow{1}{*}{OmniSeg3D} & \multicolumn{3}{c}{LUDVIG (Ours)}  \\
\cmidrule(lr){6-8}
3D representation: & NeRF & GS & GS & NeRF &  \multicolumn{3}{c}{GS} \\
Uplifting: &  & SAM & SAM & SAM & DINOv2 & SAM & SAM2 \\
\midrule
Orchids      & 92.7 & 84.7 & - & 92.3 & 92.6 & 92.2 & 91.0 \\
Leaves  & 94.9 & 97.2  & - & 96.0  & 96.2 & 96.3 & 96.3 \\
Fern & 94.3 & 96.7 & - & 97.5 & 96.3 & 97.0 & 97.0 \\
Room   & 95.6 & 93.7 & - & 97.9 & 95.7 & 96.5 & 96.1 \\
Horns  & 92.8 & 95.3 & - & 91.5 & 95.1 & 94.5 & 94.8 \\
Fortress   & 97.7 & 98.1 & - & 97.9  & 97.5 & 98.3 & 98.3 \\
\midrule
Fork   & 87.9 & 87.9 & - & 90.4 & 85.0 & 86.8 & 86.7 \\
Pinecone   & 93.4 & 91.6 & - & 92.1 & 93.2 & 88.8 & 90.7  \\
Truck   & 85.2 & 94.8 & - & 96.1 & 95.6 & 94.9 & 93.9 \\
Lego   & 74.9 & 92.0 & - & 90.8 & 91.1 & 92.7 & 92.9 \\
\midrule
Average  & 90.9 &  93.2  & 93.4 & 94.3 & 93.8 & 93.8 & 93.8 \\
\hline
\end{tabular}
  \caption{Segmentation (IoU) on SPIn-NeRF~\citep{mirzaei2023spin} with DINOv2, SAM and SAM2.}
  \label{tab:spin}
\end{table*}
\begin{table*}[t!]
  \centering
  \footnotesize
\begin{tabular}{lccccccccc}
\toprule
& Fern & Flower & Fortress & HornsC & HornsL & Leaves & Orchids & Trex & Average \\
\midrule
NVOS & - & - & - & - & - & - & - & - & 70.1  \\
SA3D & 82.9 & 94.6 & 98.3 & 96.2 & 90.2 & 93.2 & 85.5 & 82.0 & 90.3  \\
OmniSeg3D & 82.7 & 95.3 & 98.5 & 97.7 & 95.6 & 92.7 & 84.0 & 87.4 & 91.7  \\
SA3D-GS & - & - & - & - & - & - & - & - & 92.2  \\
SAGA & - & - & - & - & - & - & - & - & 92.6  \\
Ours-DINOv2 & 84.5 & 95.6 & 97.5 & 97.3 & 93.4 &  96.3 & 91.7 & 84.7 & 92.4    \\
Ours-SAM & 85.5 & 97.6 & 98.1 & 97.9 & 94.1 & 96.4 & 73.1 & 88.0  & 91.3    \\
Ours-SAM2 & 84.8 & 97.3 & 98.3 & 97.7 & 93.4 & 96.7 & 73.1 & 89.1 & 91.3    \\
\hline
\end{tabular}
  \caption{Segmentation (IoU) on NVOS~\citep{ren2022nvos} with DINOv2, SAM and SAM2.}
  \label{tab:nvos}
\end{table*}
\begin{table*}[t!] 
\centering \footnotesize 
\begin{tabular}{lcccccc} 
\toprule
& Geometry only & \multicolumn{2}{c}{Single view} & \multicolumn{2}{c}{Uplifting} & \multicolumn{1}{c}{Graph diffusion} \\
\cmidrule(lr){3-4} \cmidrule(lr){5-7}
Model: & Reference mask & DINOv2 & SAM2 & DINOv2 & SAM2 & DINOv2 \\ 
\midrule 
Orchids & 71.3 & 91.5 & 78.4 & 91.5 & 91.0 & 92.6 \\ 
Leaves &  72.4 & 89.3 & 96.6 & 94.1 & 96.3 & 96.2 \\ 
Fern &  93.9 & 95.1 & 96.7 & 96.7 & 97.0 & 96.3 \\ 
Room &  77.4 & 95.4 & 95.6 & 97.3 & 96.1 & 95.7 \\ 
Horns &  80.7 & 90.9 & 93.0 & 94.2 & 94.8 & 95.1 \\ 
Fortress &  94.3 & 96.8 & 97.7 & 98.6 & 98.3 & 97.5 \\ 
\midrule 
Fork &  67.5 & 85.6 & 80.5 & 88.8 &  86.7 & 85.0 \\ 
Pinecone &  56.5 & 92.8 & 67.8 & 89.6 & 90.7 & 93.2 \\ 
Truck &  60.1 & 83.6 & 90.9 & 95.2 & 93.9 & 95.6 \\ 
Lego &  57.3 & 64.4 & 89.0 & 69.9 & 92.9 & 91.1 \\
\midrule 
Average & 73.1 & 88.5 & 88.6 & 91.6 & 93.8 & 93.8 \\
\bottomrule 
\end{tabular} 
\caption{\textbf{Segmentation (IoU) on SPIn-NeRF~\citep{mirzaei2023spin}}. We compare purely geometrical reference mask uplifting and reprojection, single-view prediction, feature/mask uplifting, and graph diffusion leveraging DINOv2 or SAM2.} \label{tab:ablation-spin-dino-sam}
\end{table*}

\section{Additional results}

\subsection{Per-scene multi-view segmentation results}
\label{sec:appendix-seg_scene}

In this section, we present per-scene segmentation results on NVOS and SPIn-NeRF in Tables~\ref{tab:spin},~\ref{tab:nvos} and \ref{tab:ablation-spin-dino-sam}, along with an extended analysis of these results.

\myparagraph{Segmentation on SPIn-NeRF}
We report our segmentation results for the SPin-NeRF dataset~\citep{mirzaei2023spin} in Table~\ref{tab:spin}.
Our results are comparable to the state of the art while not relying on optimization-based approaches. Surprisingly, our segmentation with DINOv2 using graph diffusion also gives results on par with models leveraging SAM masks. 
Our lower segmentation results compared to OmniSeg's can be partly attributed to poor Gaussian Splatting reconstruction of highly specular scenes such as the Fork, in which semi-transparent Gaussians floating over the object try to represent reflections or surface effects that are difficult to capture with standard rasterization techniques~\citep{jiang2024gaussianshader}.

\myparagraph{Segmentation on NVOS}
We report our segmentation results for the NVOS dataset~\citep{ren2022nvos} in Table~\ref{tab:nvos}. 
Our results are comparable to those obtained by prior work. Again, DINOv2 performs surprisingly well while not having been trained on billions of labeled images like SAM. Compared to SAM, DINOv2 better captures complex objects, but sometimes also captures some background noise. This can be seen in Appendix Fig.~\ref{fig:appendix-nvos} with the example of Trex: while SAM misses out the end of the tail as well as the end of the ribs, DINOv2 captures the whole Trex, but also captures part of the stairs behind. Visualisations of Orchids in Appendix Fig.~\ref{fig:appendix-nvos} also explain the lower performance of SAM on this scene: the two orchids SAM is missing are not covered by the positive scribbles, which makes the task ambiguous.

\myparagraph{Ablation study}
In Table \ref{tab:ablation-spin-dino-sam}, we compare our segmentation protocol using DINOv2 and SAM2 to multiple simpler variants. More precisely, we evaluate i) a purely geometrical variant that does not use SAM2 or DINOv2, ii) single-view segmentation in 2D based on SAM2 or DINOv2 2D predictions, iii) uplifting DINOv2 features or SAM2 masks into 3D then rendering them for segmentation, as described in Sec.~\ref{sec:appendix-sam_seg} and \ref{sec:appendix-dino_seg}, and iv) segmenting using graph diffusion over DINOv2 3D feature similarities.

The purely geometrical approach works well on the forward-facing LLFF scenes (Orchids to Fortress). In these scenes, the reference mask is accurately uplifted and reprojected as the viewing direction changes only a little between each frame. However, it fails on the 360-degree scenes (Fork, Pinecone, Truck, Lego). This points to a suboptimal 3D reconstruction of the scene, likely due to overfitting on the limited numbers of available views~\citep{chung2024depth}.

The single-view variants use a similar process for constructing the features and using them for segmentation as in Sec.~\ref{sec:appendix-sam_seg} and \ref{sec:appendix-dino_seg} but without uplifting and rendering. It improves from a purely geometrical approach and performs reasonably well on average, the foreground being well isolated from the rest of the scene. However, as illustrated in Fig.~3, 
the semantic features are at a much lower resolution than those resulting from 3D uplifting, leading to a coarser segmentation.

3D uplifting considerably boosts results compared to single-view approaches. However, performing segmentation in 2D based on the uplifted DINOv2 features does not benefit from the 3D spatial information and typically fails on the 360-degree scenes (Pinecone, Truck and Lego) which have higher variability across different views. Introducing 3D spatial information through 3D graph diffusion results in a boosted performance on these scenes.

\section{Additional visualizations}

\subsection{Segmentation tasks}

\begin{figure*}[t!]
    \centering
    \begin{subfigure}{.24\linewidth}
        \includegraphics[width=\linewidth]{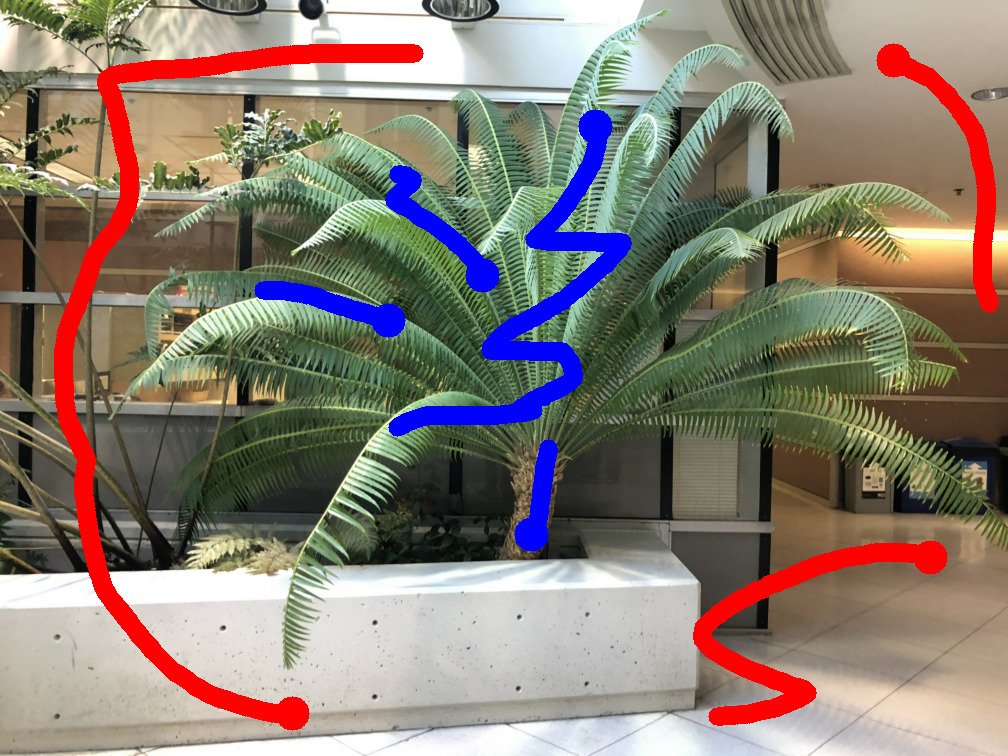}
        \includegraphics[width=\linewidth]{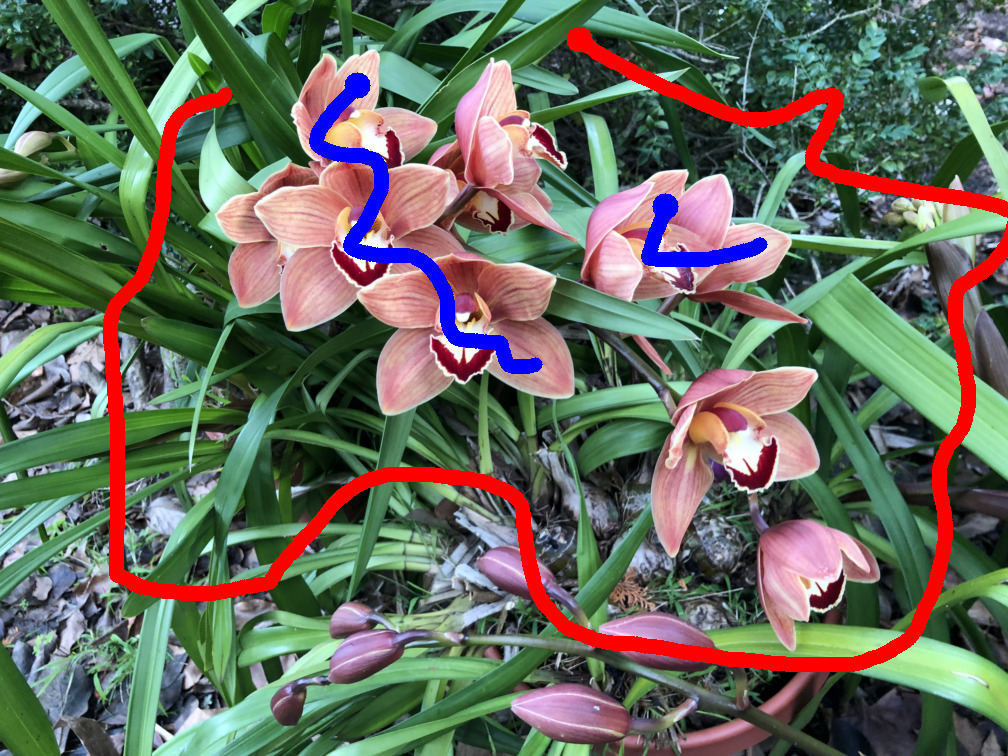}
        \includegraphics[width=\linewidth]{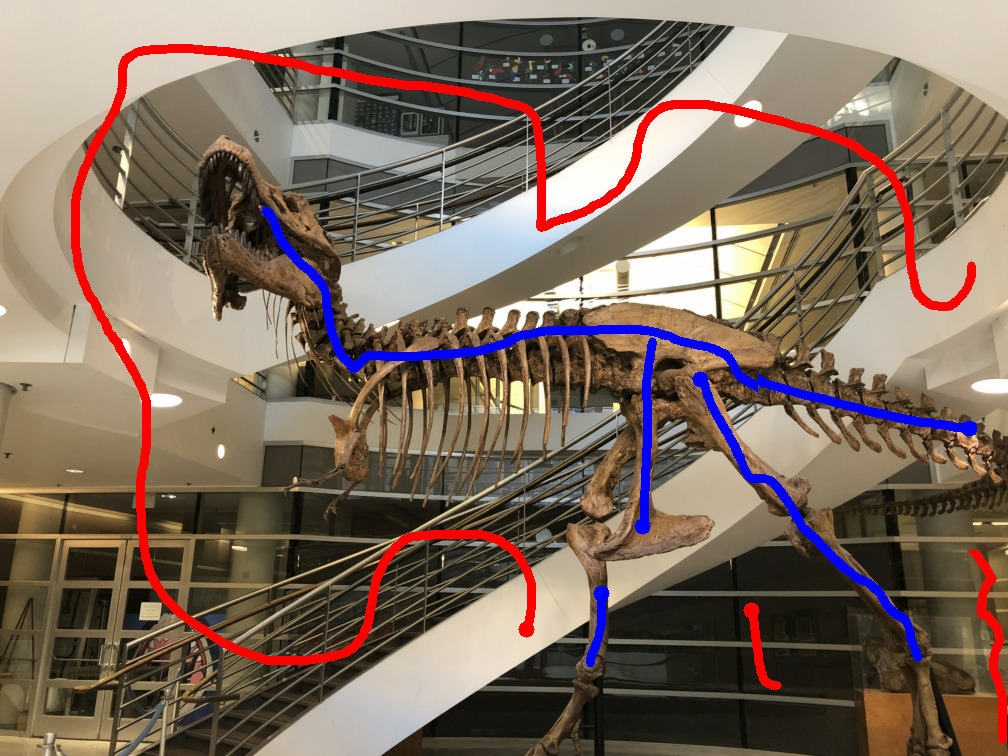}
        \caption{Reference image}
    \end{subfigure}
    \begin{subfigure}{.24\linewidth}
        \includegraphics[width=\linewidth]{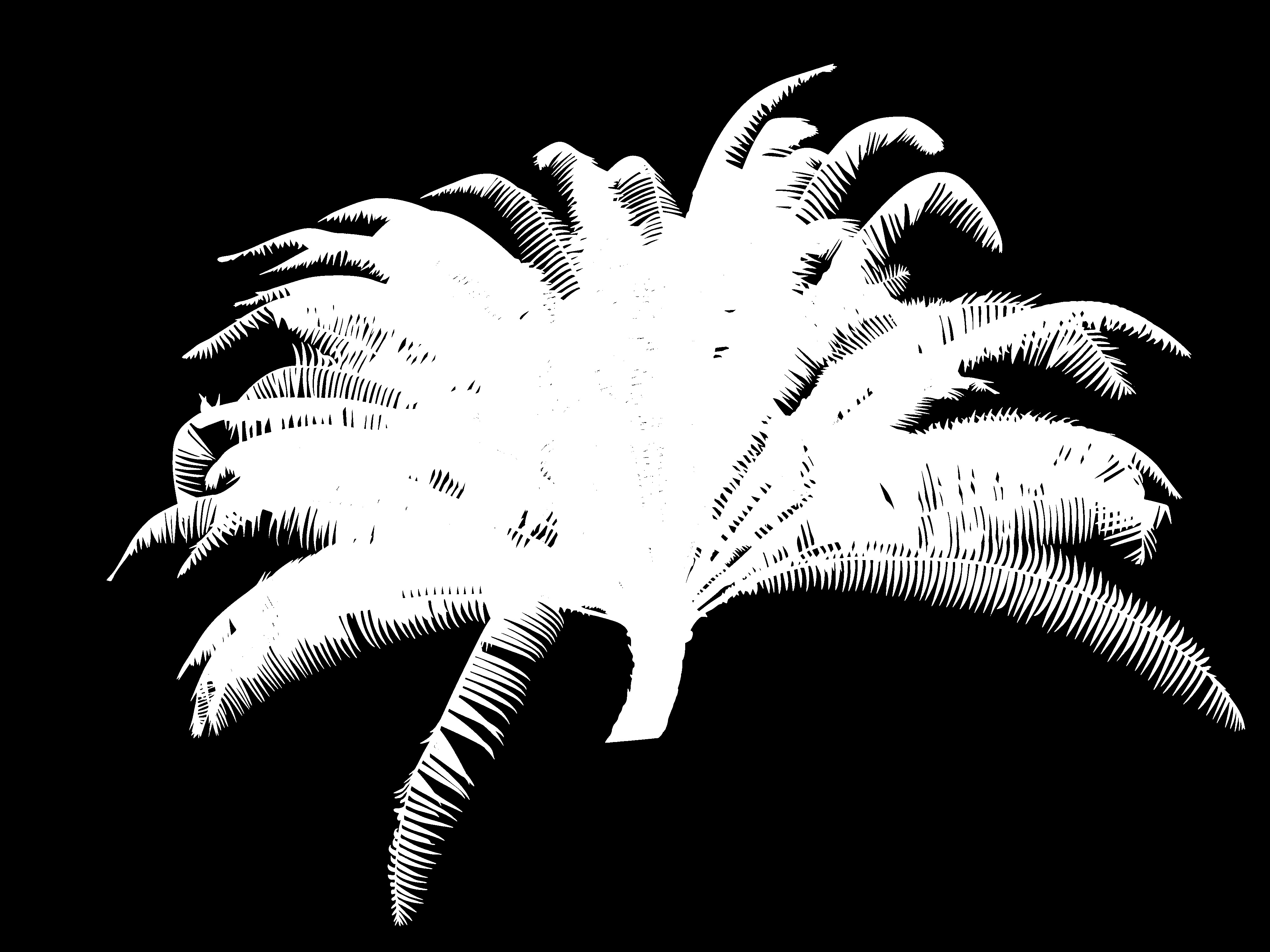}
        \includegraphics[width=\linewidth]{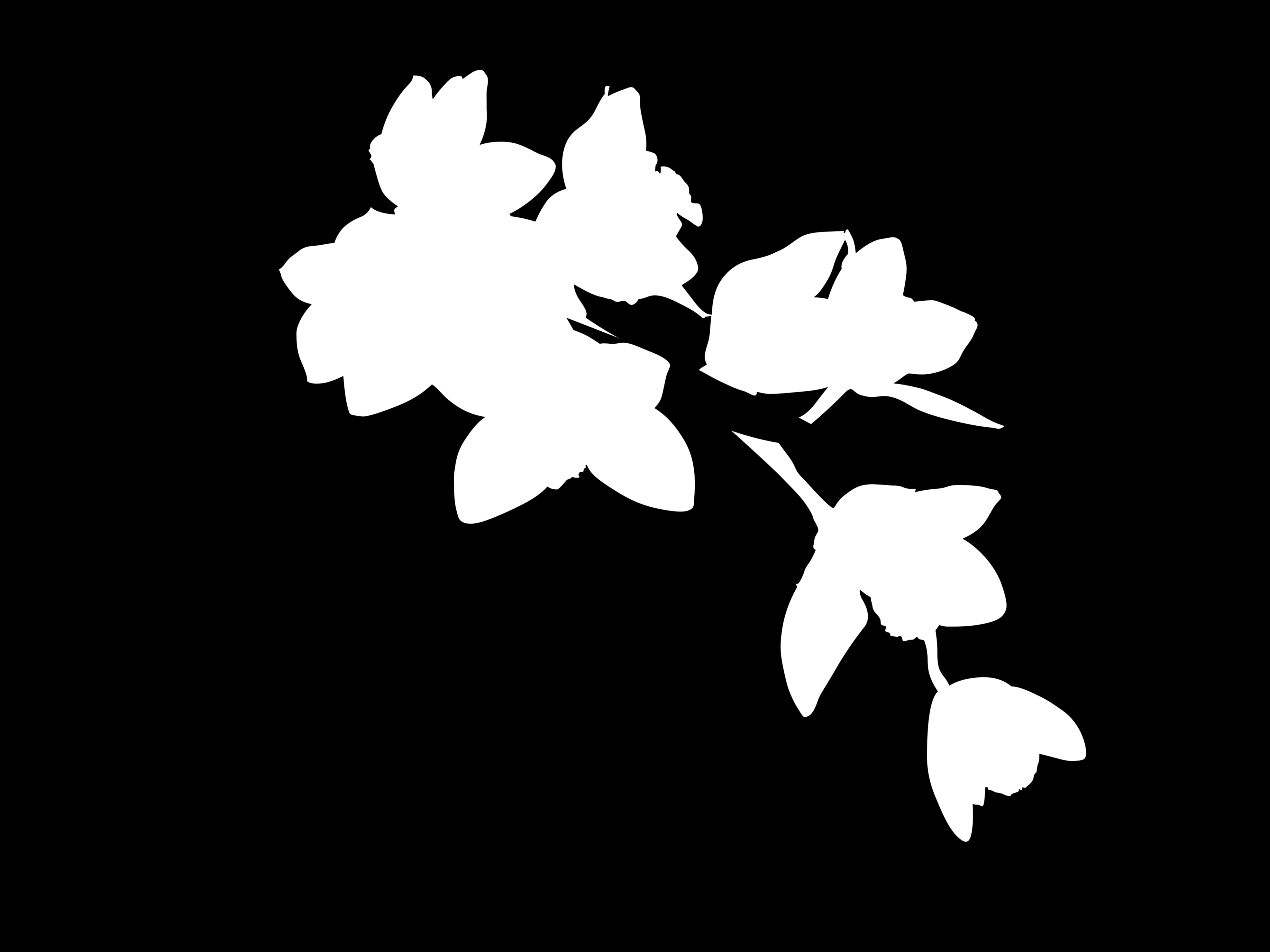}
        \includegraphics[width=\linewidth]{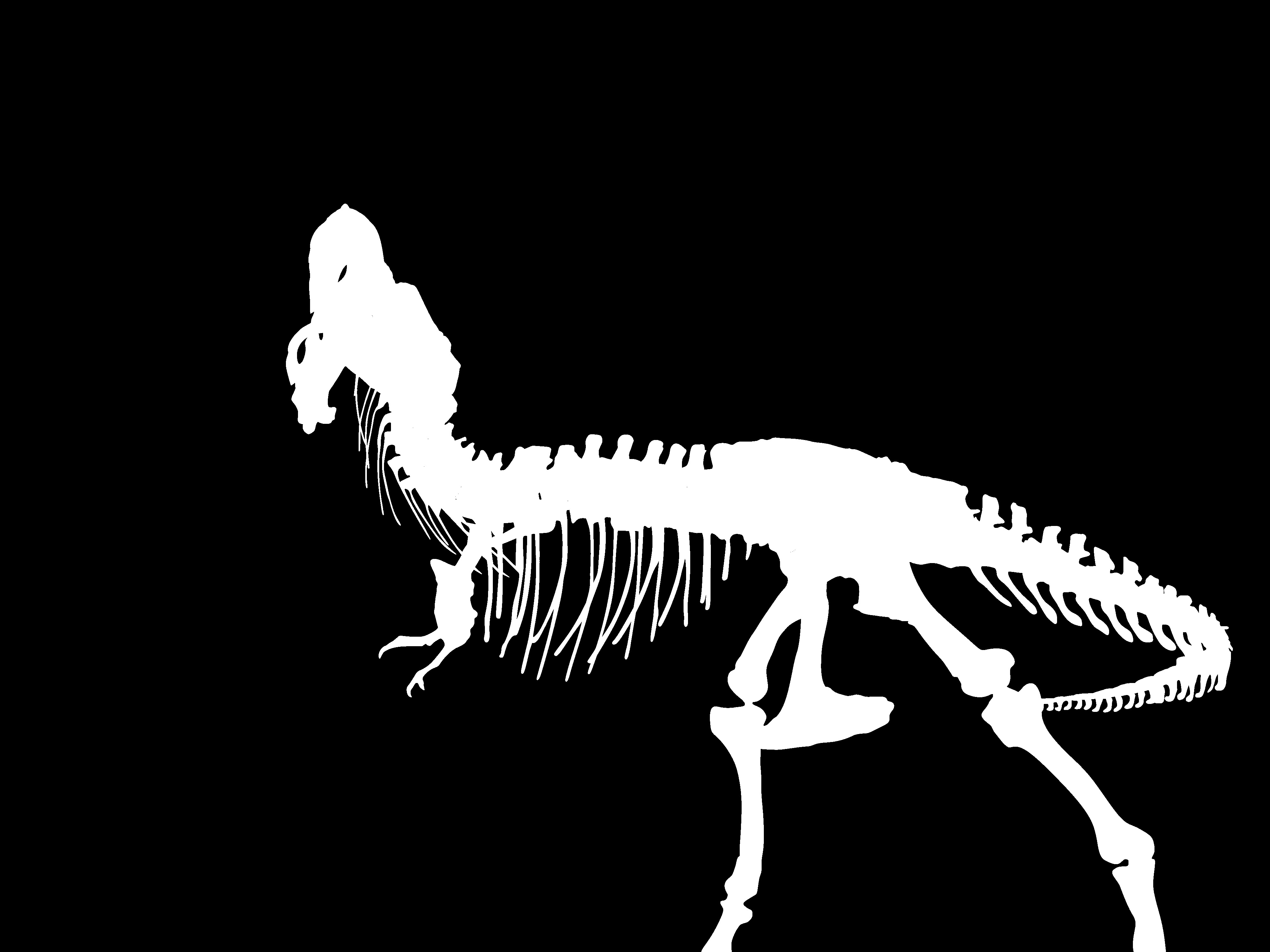}
        \caption{Ground truth mask}
    \end{subfigure}
    \begin{subfigure}{.24\linewidth}
        \includegraphics[width=\linewidth]{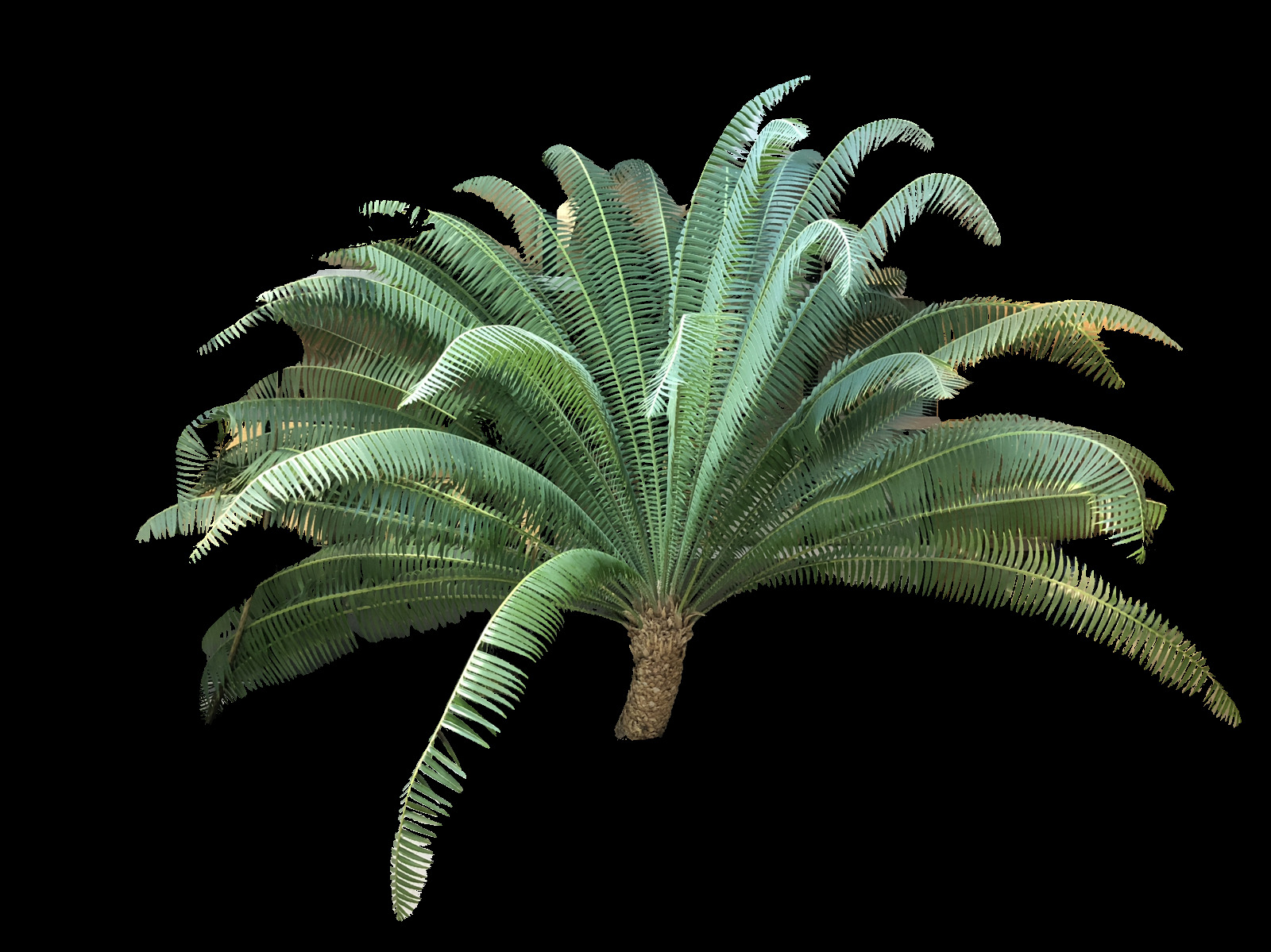}
        \includegraphics[width=\linewidth]{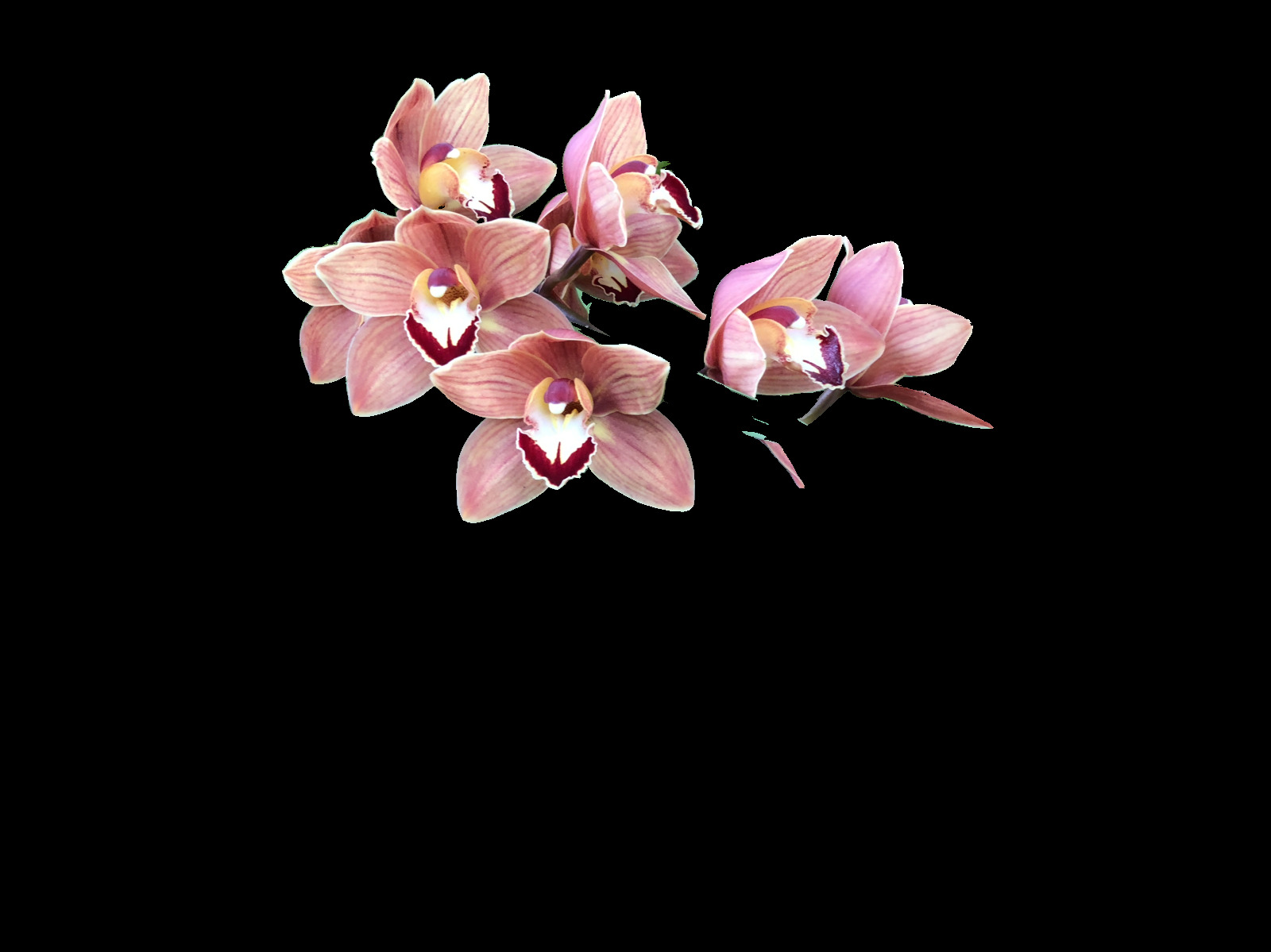}
        \includegraphics[width=\linewidth]{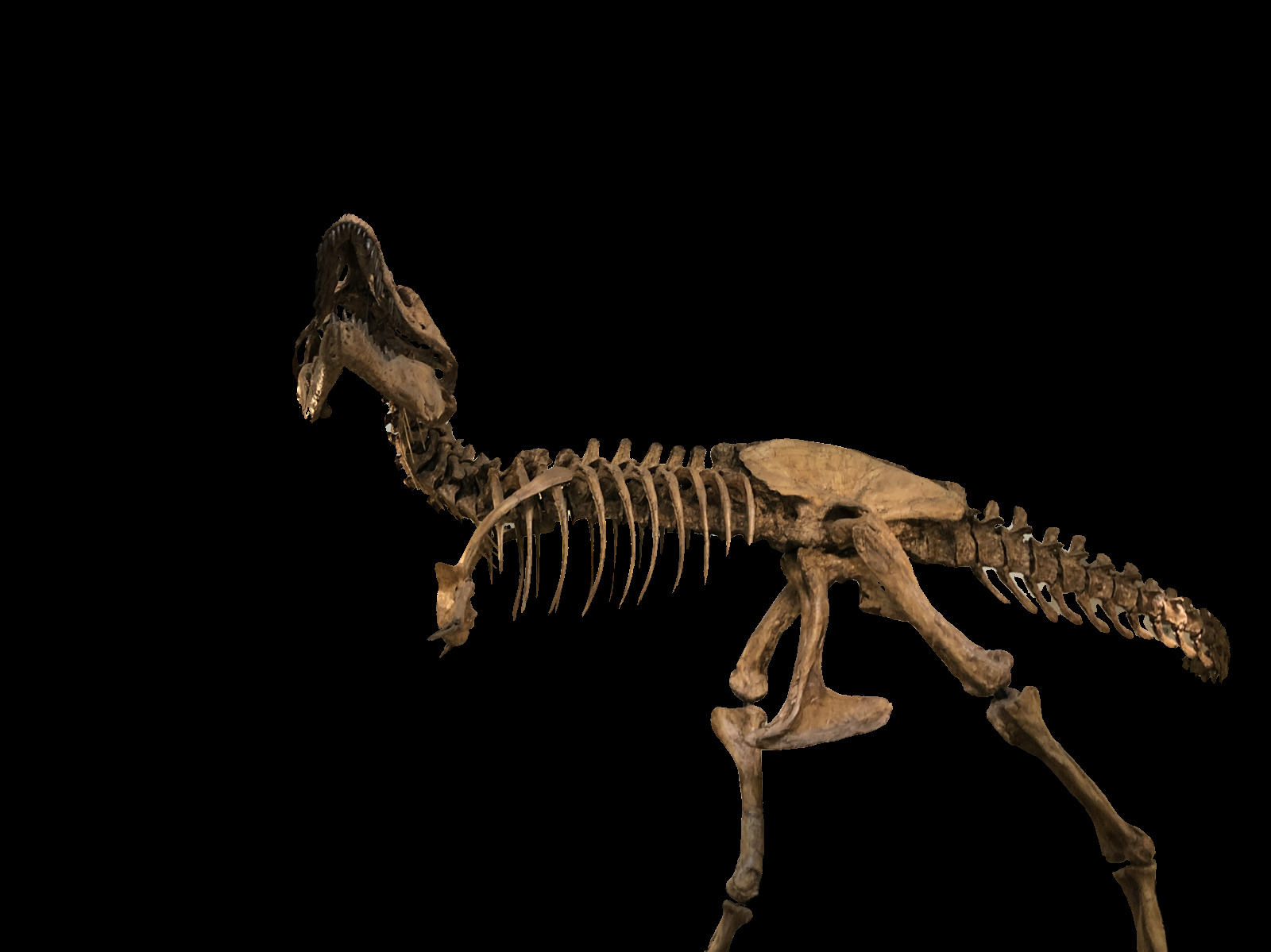}
        \caption{DINOv2 mask}
    \end{subfigure}
    \begin{subfigure}{.24\linewidth}
        \includegraphics[width=\linewidth]{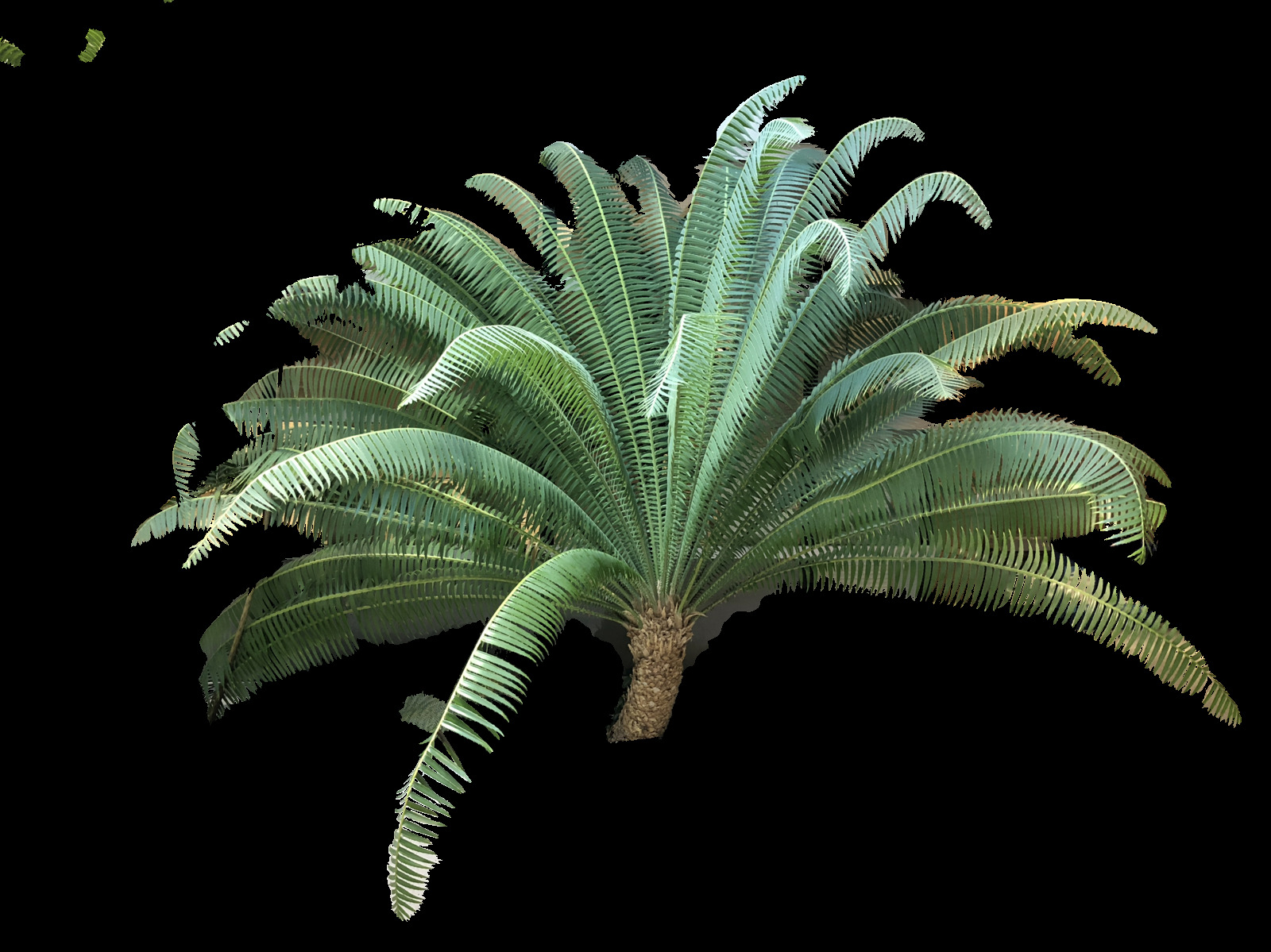}
        \includegraphics[width=\linewidth]{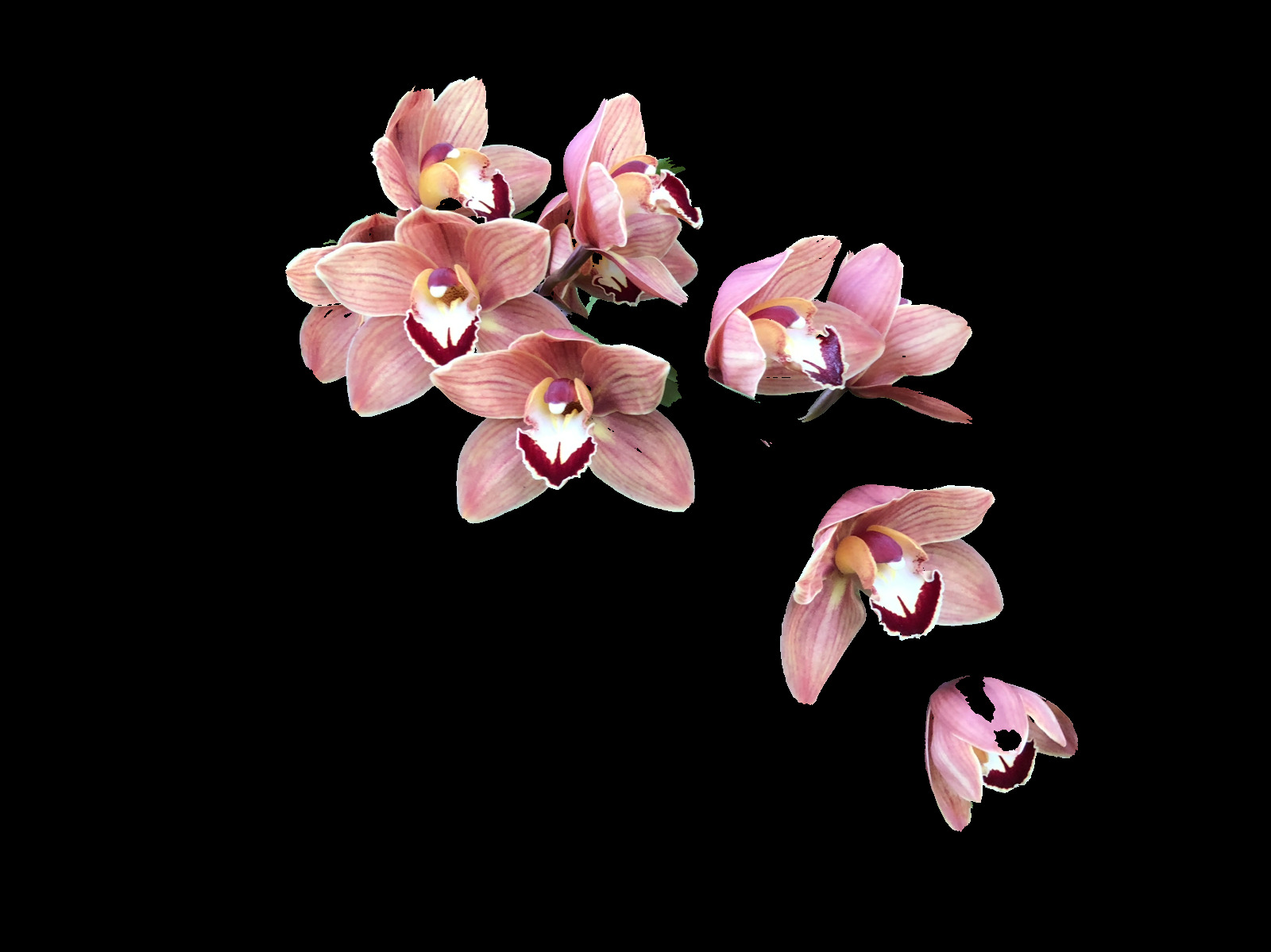}
        \includegraphics[width=\linewidth]{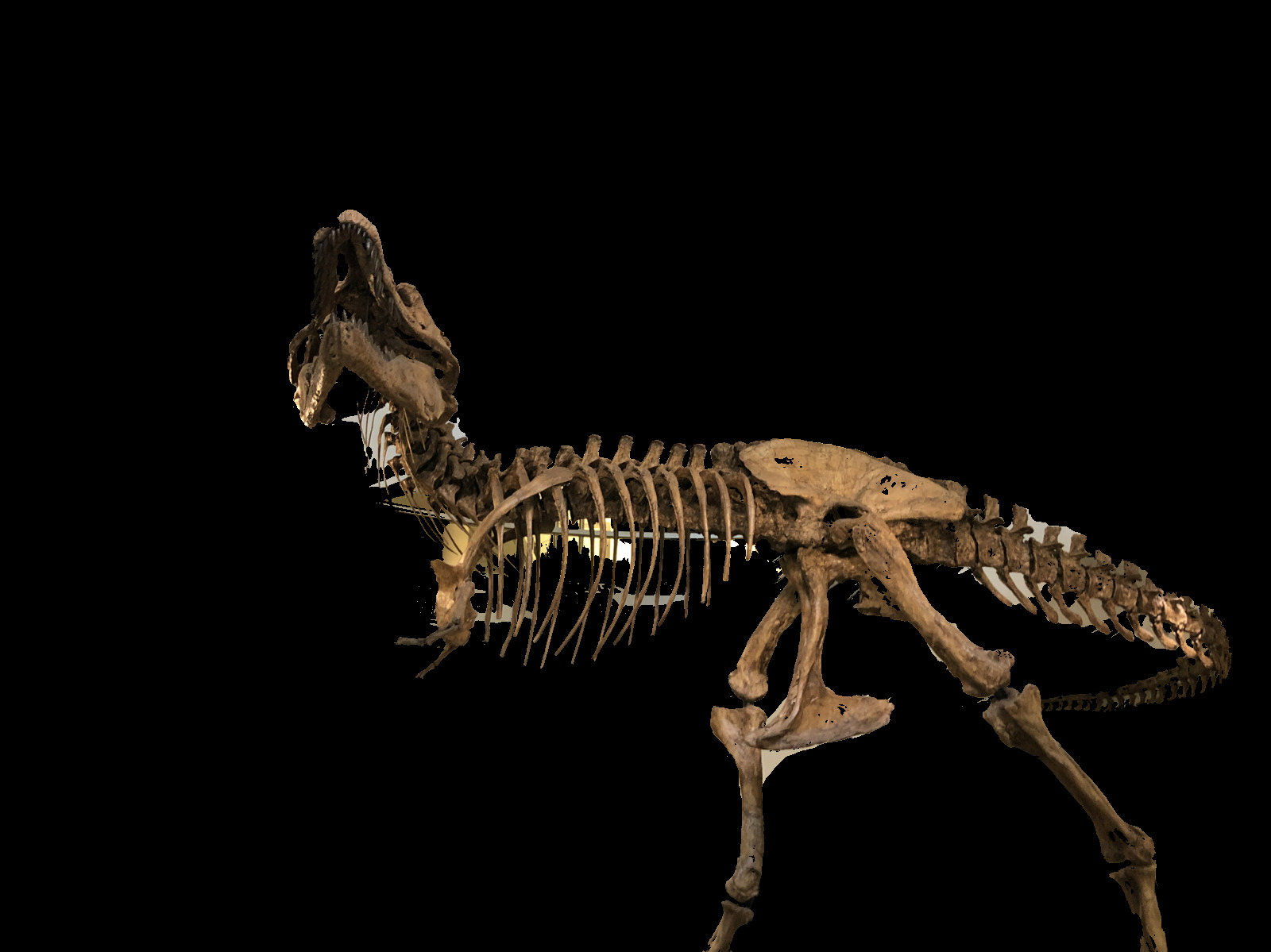}
        \caption{SAM mask}
    \end{subfigure}
    \caption{Segmentation results on NVOS~\citep{ren2022nvos} with DINOv2 and SAM.}
    \label{fig:appendix-nvos}
\end{figure*}

\myparagraph{Segmentation on NVOS}
Fig.~\ref{fig:appendix-nvos} shows our segmentation masks from SAM and DINOv2 for the three most challenging scenes of the NVOS dataset: Fern, Orchids and Trex.

\begin{figure*}[t!]
    \centering
    \includegraphics[width=0.2\linewidth]{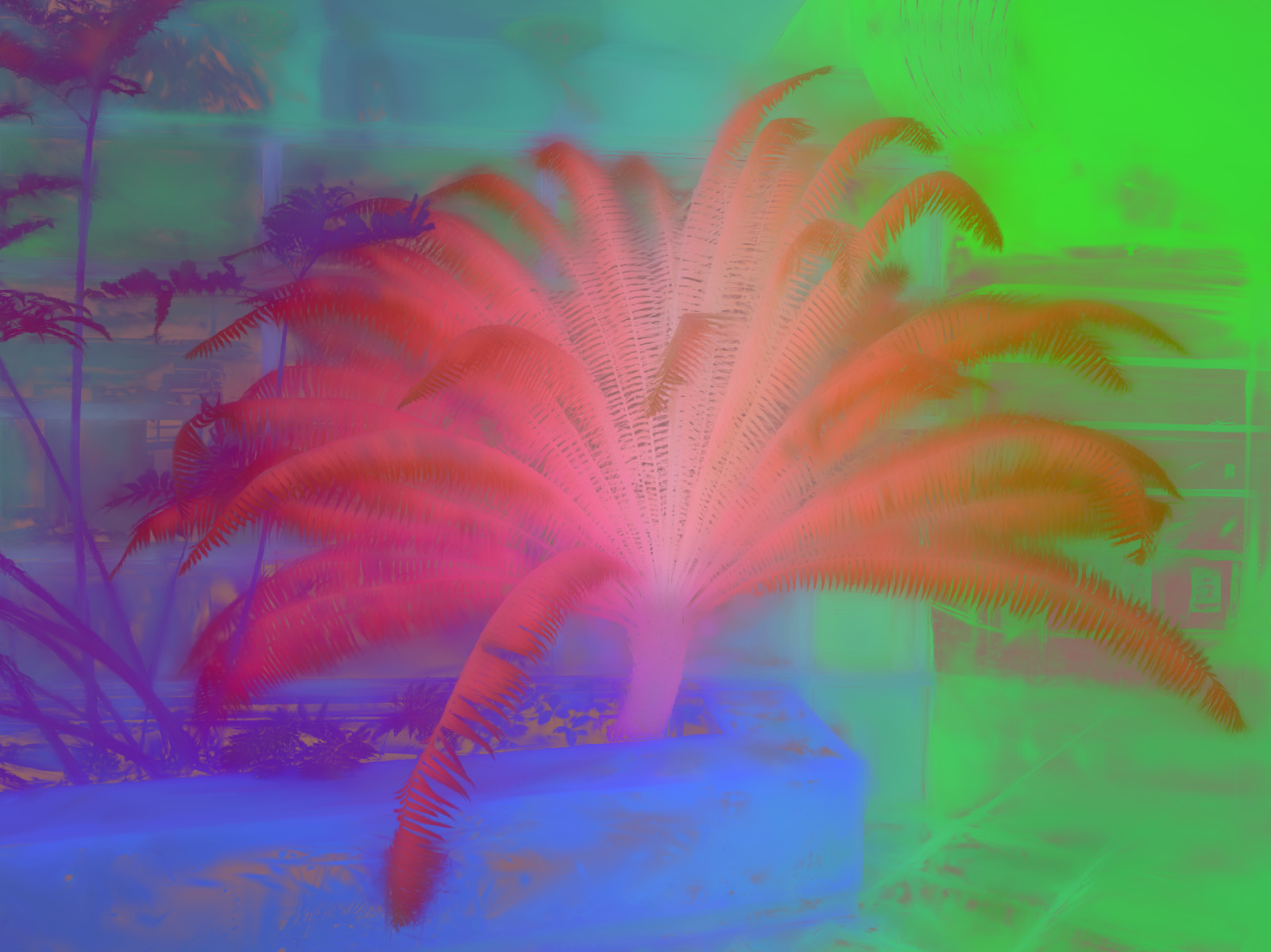}
    \includegraphics[width=0.2\linewidth]{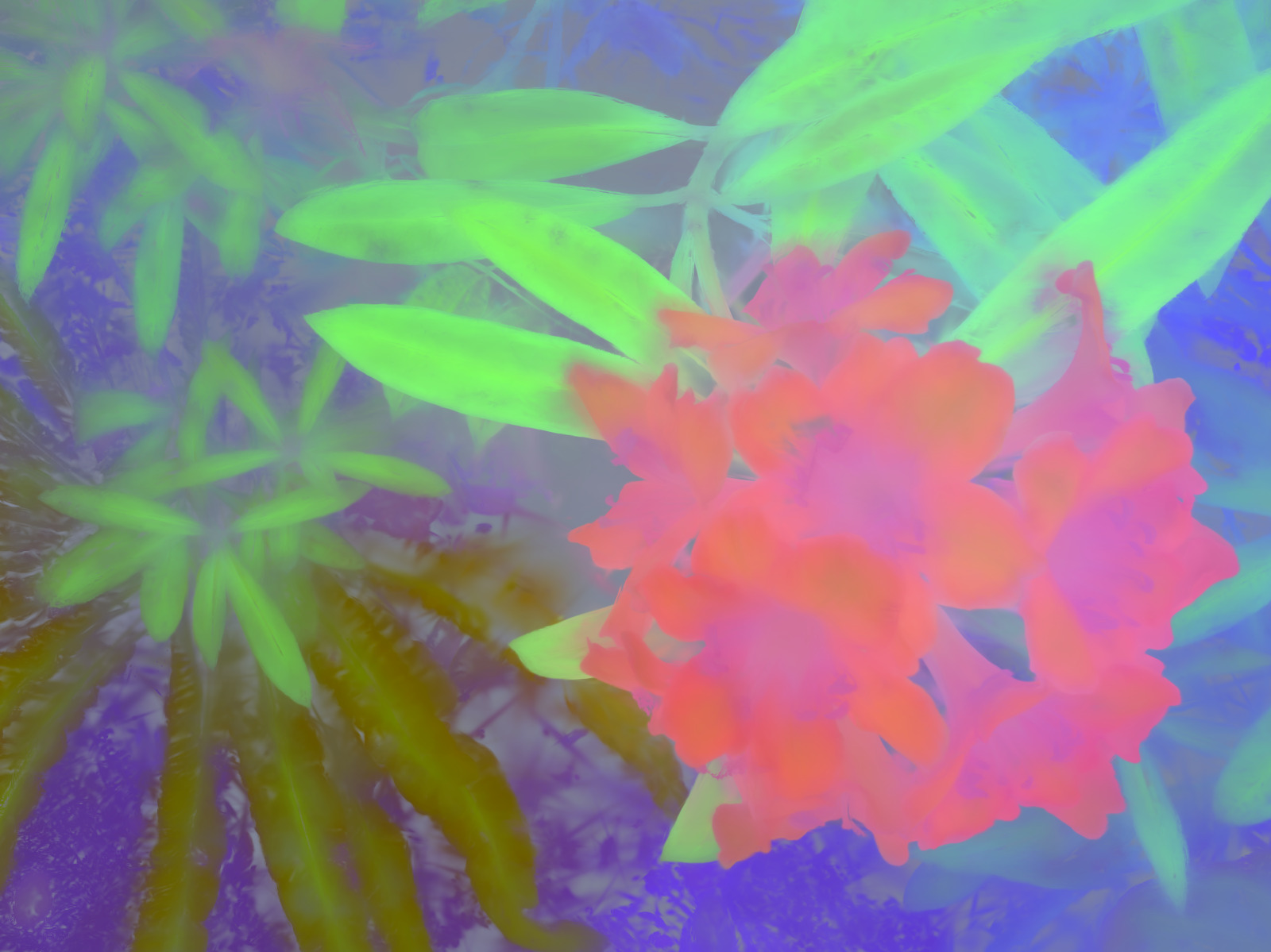}
    \includegraphics[width=0.2\linewidth]{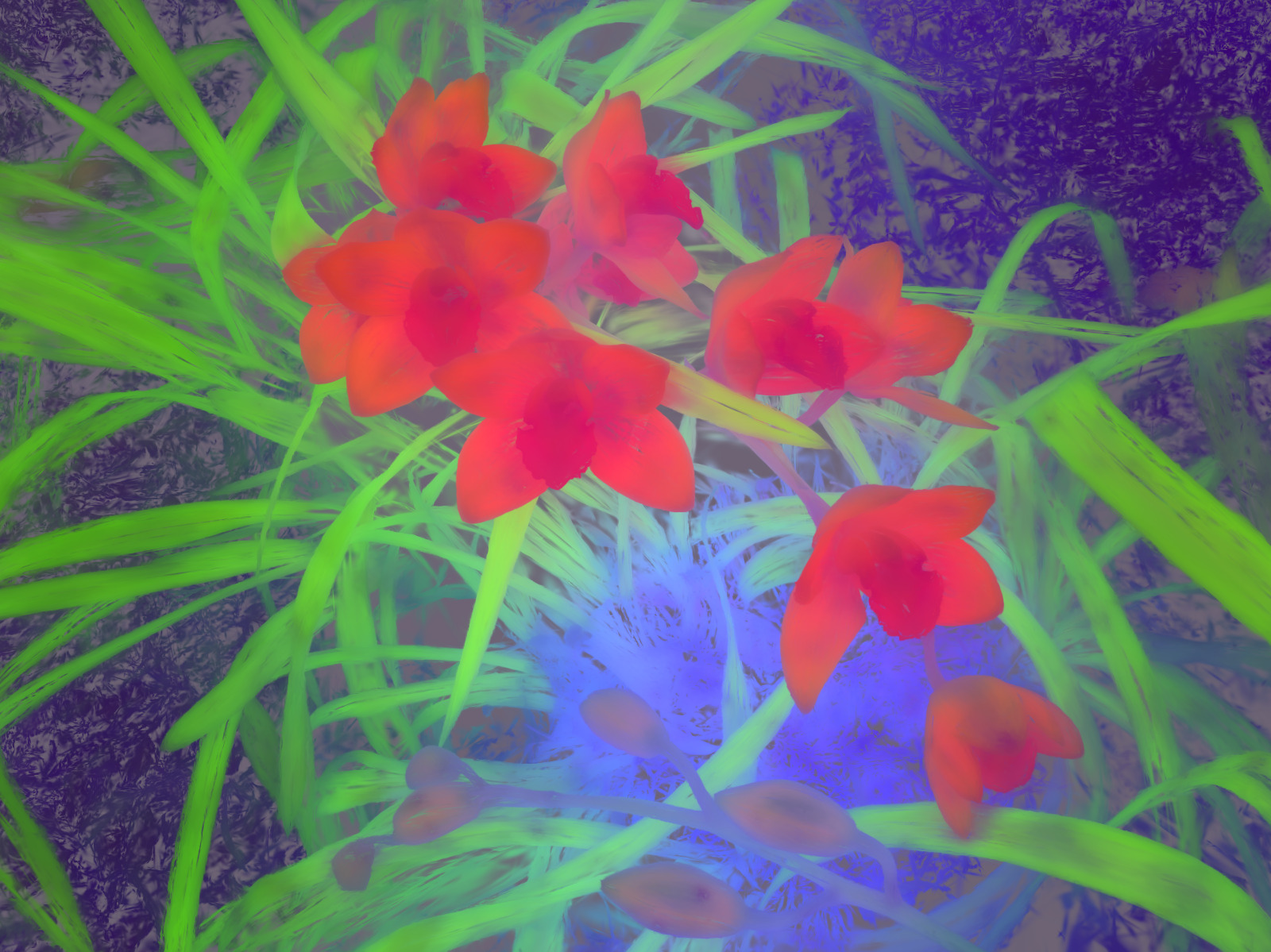}
    \includegraphics[width=0.2\linewidth]{figures/pcas/trex_pca.jpg} \\
    \includegraphics[width=0.2\linewidth]{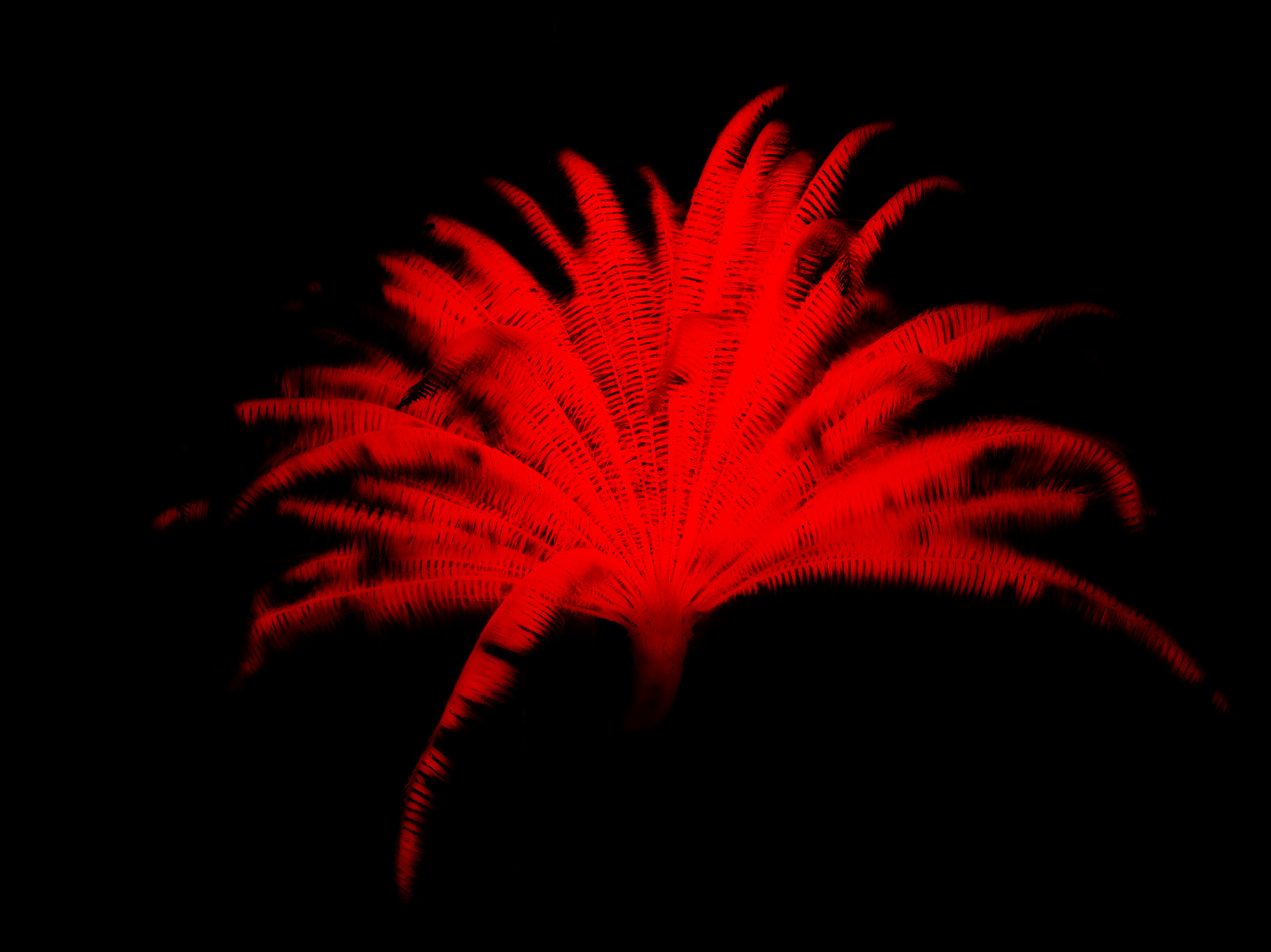}
    \includegraphics[width=0.2\linewidth]{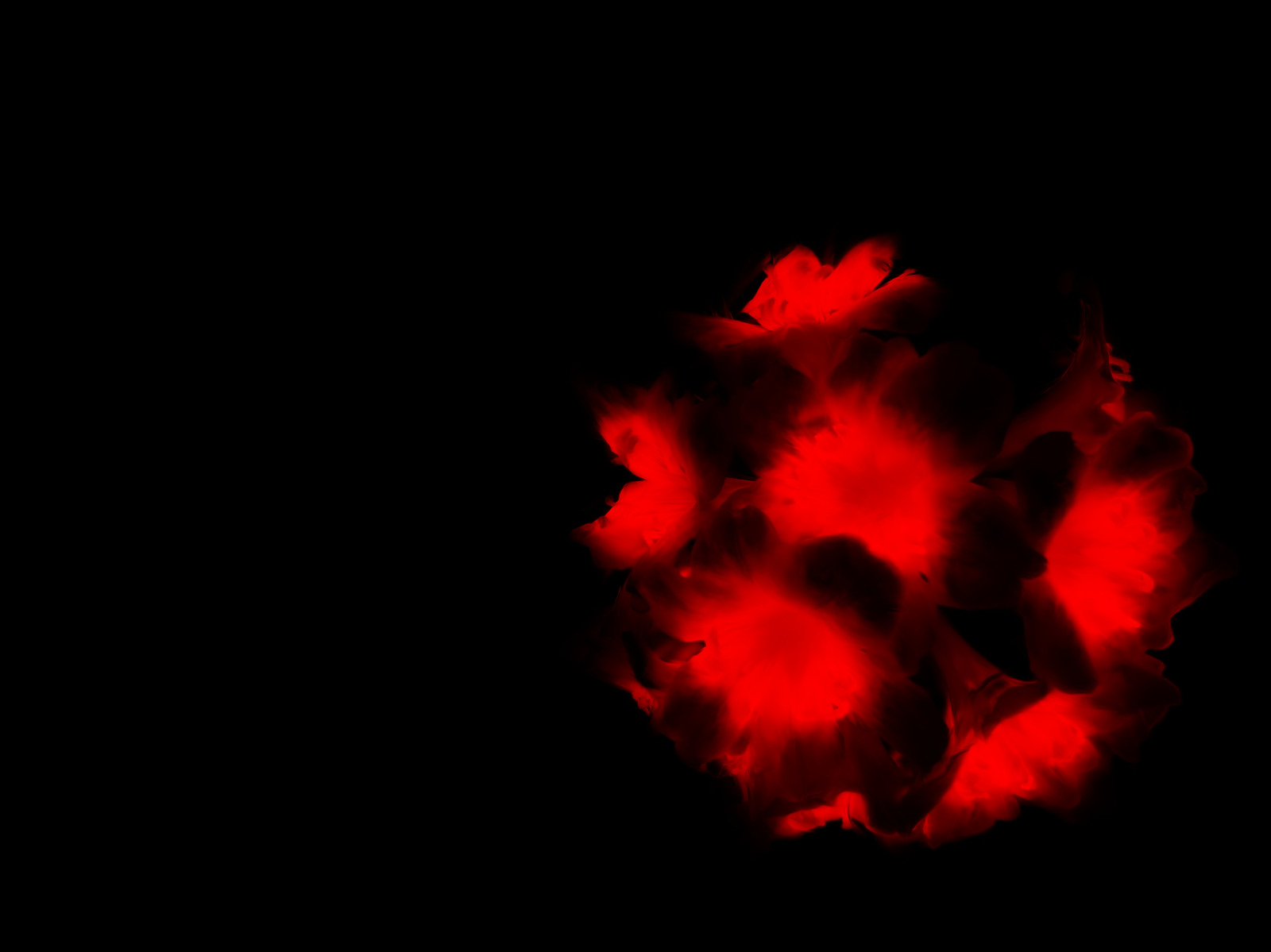}
    \includegraphics[width=0.2\linewidth]{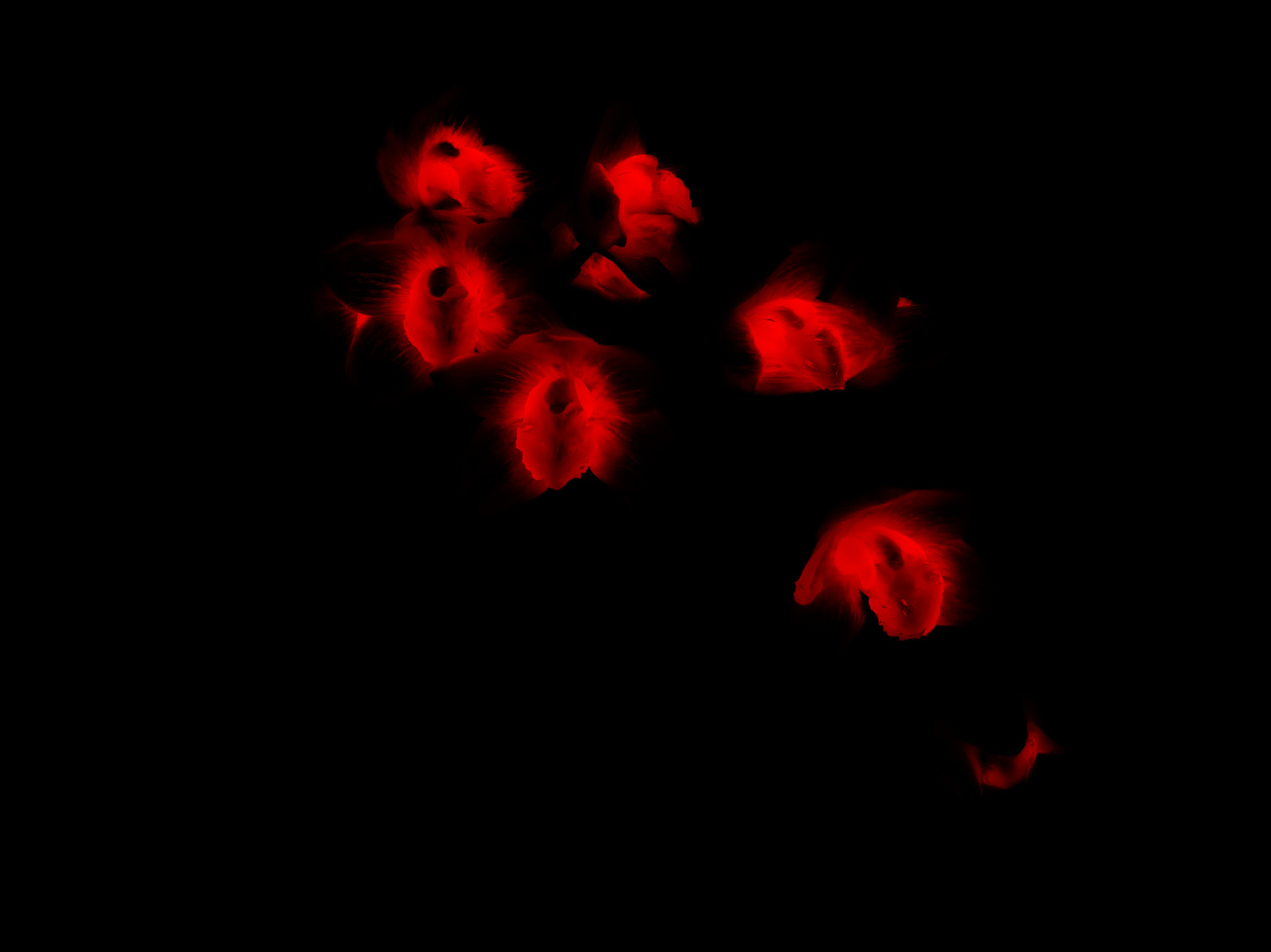}
    \includegraphics[width=0.2\linewidth]{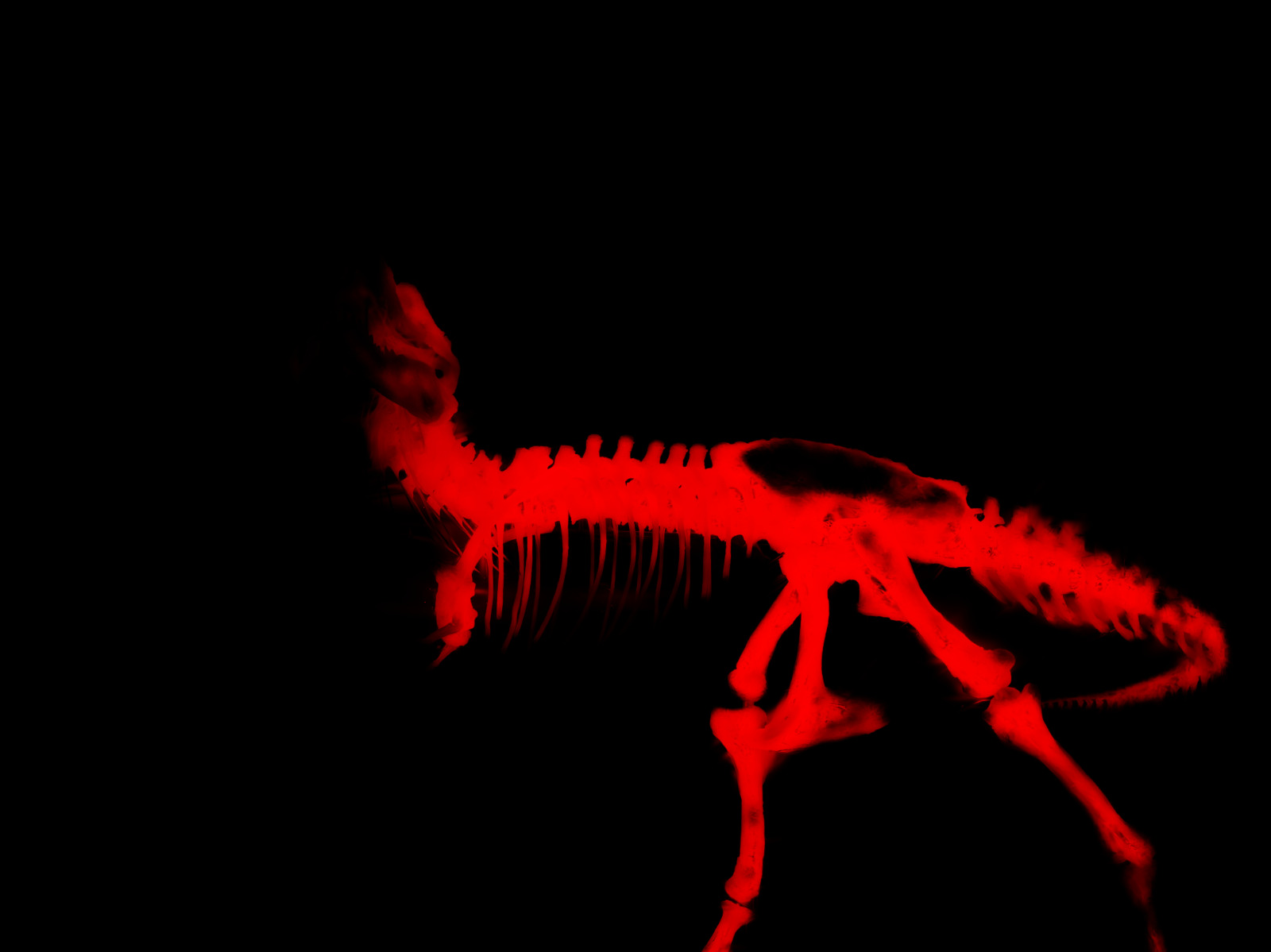} \\
    \includegraphics[width=0.2\linewidth]{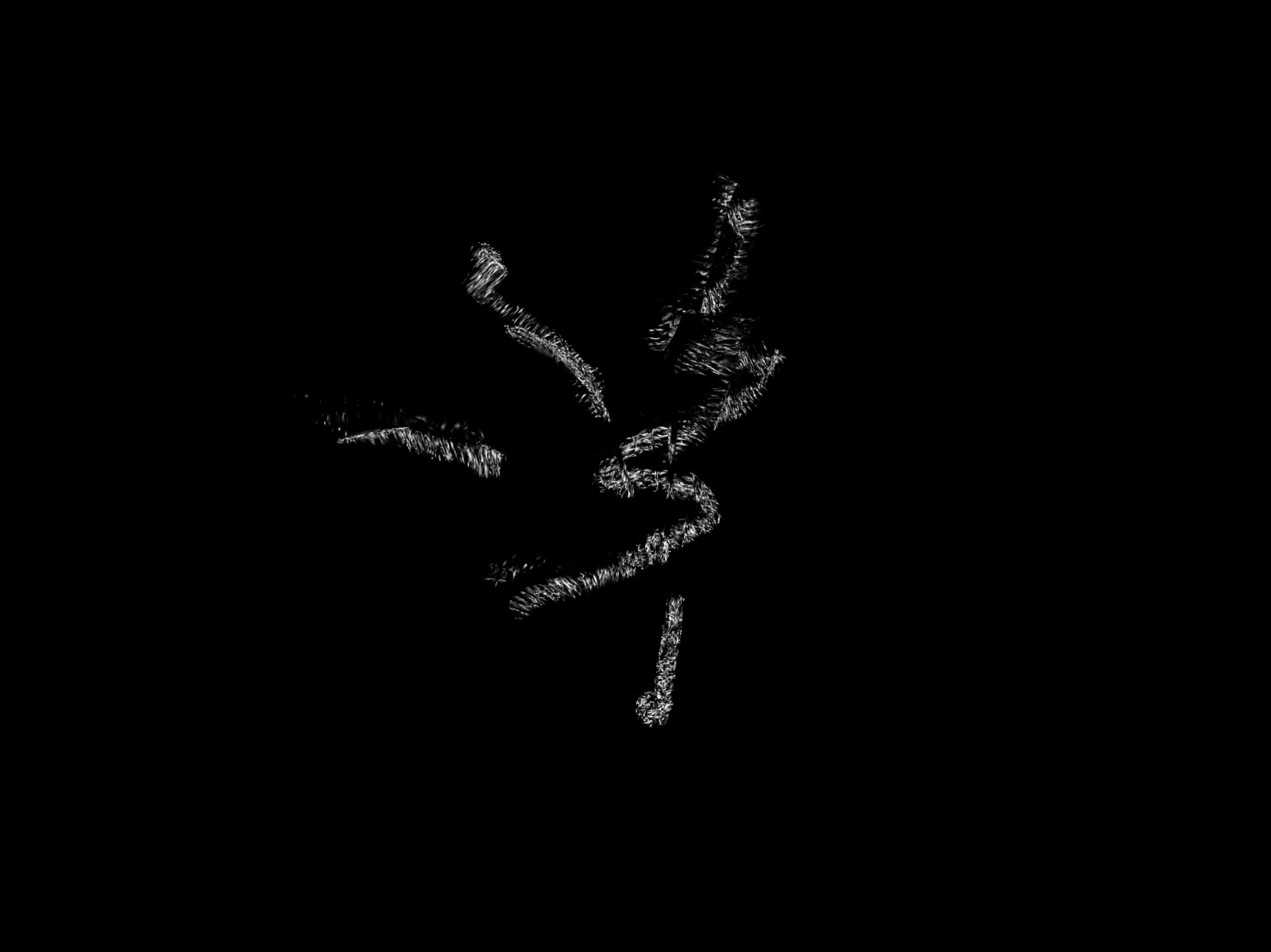}
    \includegraphics[width=0.2\linewidth]{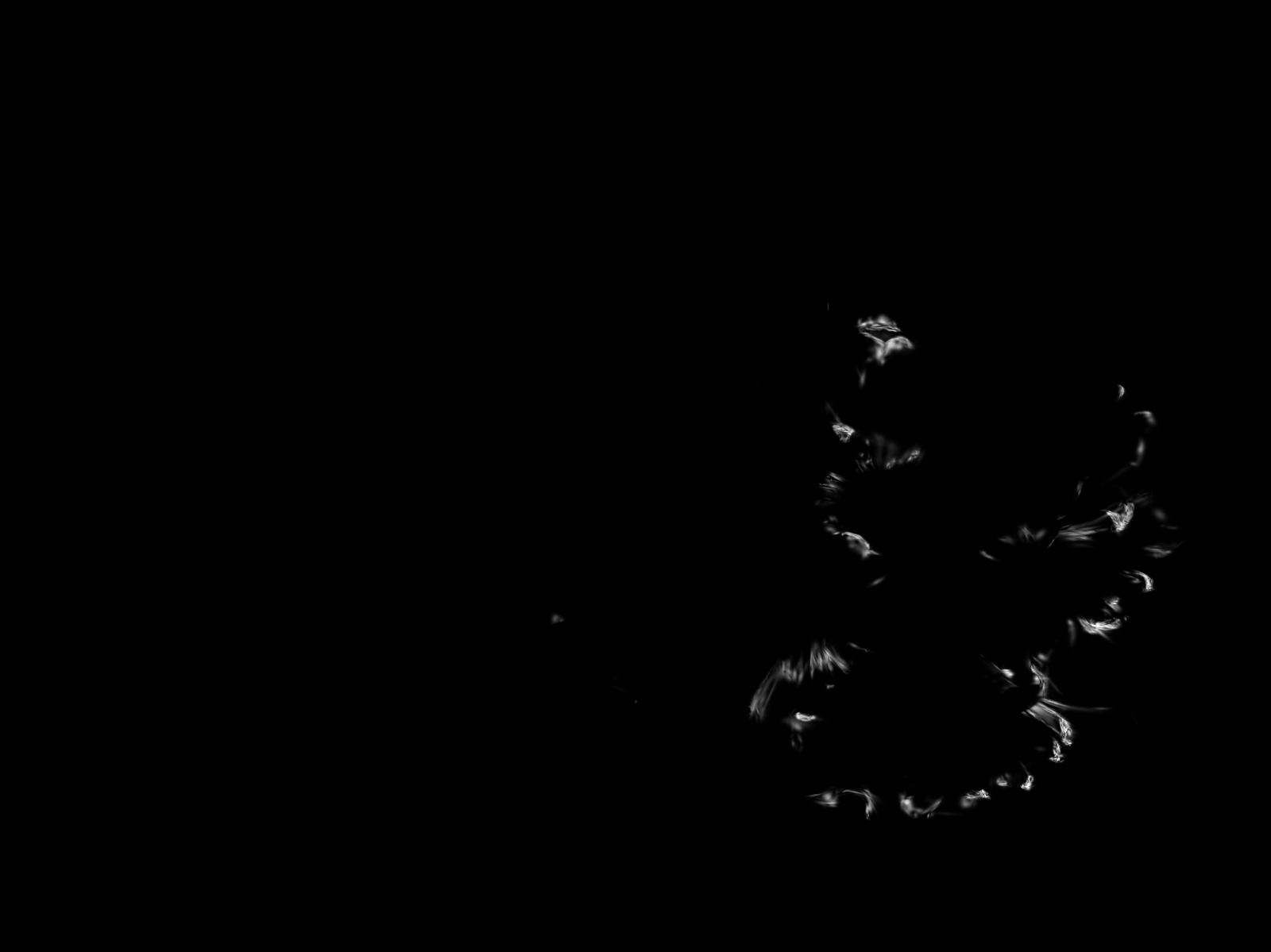}
    \includegraphics[width=0.2\linewidth]{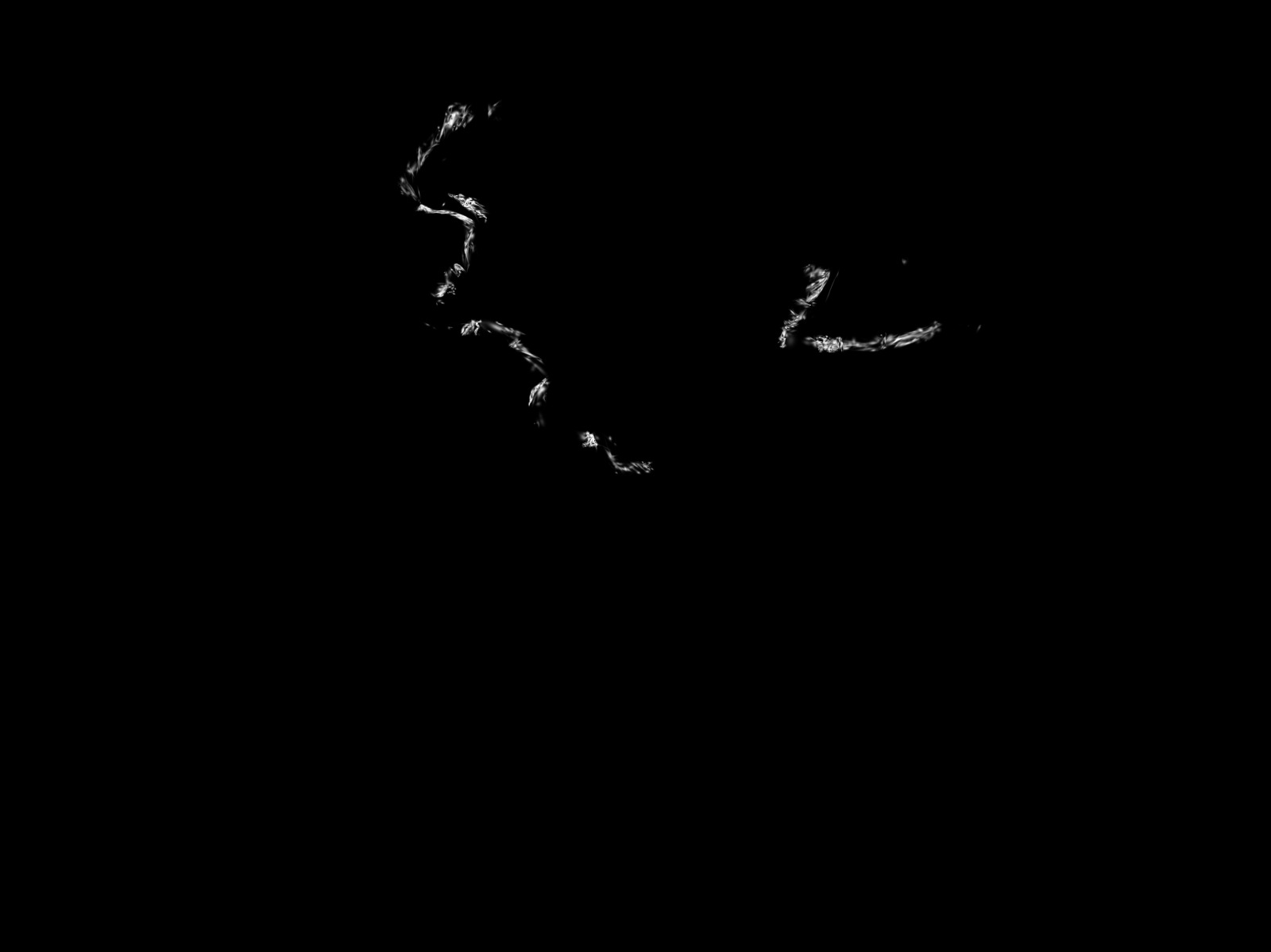}
    \includegraphics[width=0.2\linewidth]{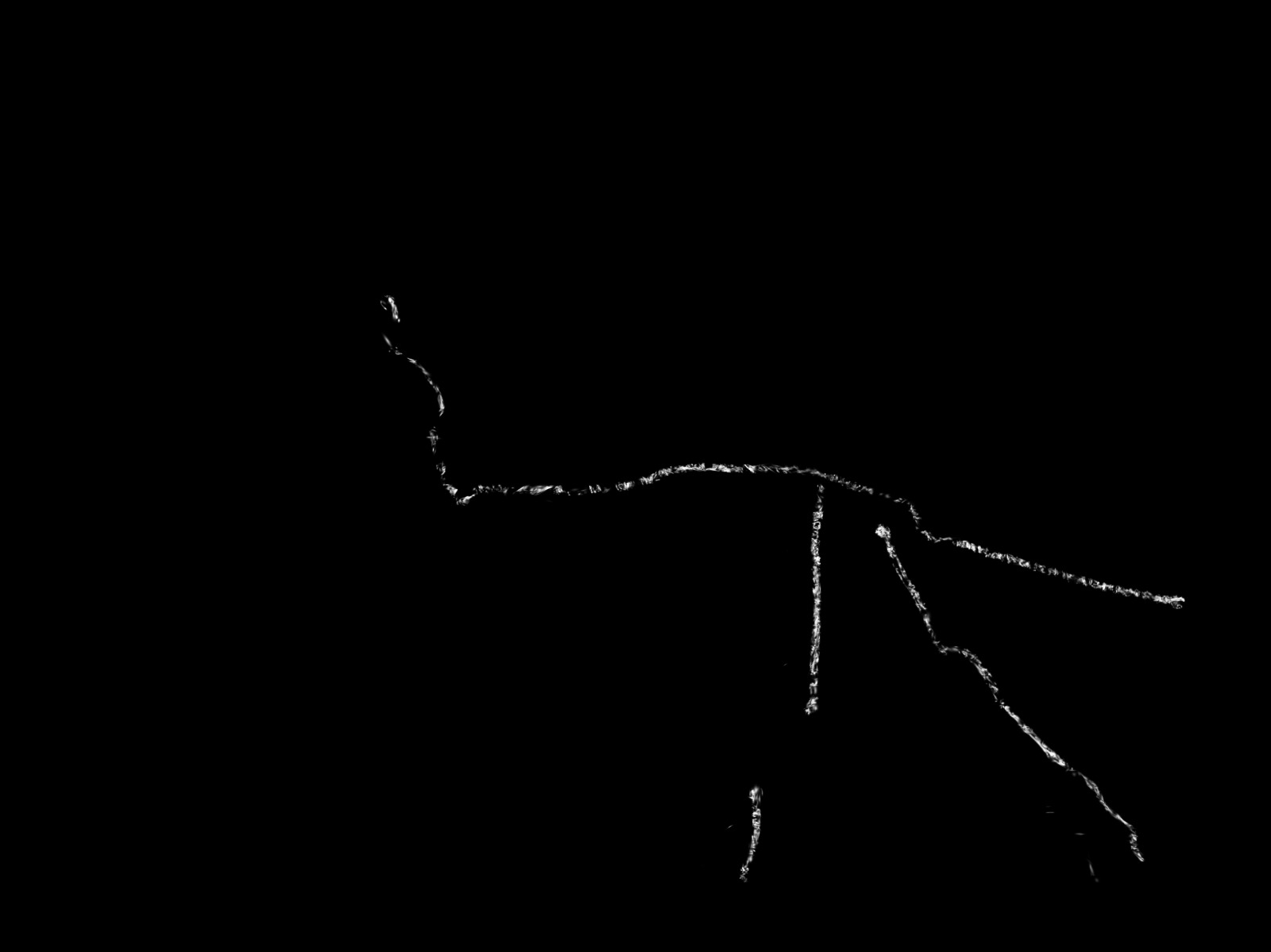} \\
    \includegraphics[width=0.2\linewidth]{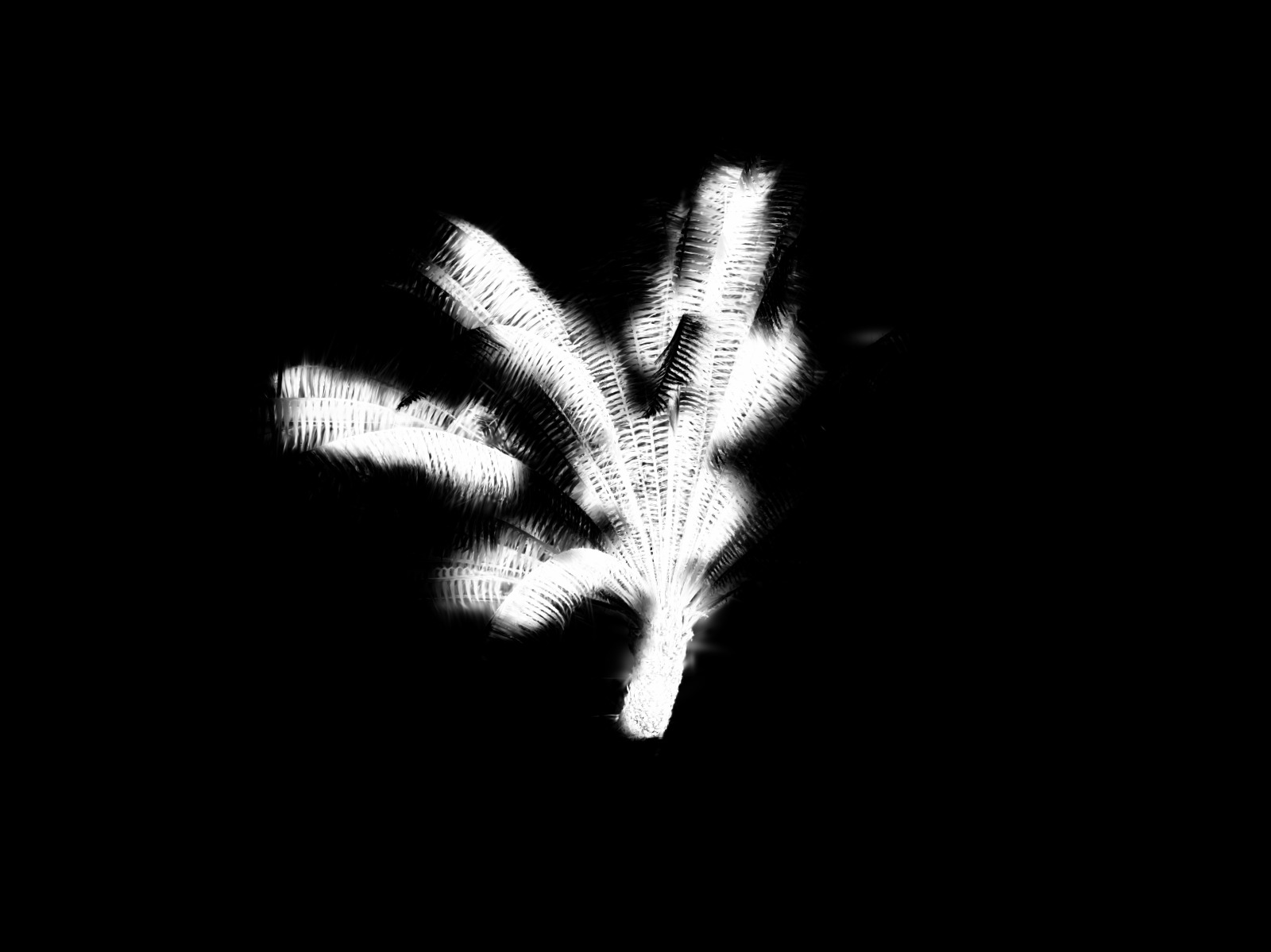}
    \includegraphics[width=0.2\linewidth]{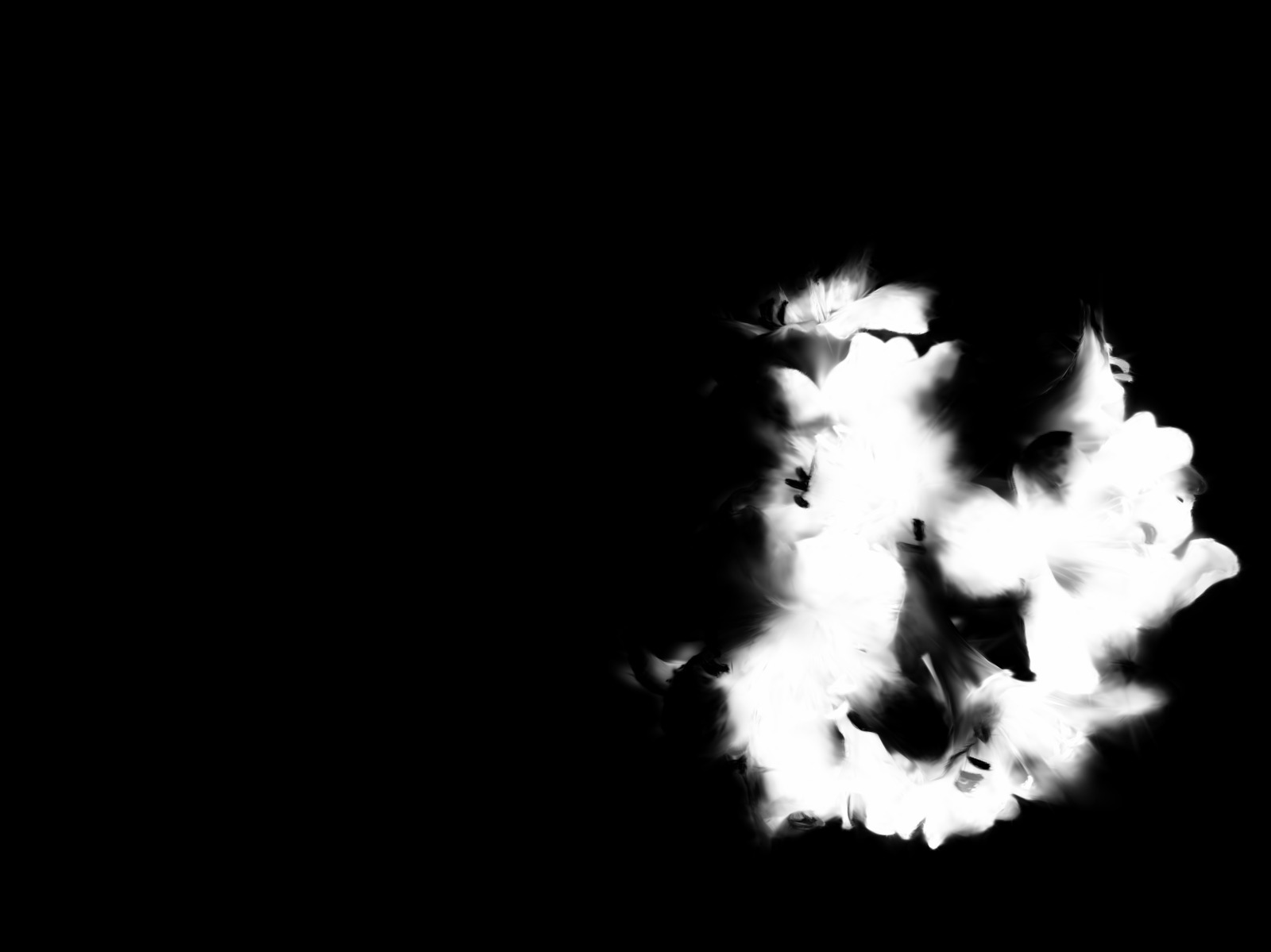}
    \includegraphics[width=0.2\linewidth]{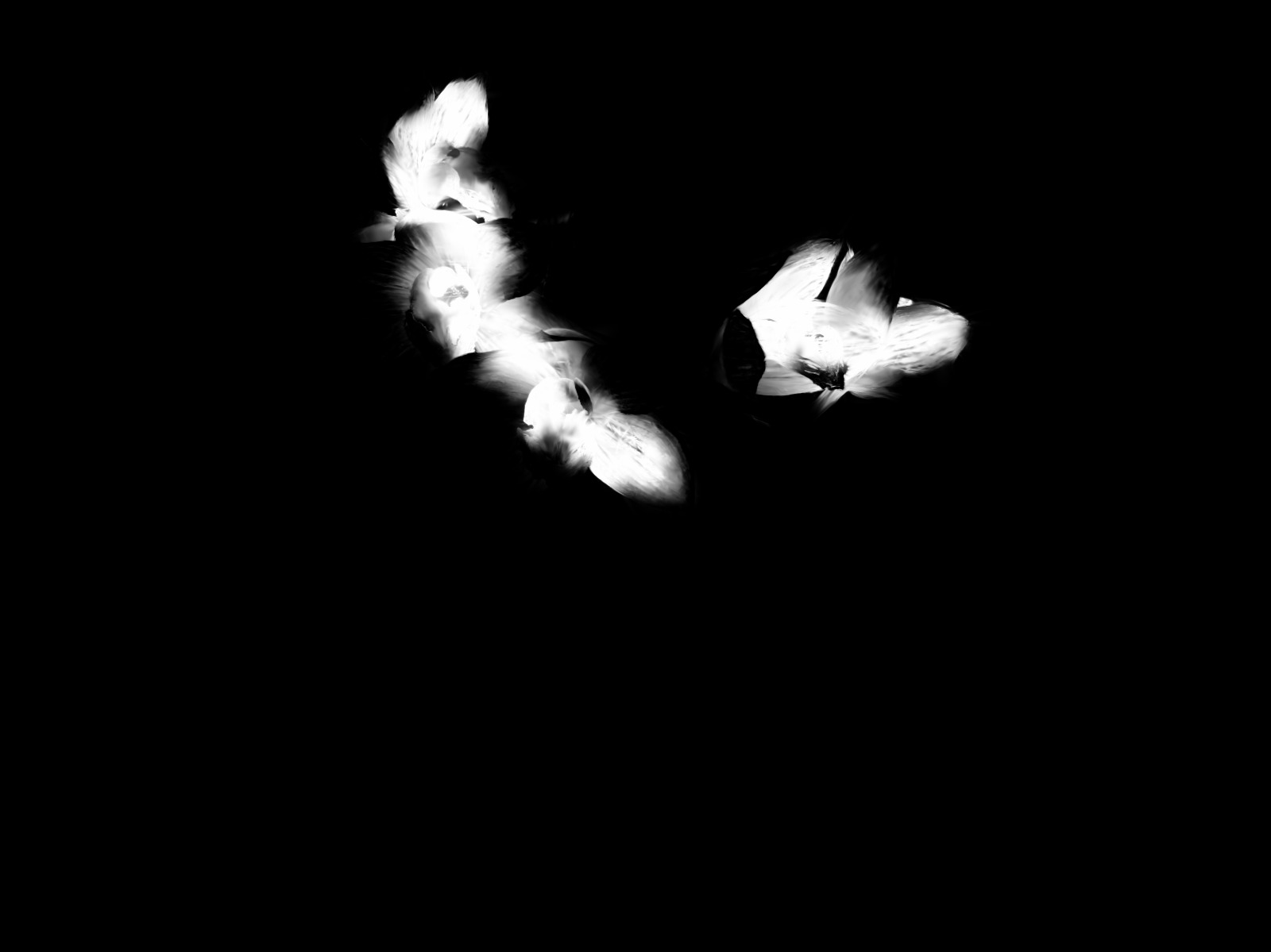}
    \includegraphics[width=0.2\linewidth]{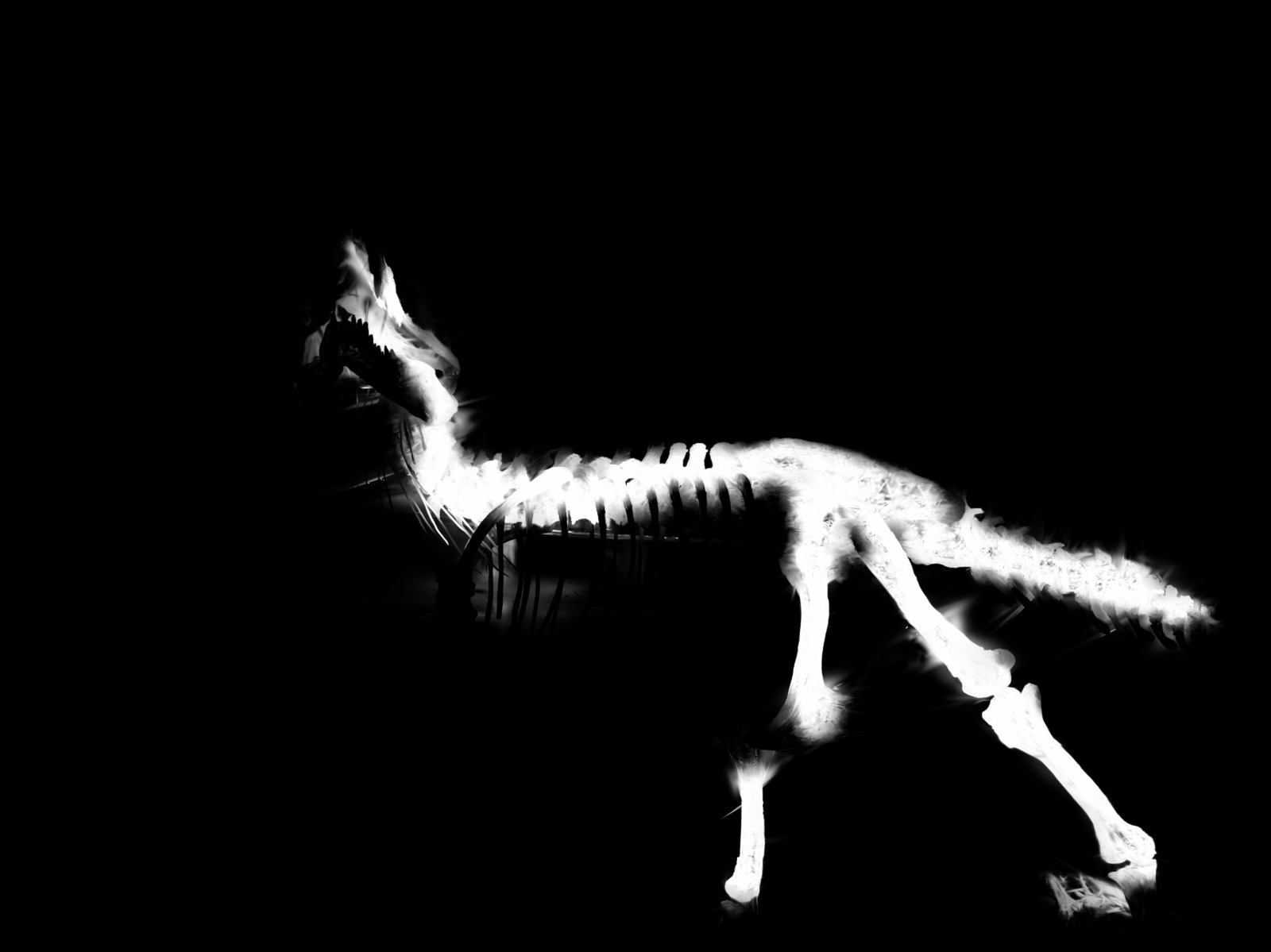} \\
    \includegraphics[width=0.2\linewidth]{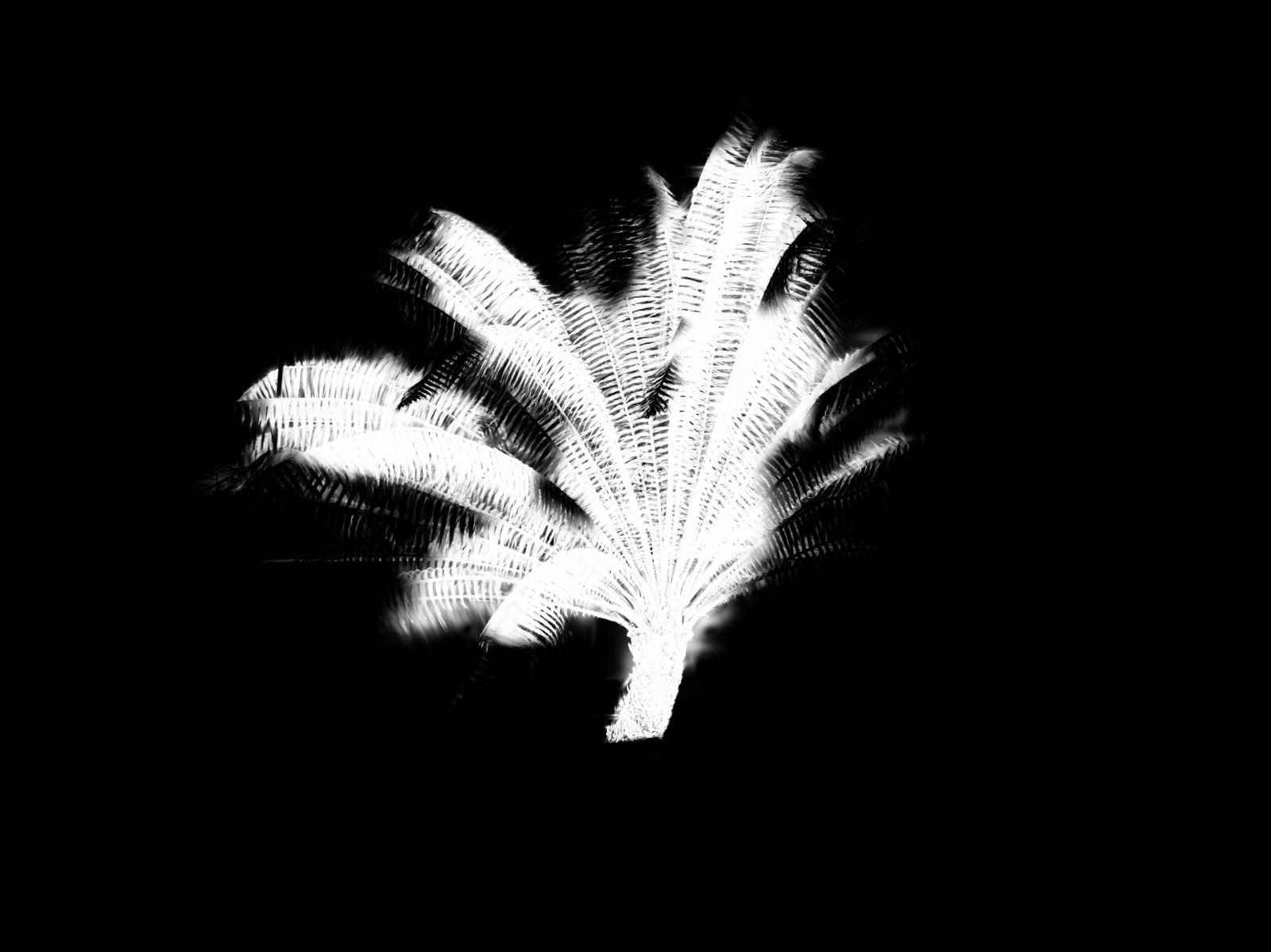}
    \includegraphics[width=0.2\linewidth]{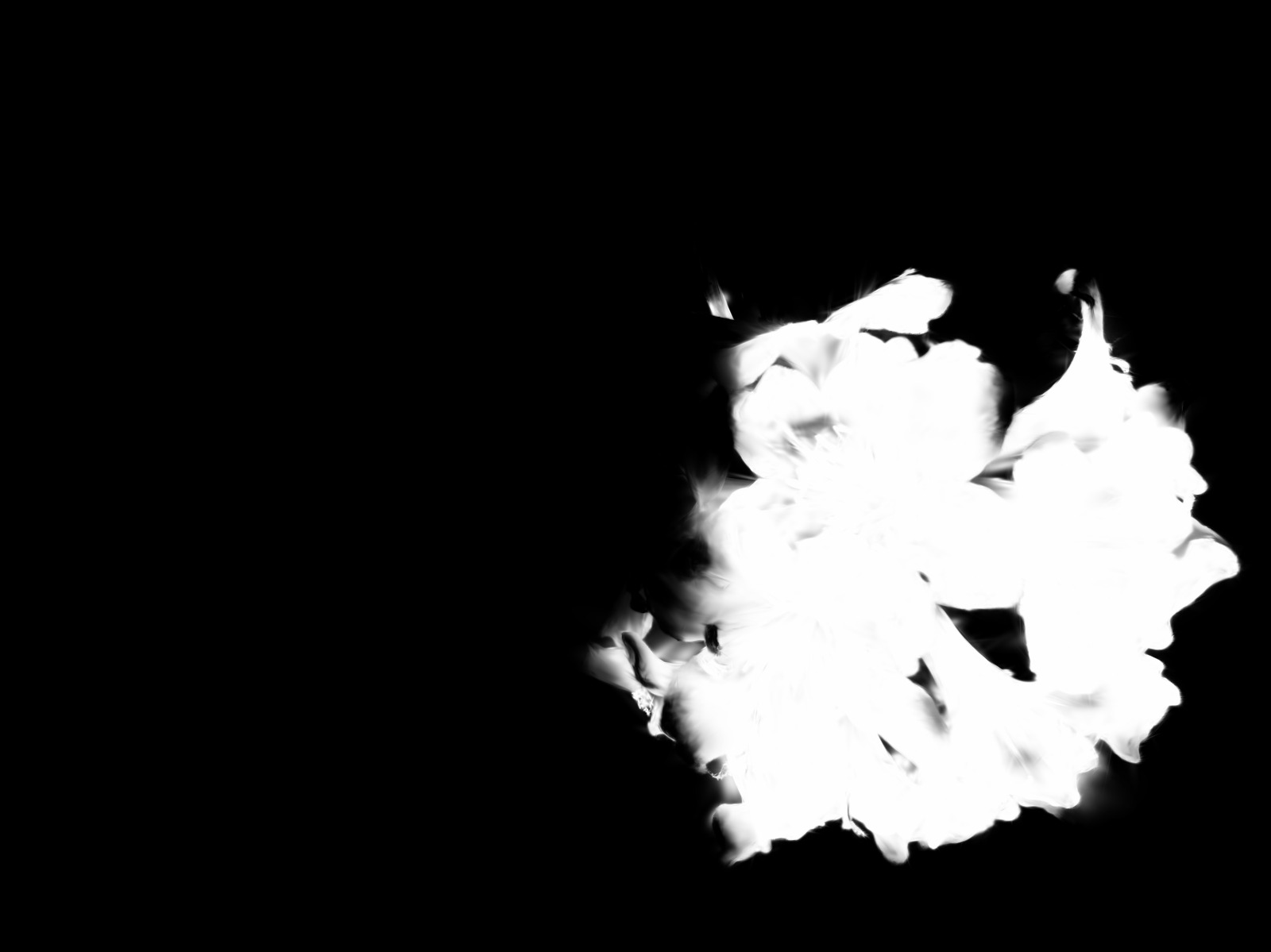}
    \includegraphics[width=0.2\linewidth]{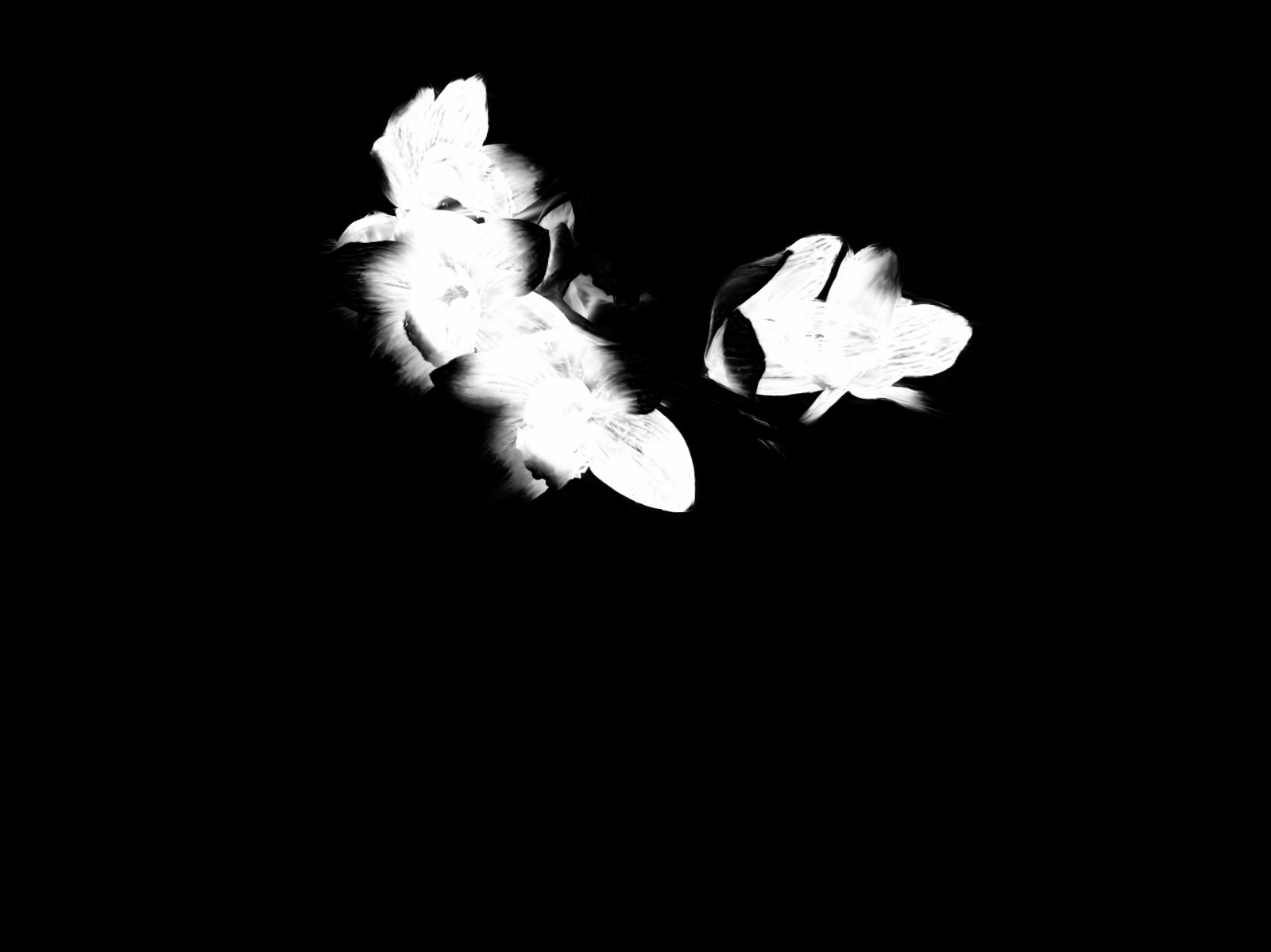}
    \includegraphics[width=0.2\linewidth]{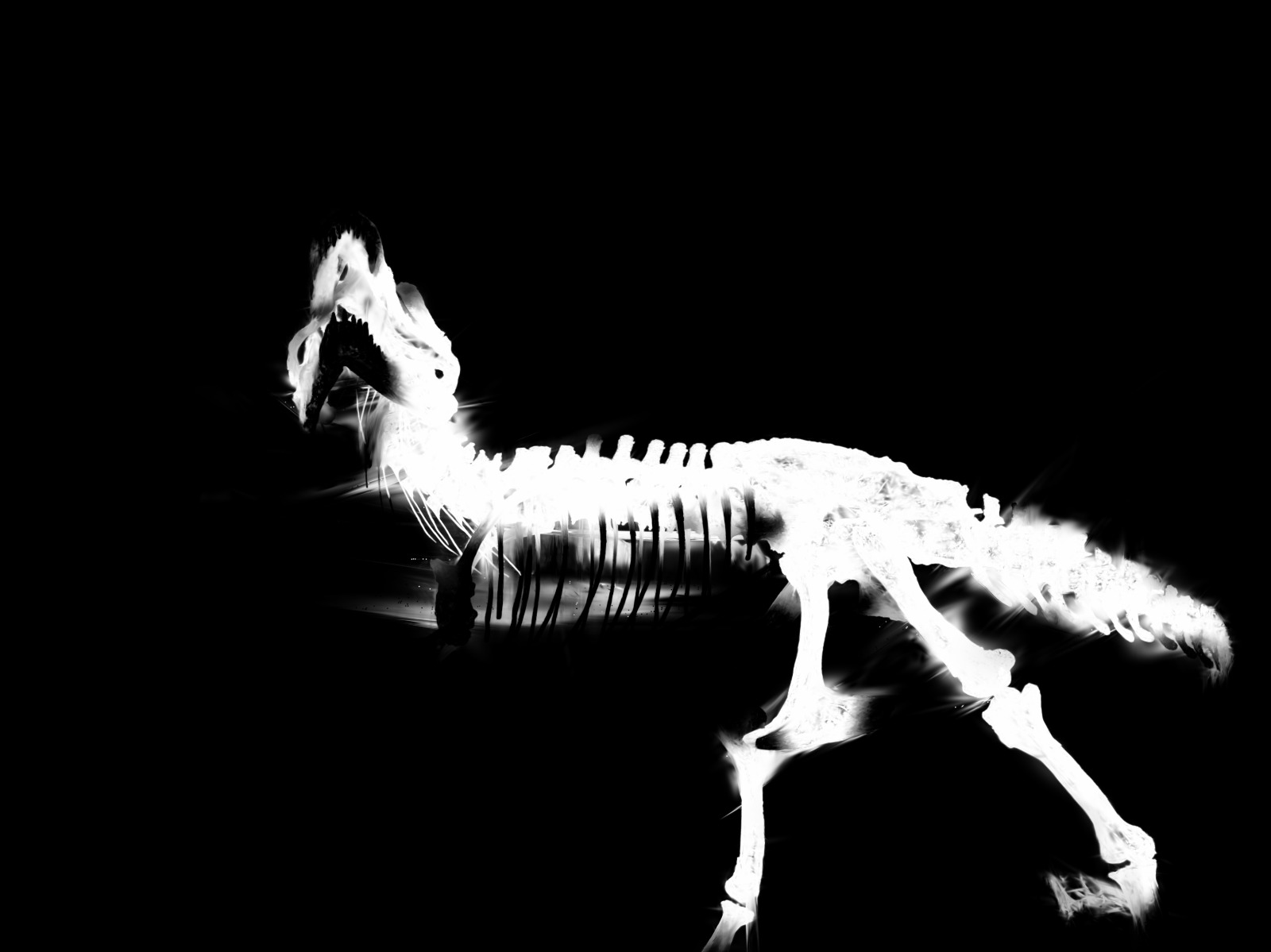} \\
    \includegraphics[width=0.2\linewidth]{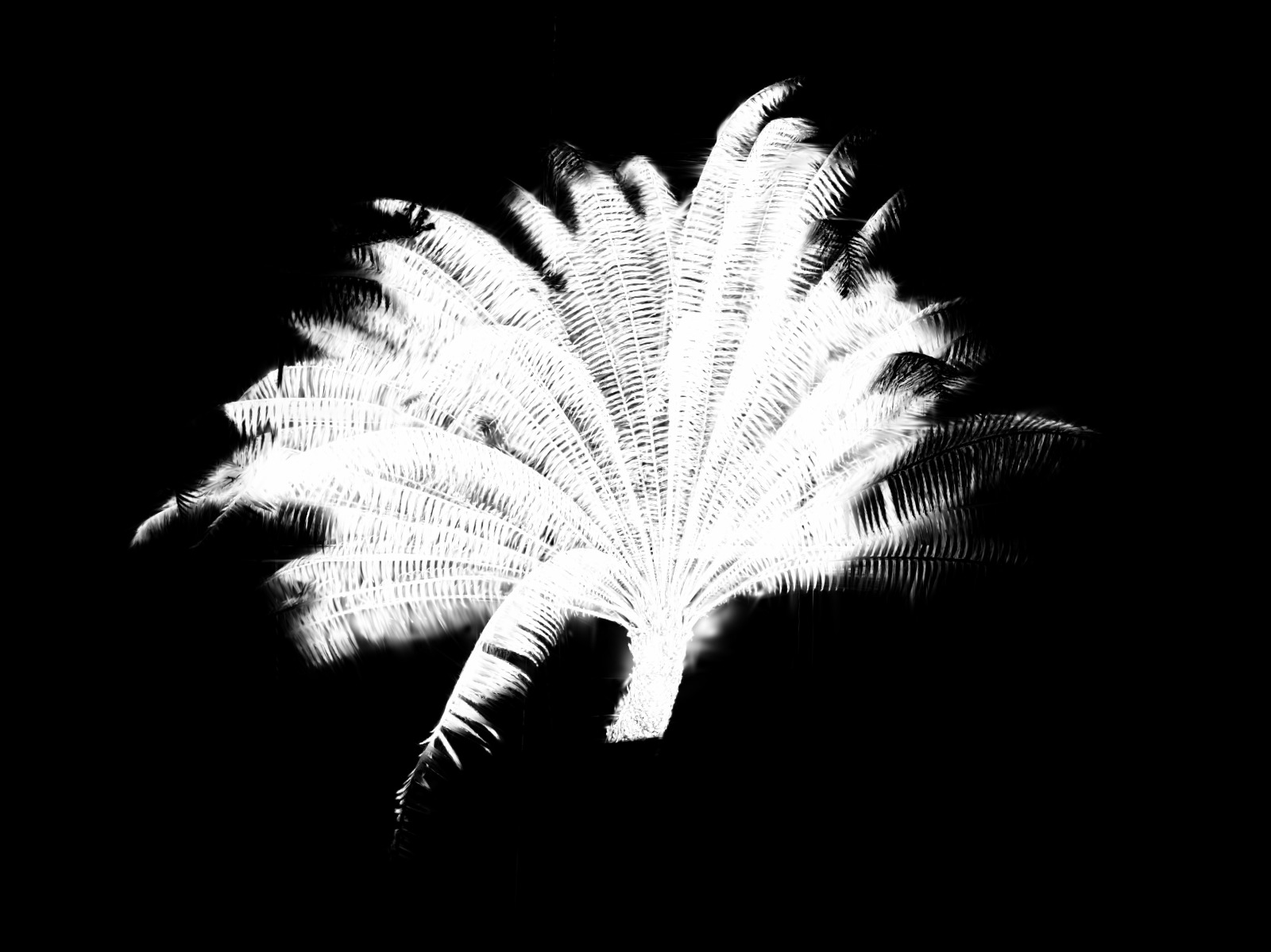}
    \includegraphics[width=0.2\linewidth]{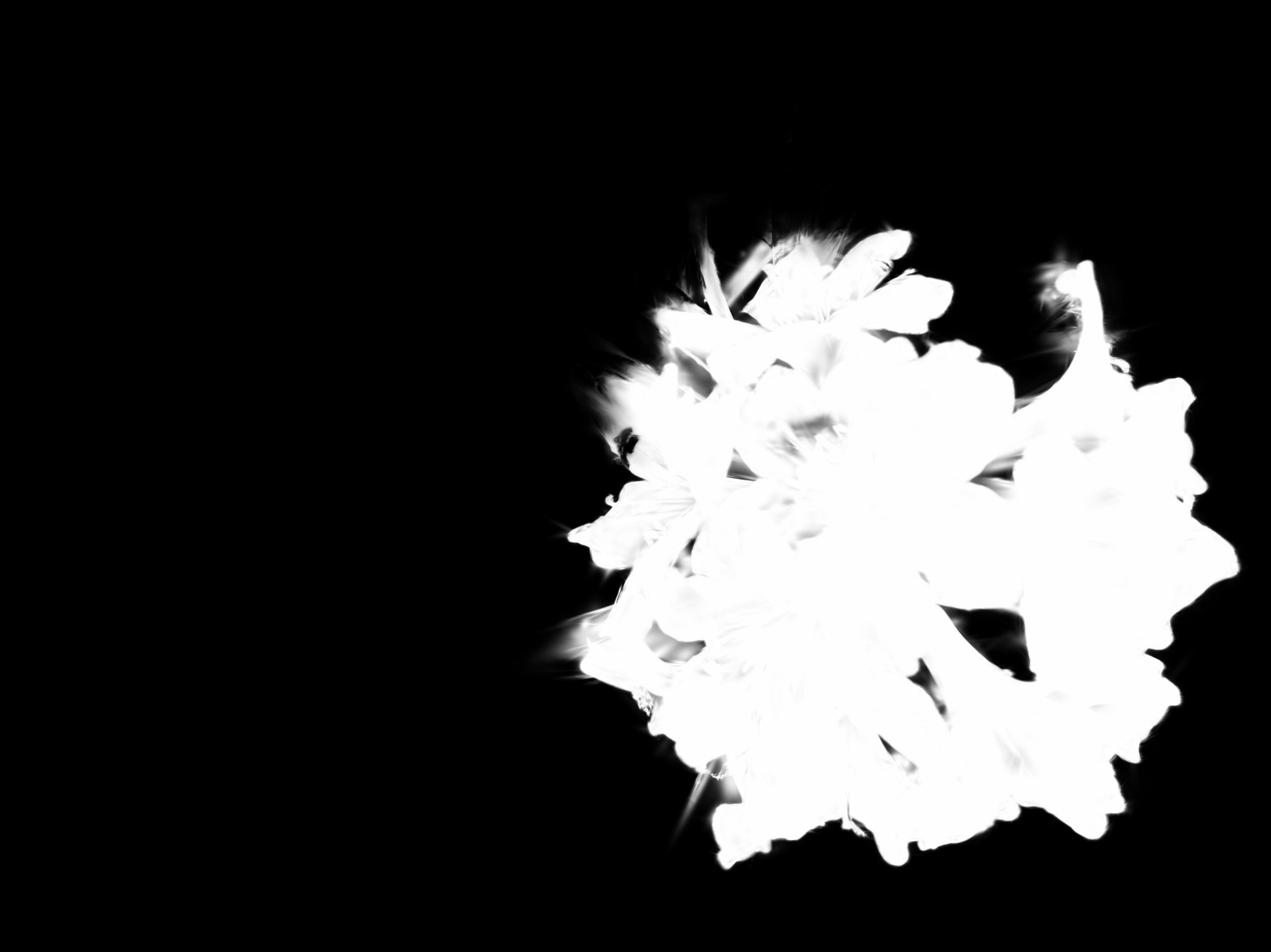}
    \includegraphics[width=0.2\linewidth]{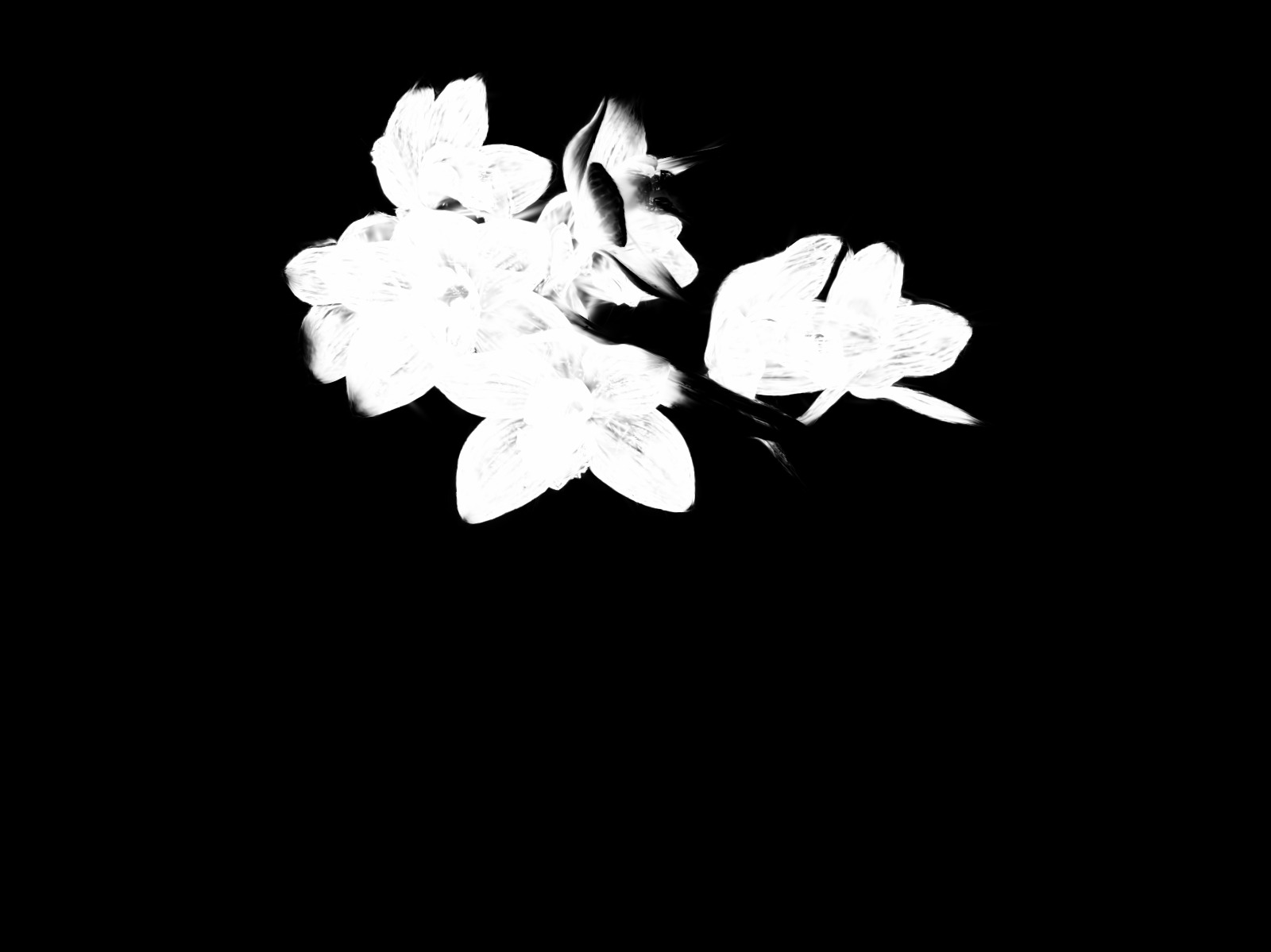}
    \includegraphics[width=0.2\linewidth]{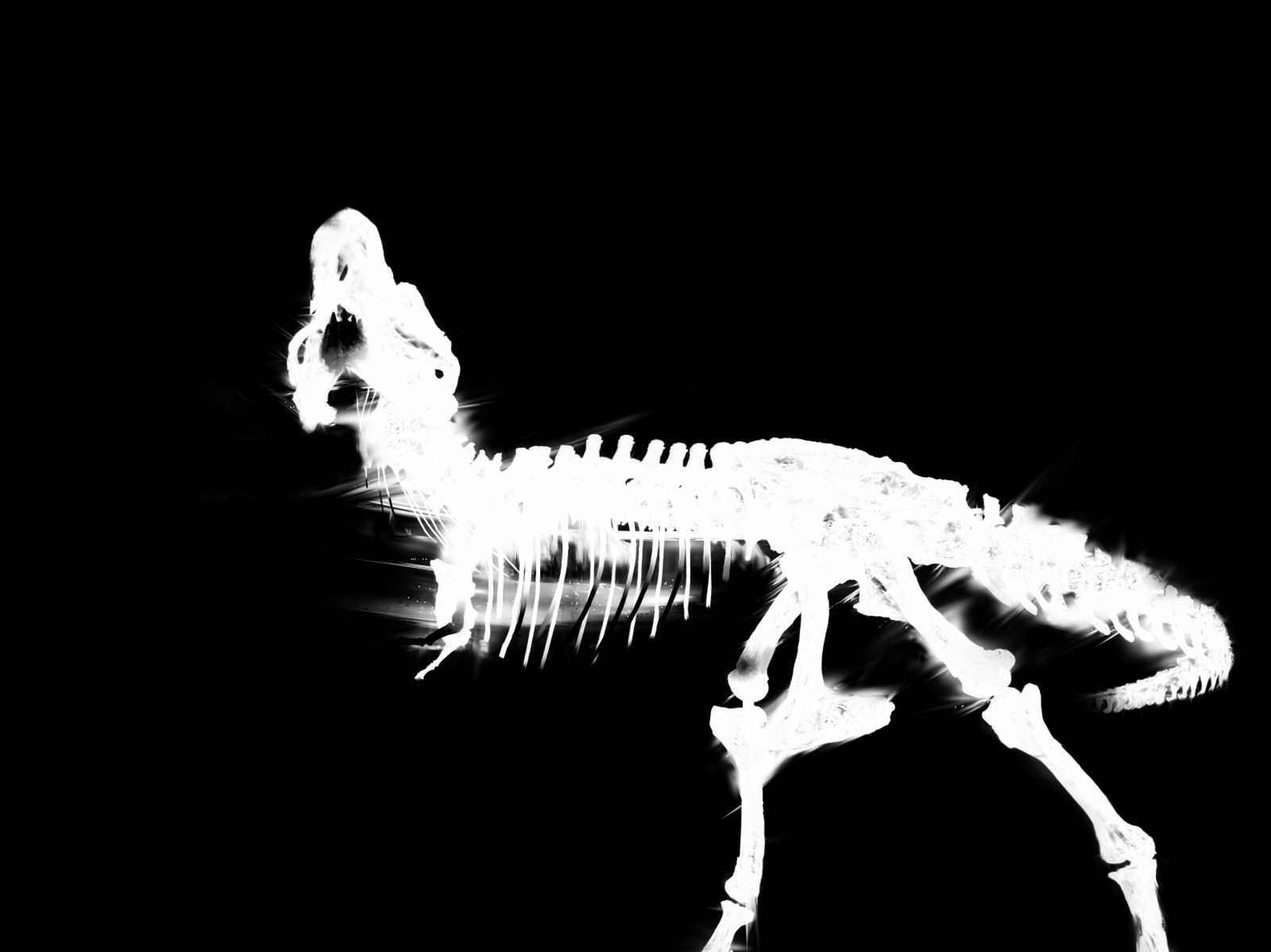} \\
    \includegraphics[width=0.2\linewidth]{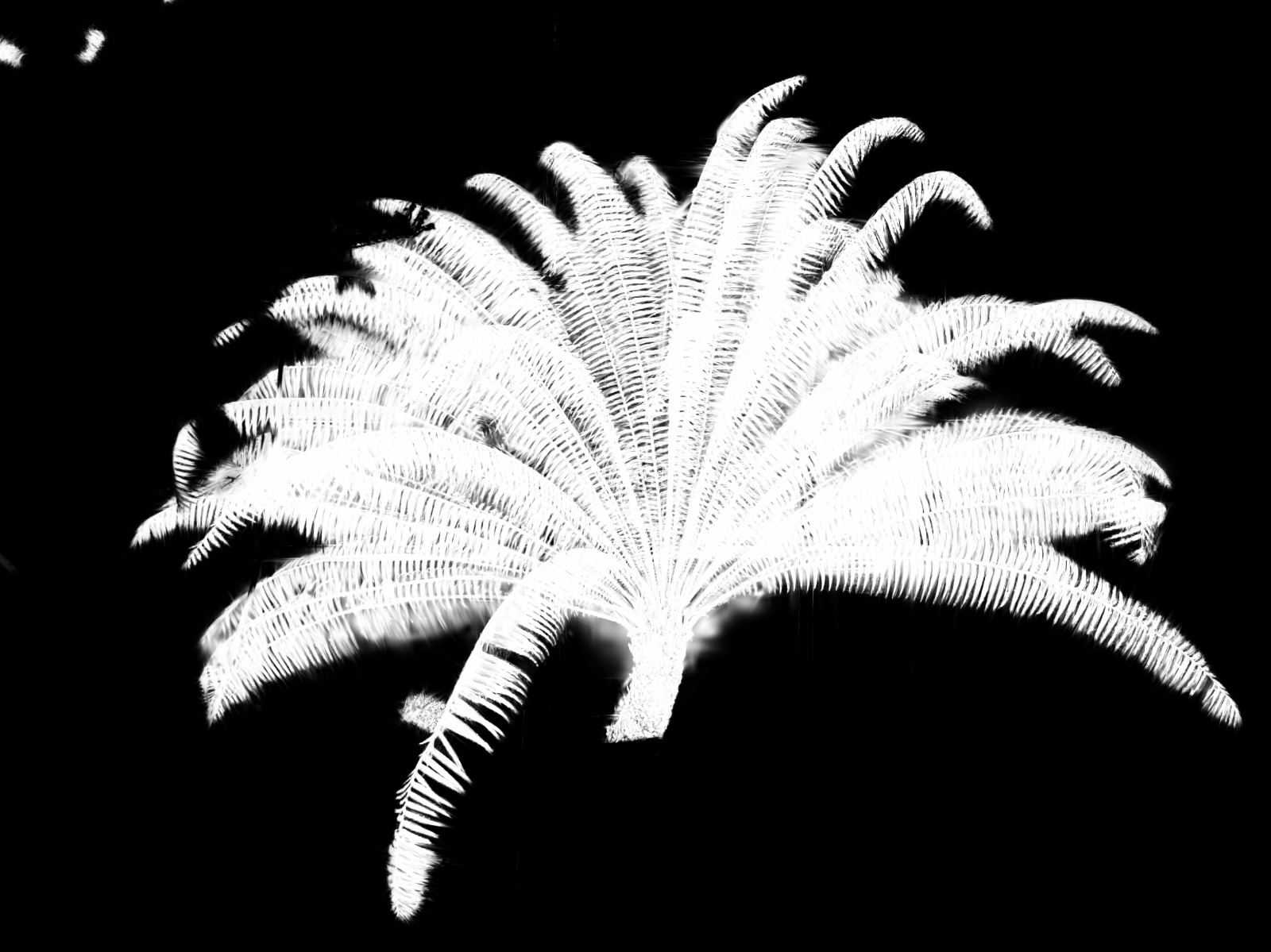}
    \includegraphics[width=0.2\linewidth]{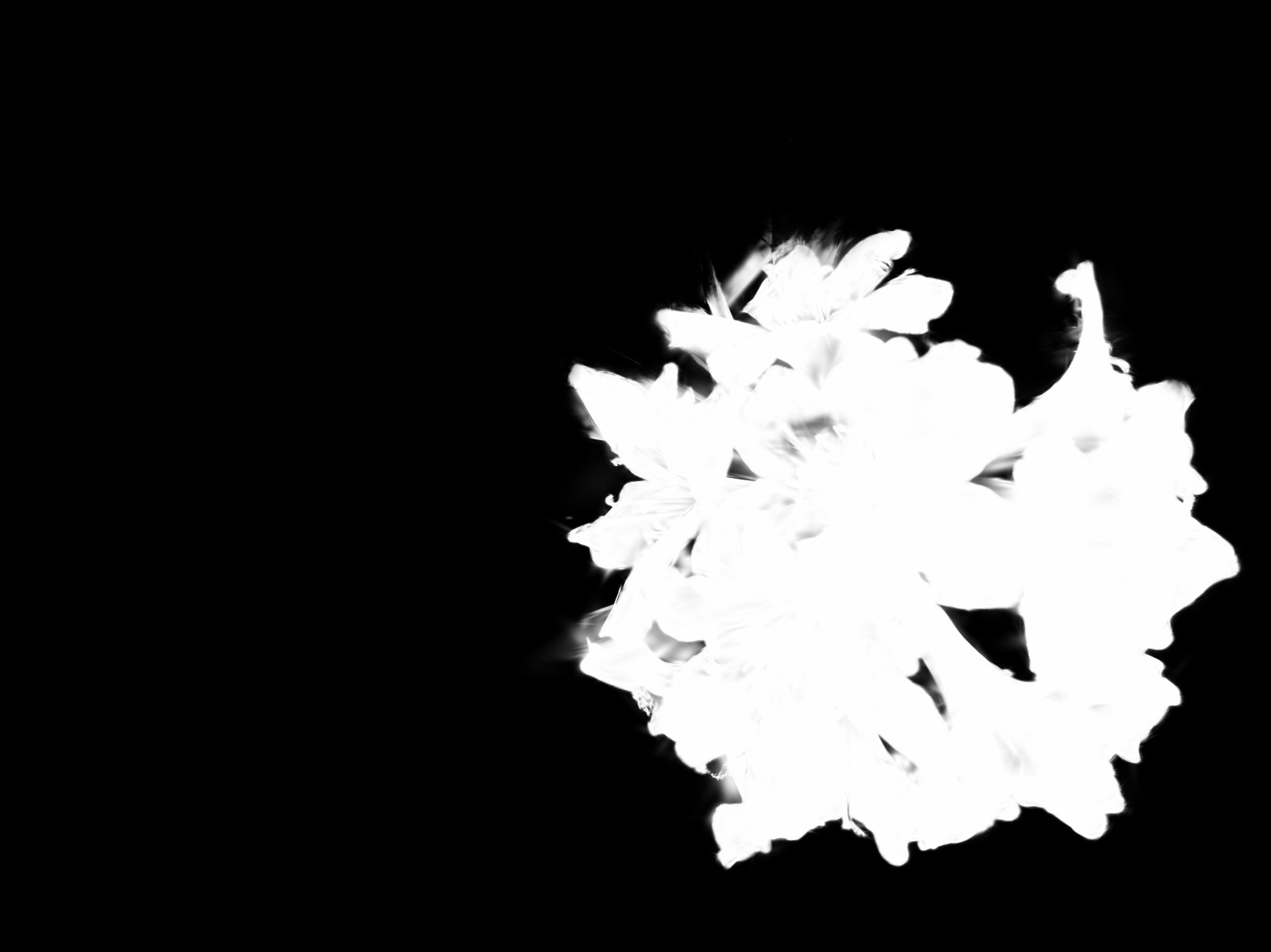}
    \includegraphics[width=0.2\linewidth]{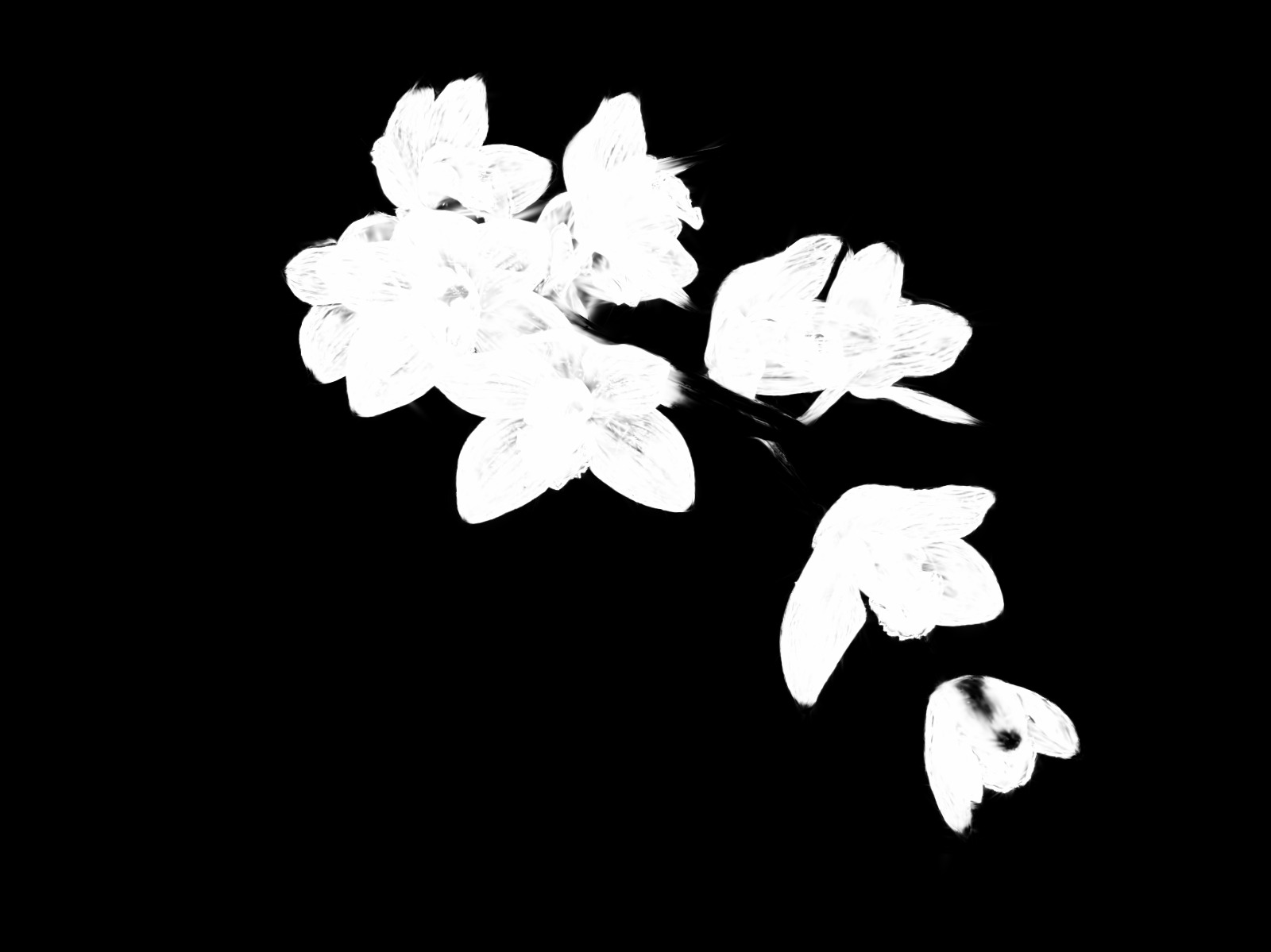}
    \includegraphics[width=0.2\linewidth]{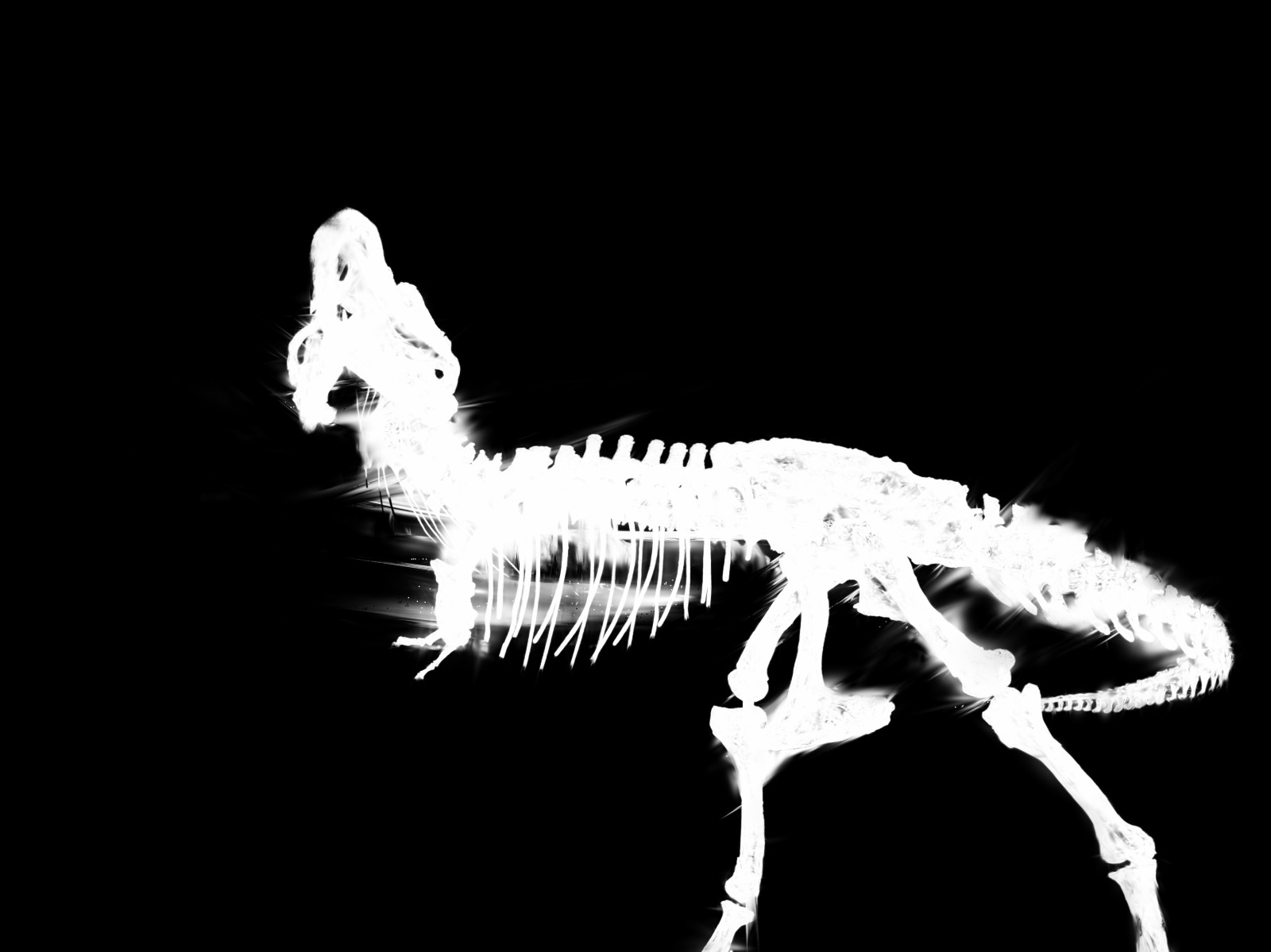} \\
    \includegraphics[width=0.2\linewidth]{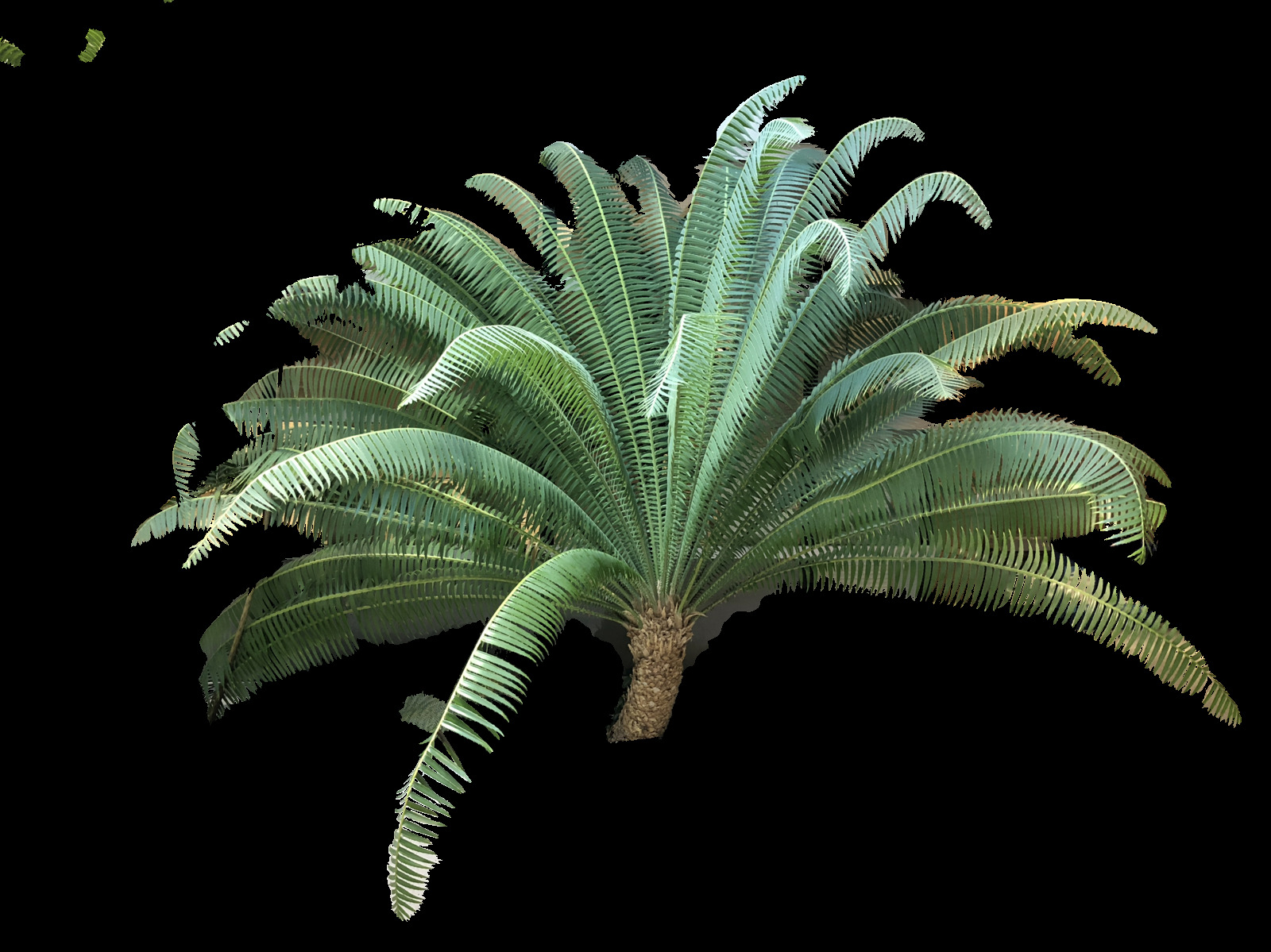}
    \includegraphics[width=0.2\linewidth]{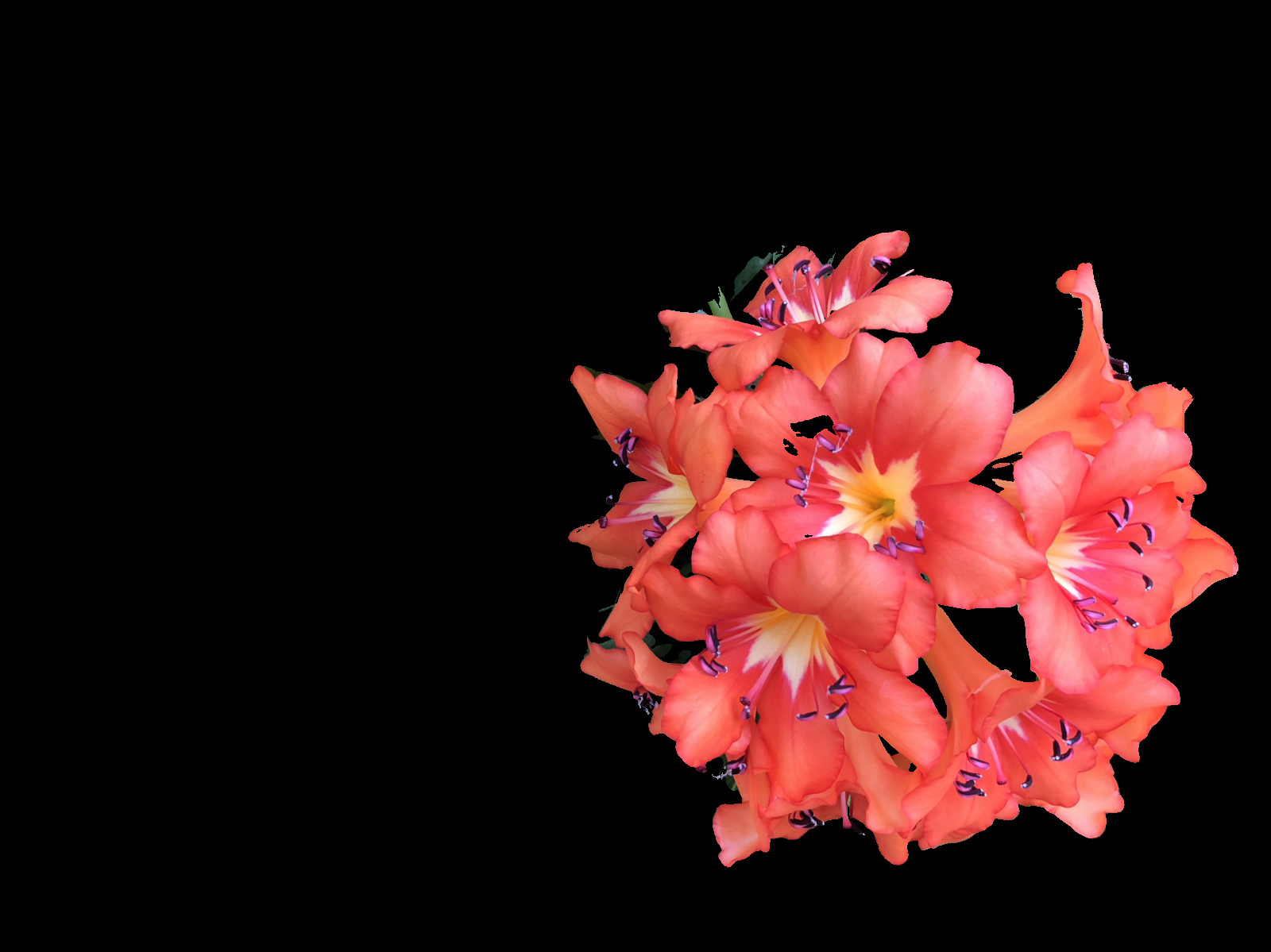}
    \includegraphics[width=0.2\linewidth]{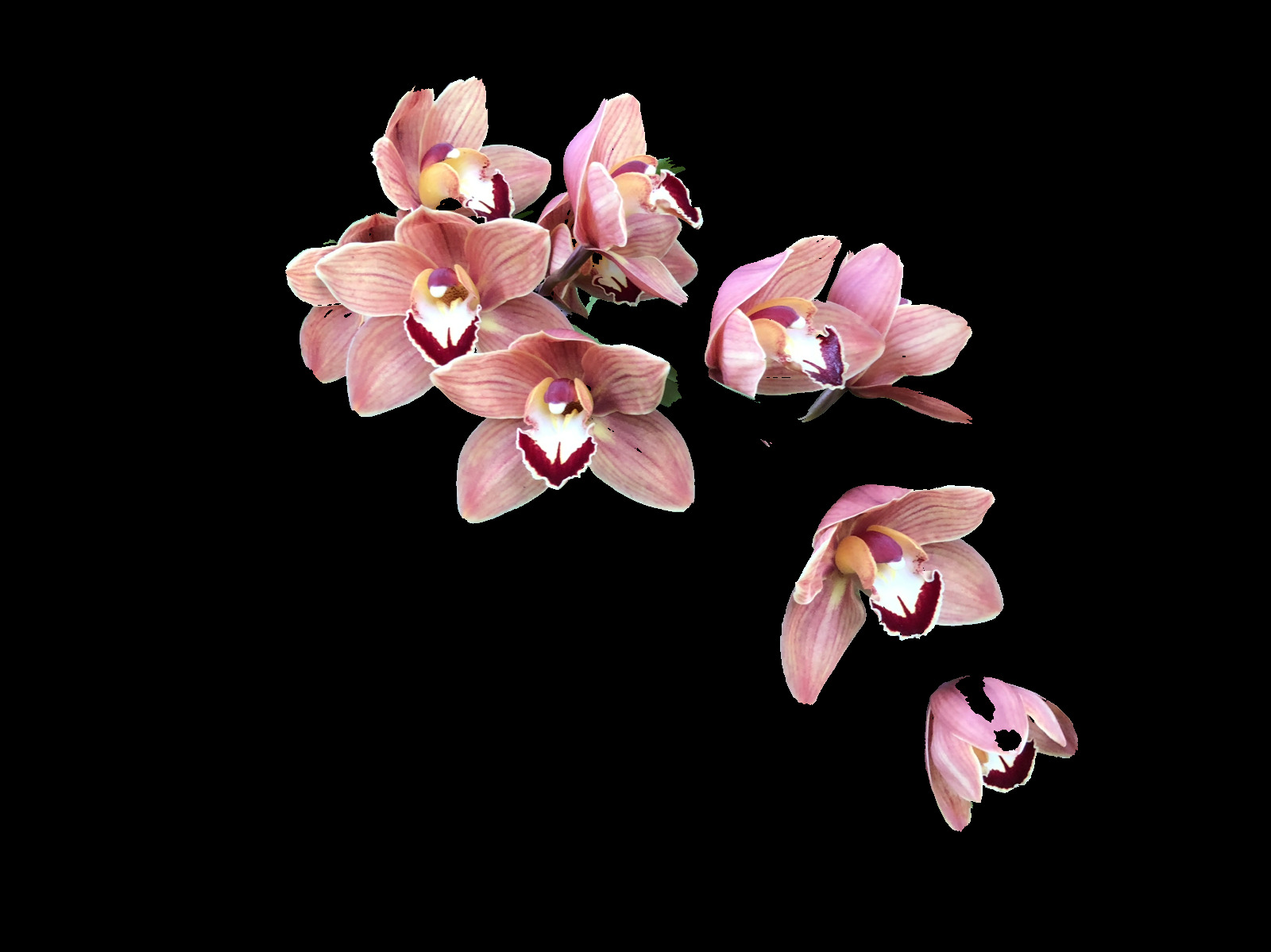}
    \includegraphics[width=0.2\linewidth]{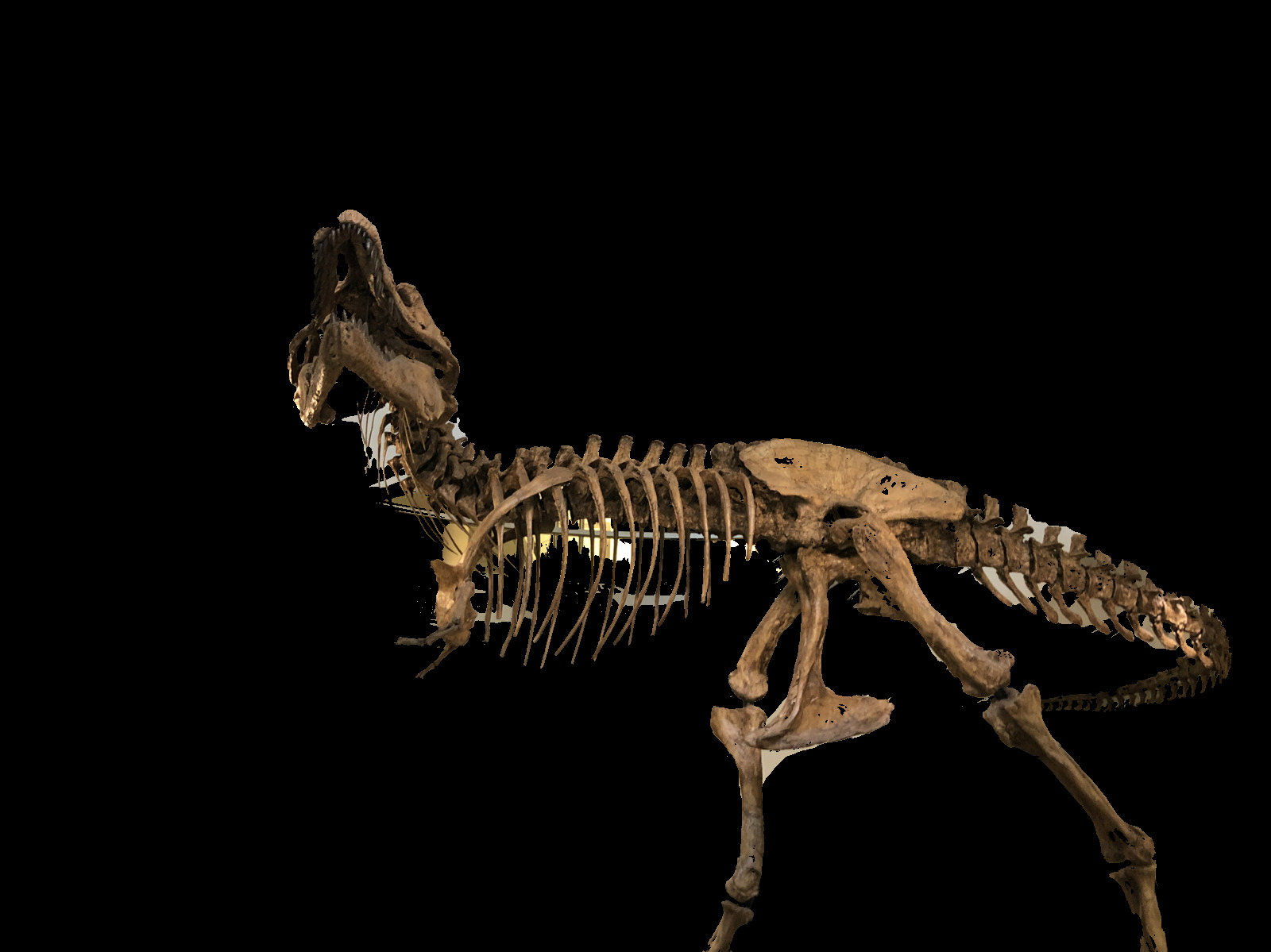} \\
    \caption{\textbf{Illustration of the graph diffusion process.} 2D projections of i) first three PCA components of DINOv2 3D features, ii) unary regularization term (red), iii) weight vector $g_t$ at timesteps $t\in \{0,3,5,10,100\}$, iv) RGB segmentation obtained using a mask based on the 2D projection of $g_{100}$.}
    \label{fig:appendix-diffusion}
\end{figure*}

\myparagraph{Diffusion process}
Fig.~\ref{fig:appendix-diffusion} illustrates different steps of the diffusion process for Fern, Leaves, Flower and Trex from the NVOS~\citep{ren2022nvos} dataset. Starting from the reference scribbles, the diffusion rapidly spreads through the large neighboring Gaussians. Covering the entire object takes more time for complex structures such as Fern, or for masks with disconnected components such as Orchids. As illustrated in the case of Flower, the last diffusion steps allow spreading to the smaller Gaussians on the flowers' edges, yielding a refined segmentation mask. For Trex, the parts being reached the latest are the head and tail. Their features are further away from the reference features (defined as the average feature over 3D reference scribbles), and therefore the regularization for diffusion is stronger in these regions. Overall once the object has been fully covered, the regularization is very effective at preventing leakage, which allows diffusion to run for an arbitrary number of steps.

\begin{figure*}[t!]
    \centering
    \begin{subfigure}{.32\linewidth}
        \includegraphics[width=\linewidth, trim=0 360 0 360, clip]{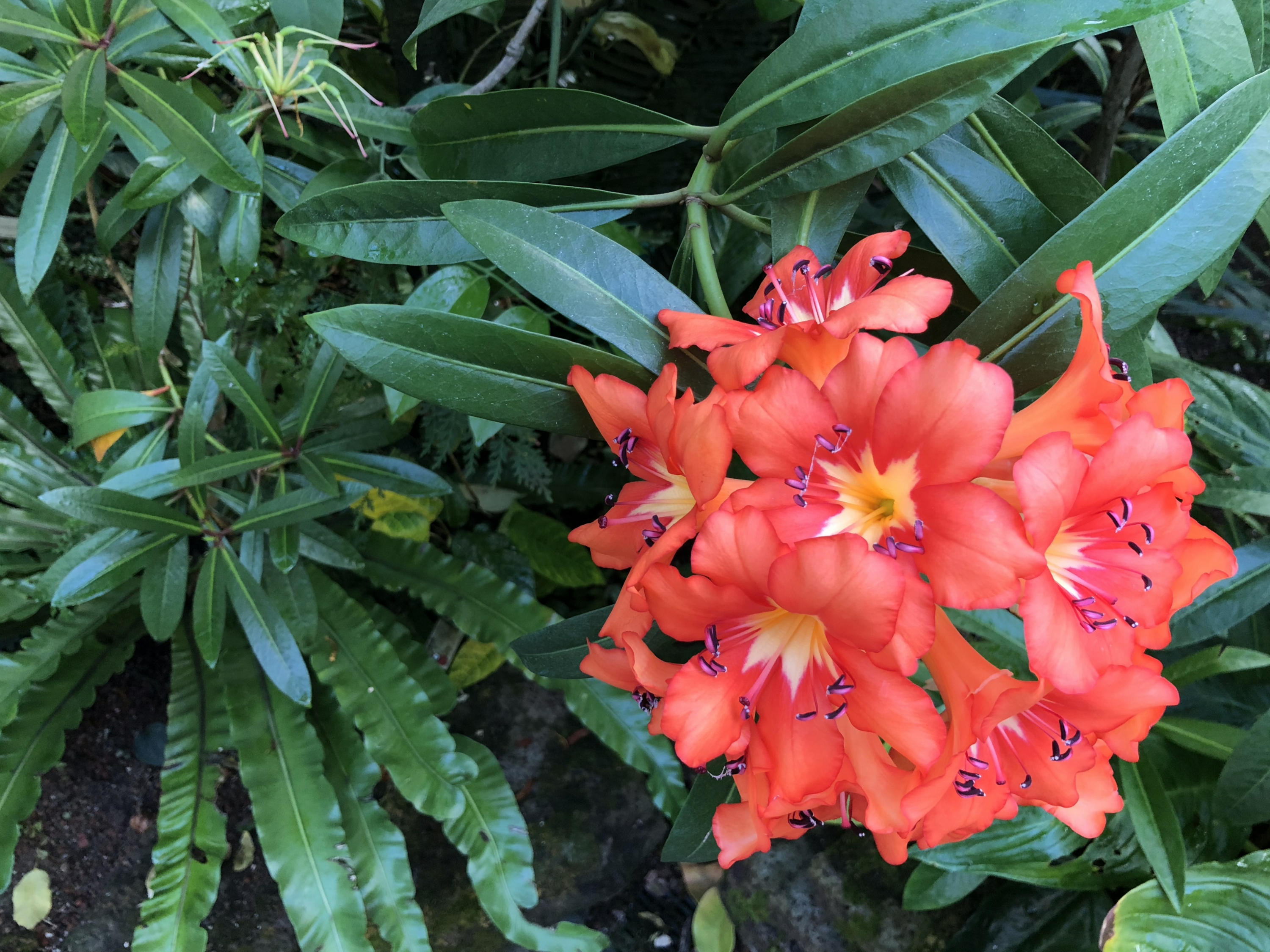} \\
        \includegraphics[width=\linewidth, trim= 0 360 0 360, clip]{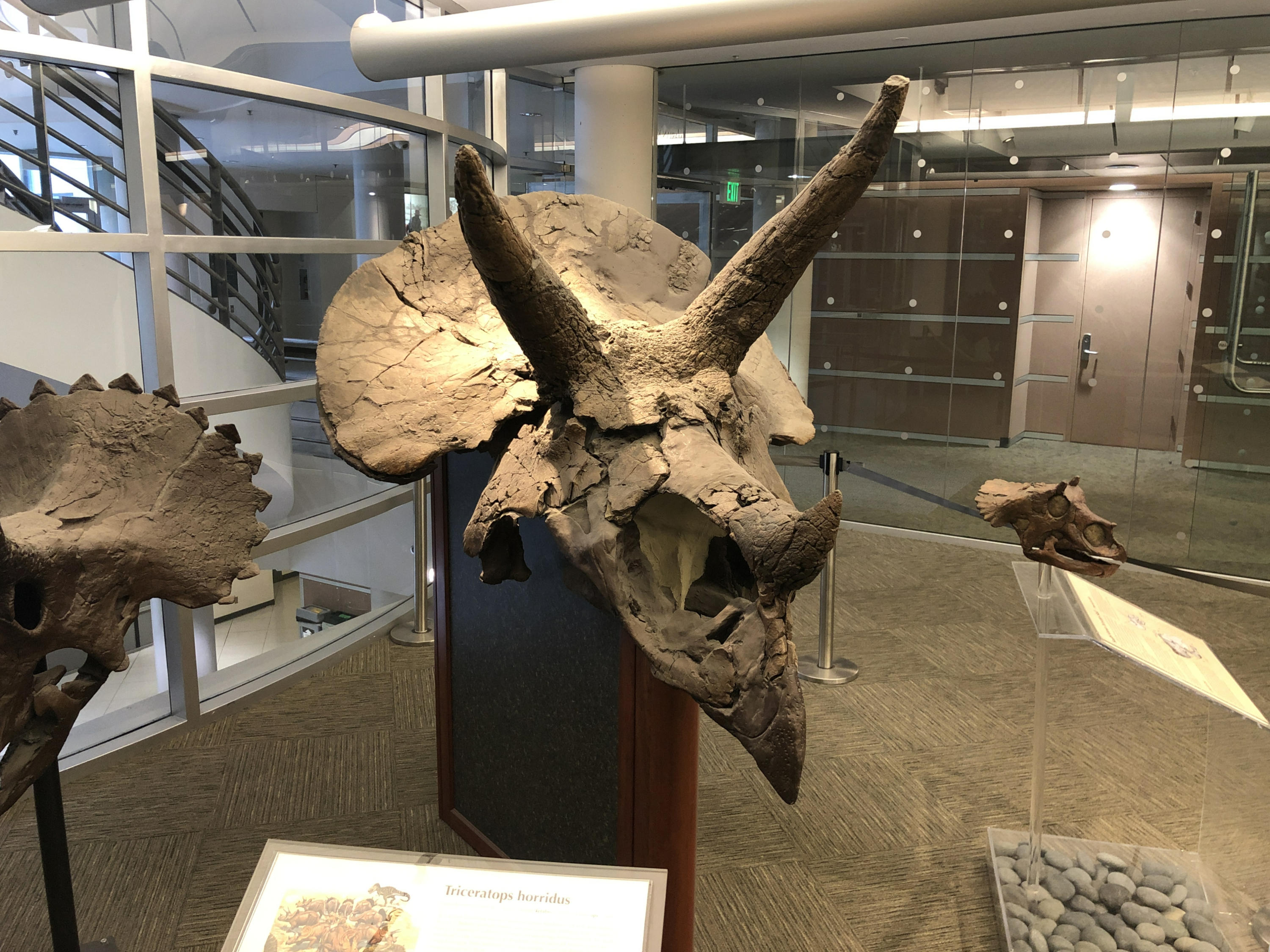} \\
        \caption{RGB image}
    \end{subfigure}
    \begin{subfigure}{.32\linewidth}
        \includegraphics[width=\linewidth]{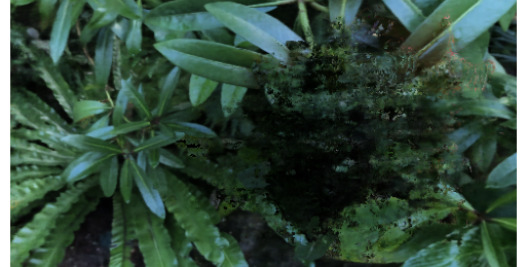} \\
        \includegraphics[width=\linewidth]{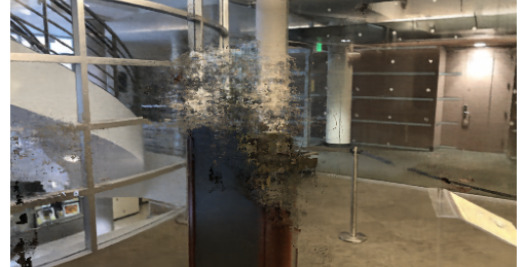} \\
        \caption{N3F~\citep{tschernezki2022n3f}}
    \end{subfigure}
    \begin{subfigure}{.32\linewidth}
        \includegraphics[width=\linewidth, trim=0 195 0 195, clip]{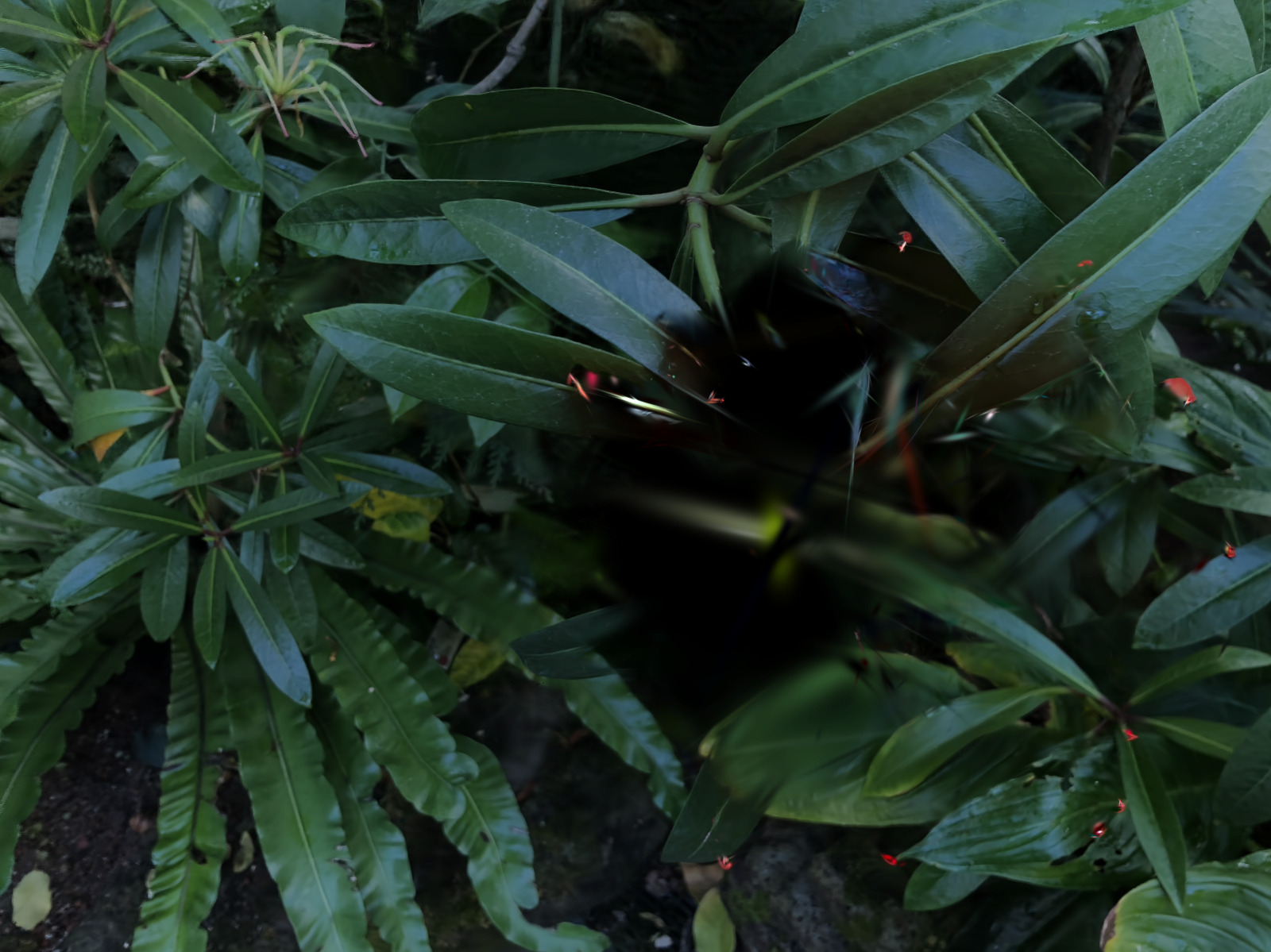} \\
        \includegraphics[width=\linewidth, trim=0 195 0 195, clip]{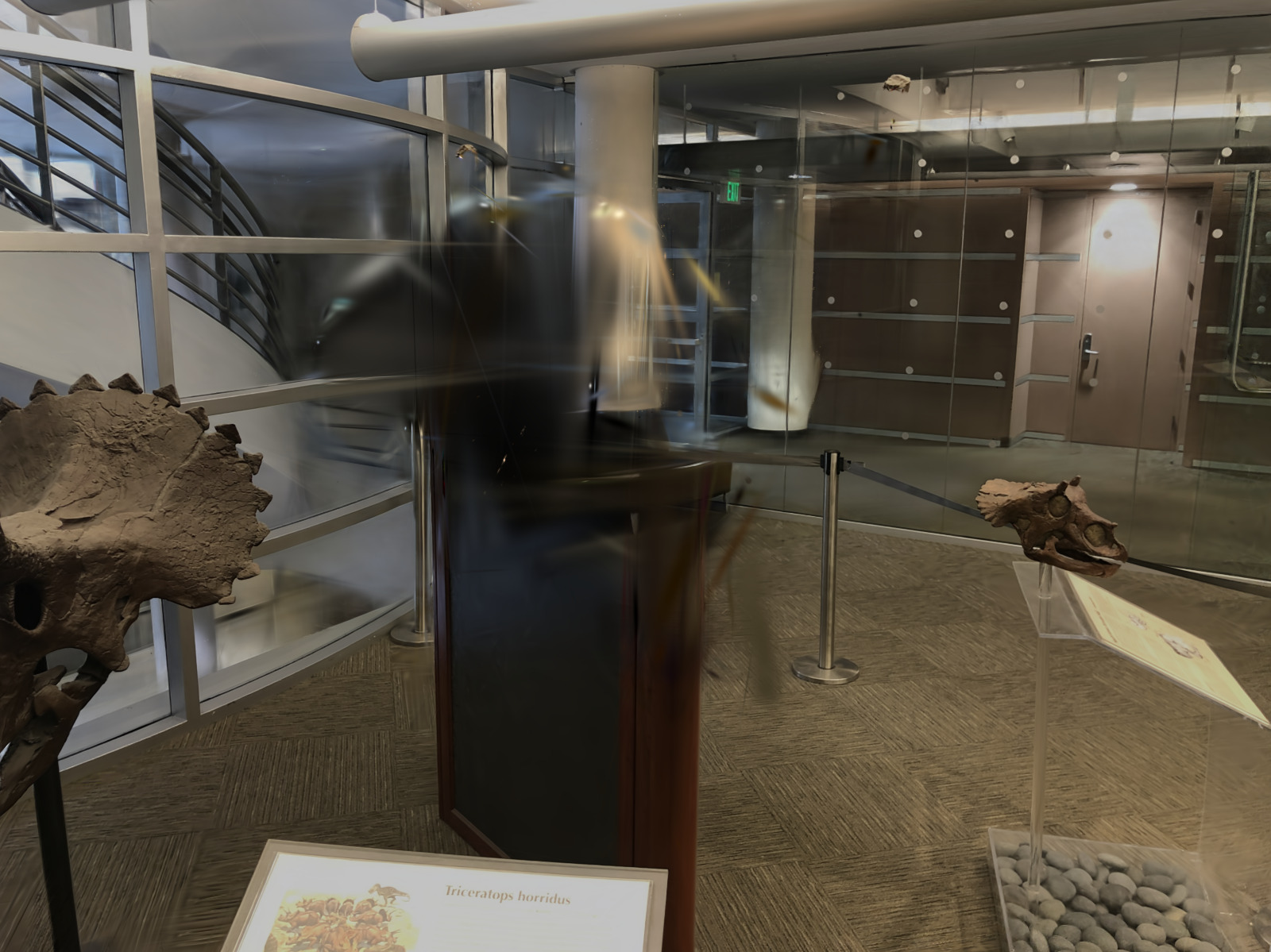} \\
        \caption{LUDVIG (ours)}
    \end{subfigure}
    \caption{\textbf{Object removal.} 3D segmentation, removal and rendering for LUDVIG and N3F~\citep{tschernezki2022n3f}. For N3F, figures are taken from~\citep{tschernezki2022n3f}.}
    \label{fig:appendix-object-removal}
\end{figure*}

\myparagraph{Object removal} Fig.~\ref{fig:appendix-object-removal} shows comparative visualizations of object removal with N3F~\citep{tschernezki2022n3f} and LUDVIG. For rendering the edited RGB image, N3F sets to zero the occupancy for all 3D points belonging to the object. For LUDVIG, we remove all Gaussians pertaining to the 3D semantic mask resulting from graph diffusion. We observe that the regions behind to segmented object are much smoother for LUDVIG than for N3F. Regions unseen from any viewpoint are black for LUDVIG (no gaussians) and result in a background partially hallucinated by NeRF for N3F.  

\subsection{Visual comparisons of uplifted features}

\begin{figure*}[t!]
    \centering
    \begin{subfigure}{.9\linewidth}
        \includegraphics[width=.24\linewidth]{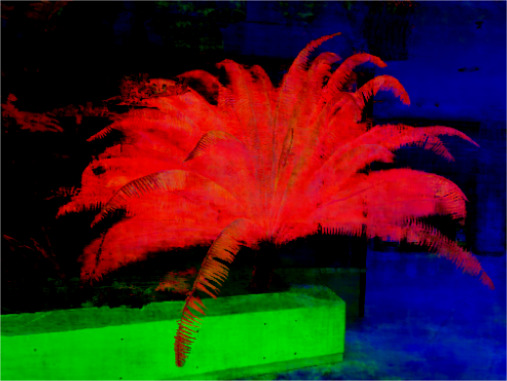}
        \includegraphics[width=.24\linewidth]{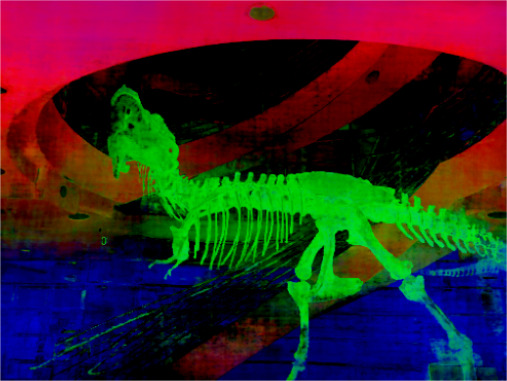}
        \includegraphics[width=.24\linewidth]{figures/editing/horns.jpg} 
        \includegraphics[width=.24\linewidth]{figures/editing/flower.jpg} 
    \end{subfigure}
    \begin{subfigure}{.9\linewidth}
        \includegraphics[width=.24\linewidth]{figures/n3f/fern.jpg}
        \includegraphics[width=.24\linewidth]{figures/n3f/trex.jpg}
        \includegraphics[width=.24\linewidth]{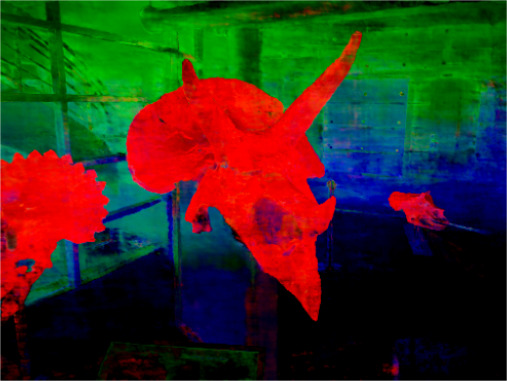} 
        \includegraphics[width=.24\linewidth]{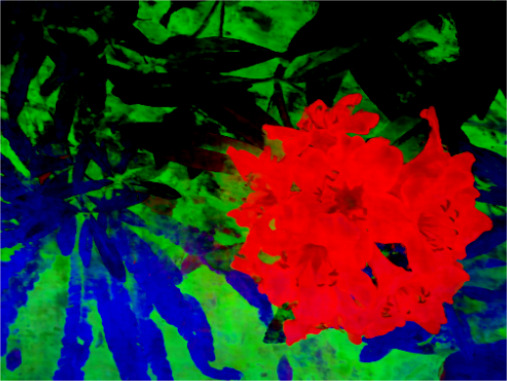} 
    \end{subfigure}
    \begin{subfigure}{.9\linewidth}
        \includegraphics[width=.24\linewidth]{figures/diffusion_appendix/fern_pca.jpg} 
        \includegraphics[width=.24\linewidth]{figures/pcas/trex_pca.jpg} 
        \includegraphics[width=.24\linewidth]{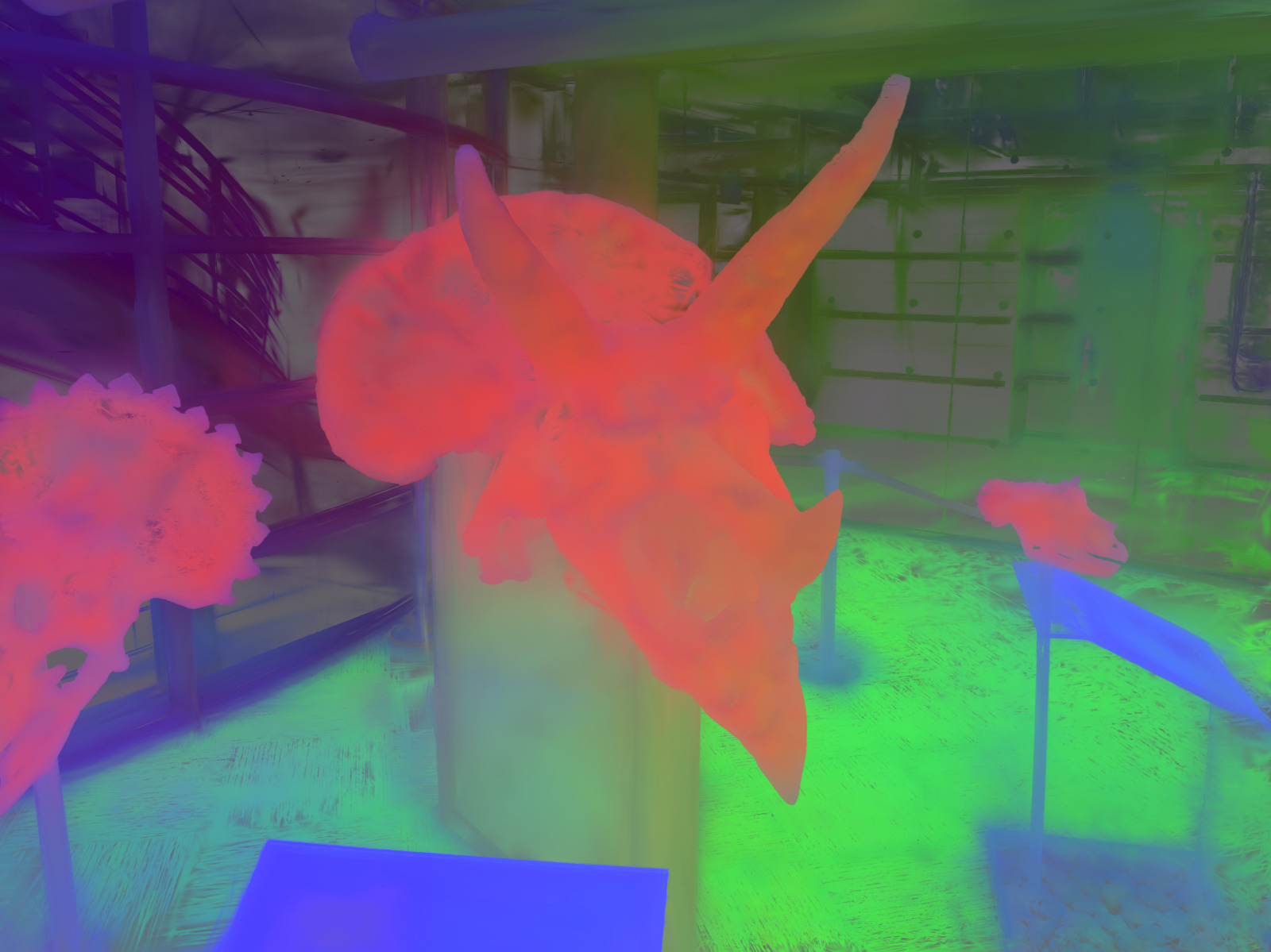}
        \includegraphics[width=.24\linewidth]{figures/diffusion_appendix/flower_pca.jpg}
    \end{subfigure}
    \caption{Comparison between LUDVIG's uplifted DINOv2 features (bottom) and N3F's~\citep{tschernezki2022n3f}
    learned DINO features (top). For N3F, figures are taken from ~\citep{tschernezki2022n3f}.}
    \label{fig:appendix-dff}
\end{figure*}

Fig.~\ref{fig:appendix-dff} show a comparison of LUDVIG's 3D DINOv2 features with learned 3D DINO features of N3D~\citep{tschernezki2022n3f}. Their figures are taken from their work. Compared to N3D, LUDVIG produces a more fine-grained reconstruction of the background (notably in the trex and horns scenes) and smoother features across all scenes.

\subsection{3D features for ScanNet semantic segmentation}

\begin{figure*}[t!]
    \centering
    \includegraphics[width=0.32\linewidth]{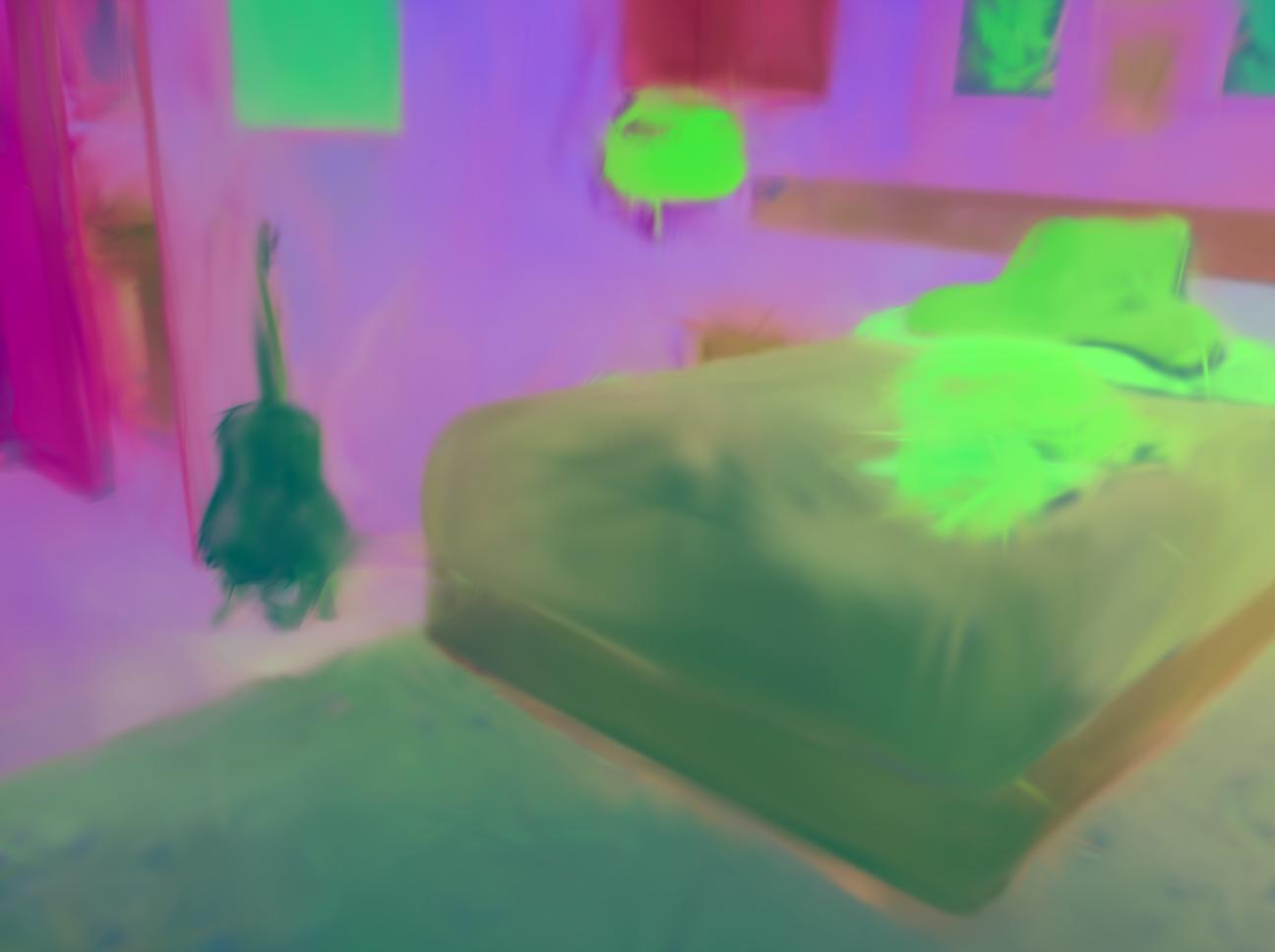}
    \includegraphics[width=0.32\linewidth]{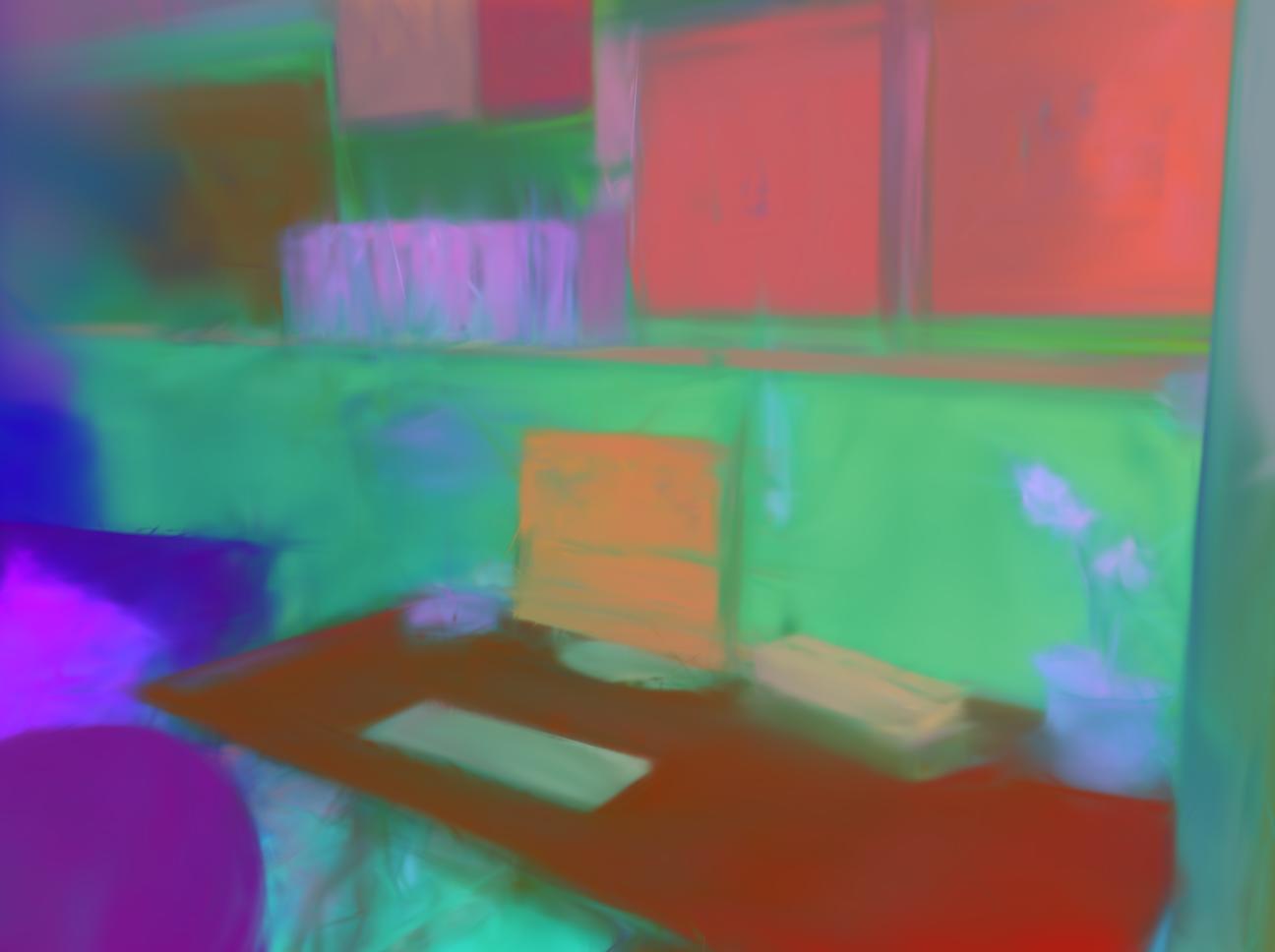}
    \includegraphics[width=0.32\linewidth]{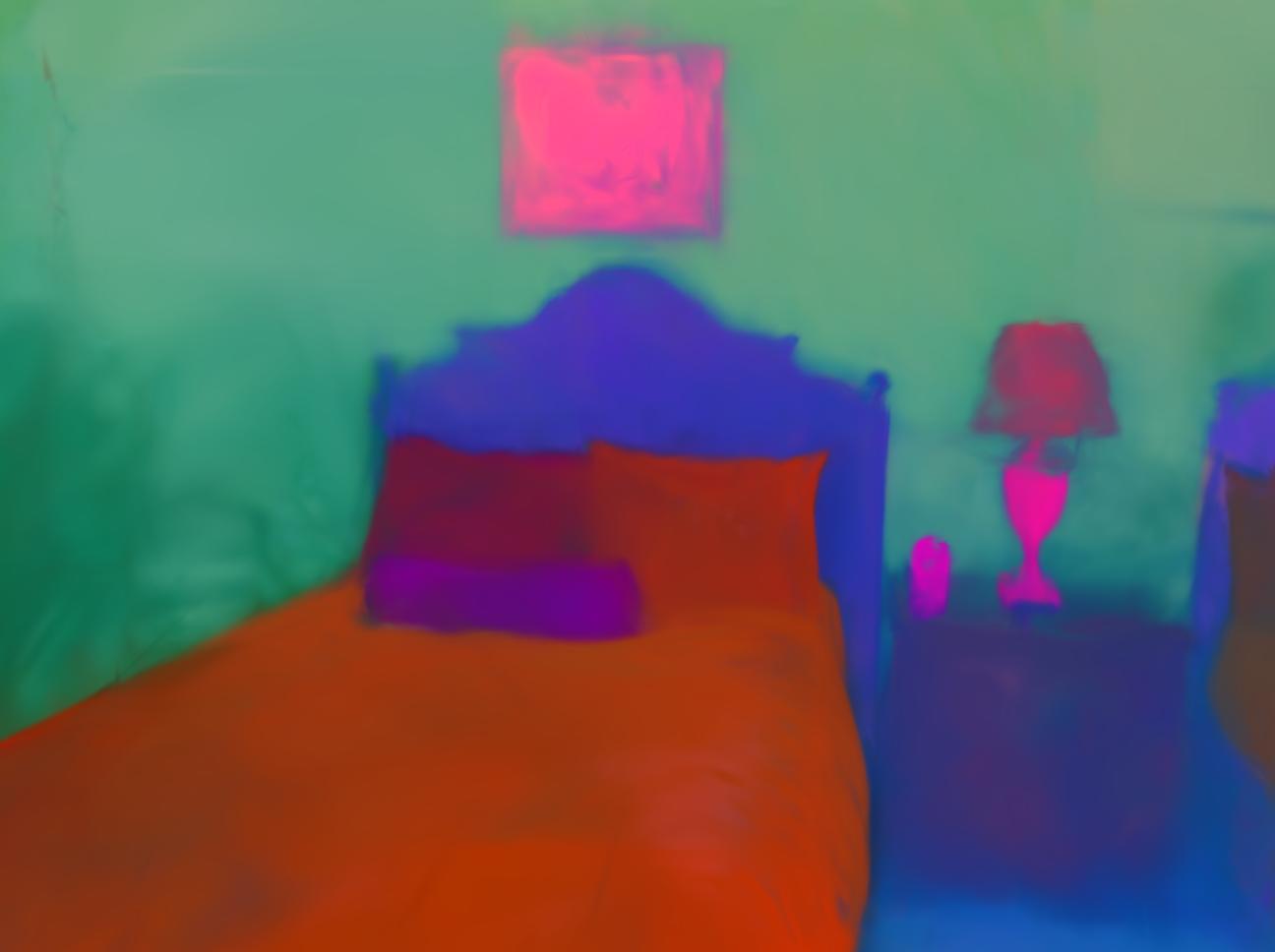}
    \caption{PCA of uplifted semantic features on ScanNetv2, obtained by assigning object-level CLIP features to SAM segments.}
    \label{fig:appendix-semantic}
\end{figure*}

Fig.~\ref{fig:appendix-semantic} presents visualizations of 3D semantic features obtained by uplifting 2D feature maps generated by OpenGaussian~\citep{wu2024opengaussian}. Following the approach of LangSplat~\citep{qin2023langsplat}, the 2D features are computed by assigning object-level CLIP embeddings to segmentation masks produced by SAM in \emph{everything} mode.

Surprisingly, despite the very constrained training conditions used during Gaussian Splatting reconstruction (frozen positions and disabled densification process), the uplifting still yields coherent and meaningful 3D semantic features.

\subsection{Comparison to GaussianEditor's uplifting}
\label{sec:appendx-gaussianeditor}

\begin{figure*}[t!]
    \centering
    \begin{subfigure}{.24\linewidth}
        \includegraphics[width=\linewidth]{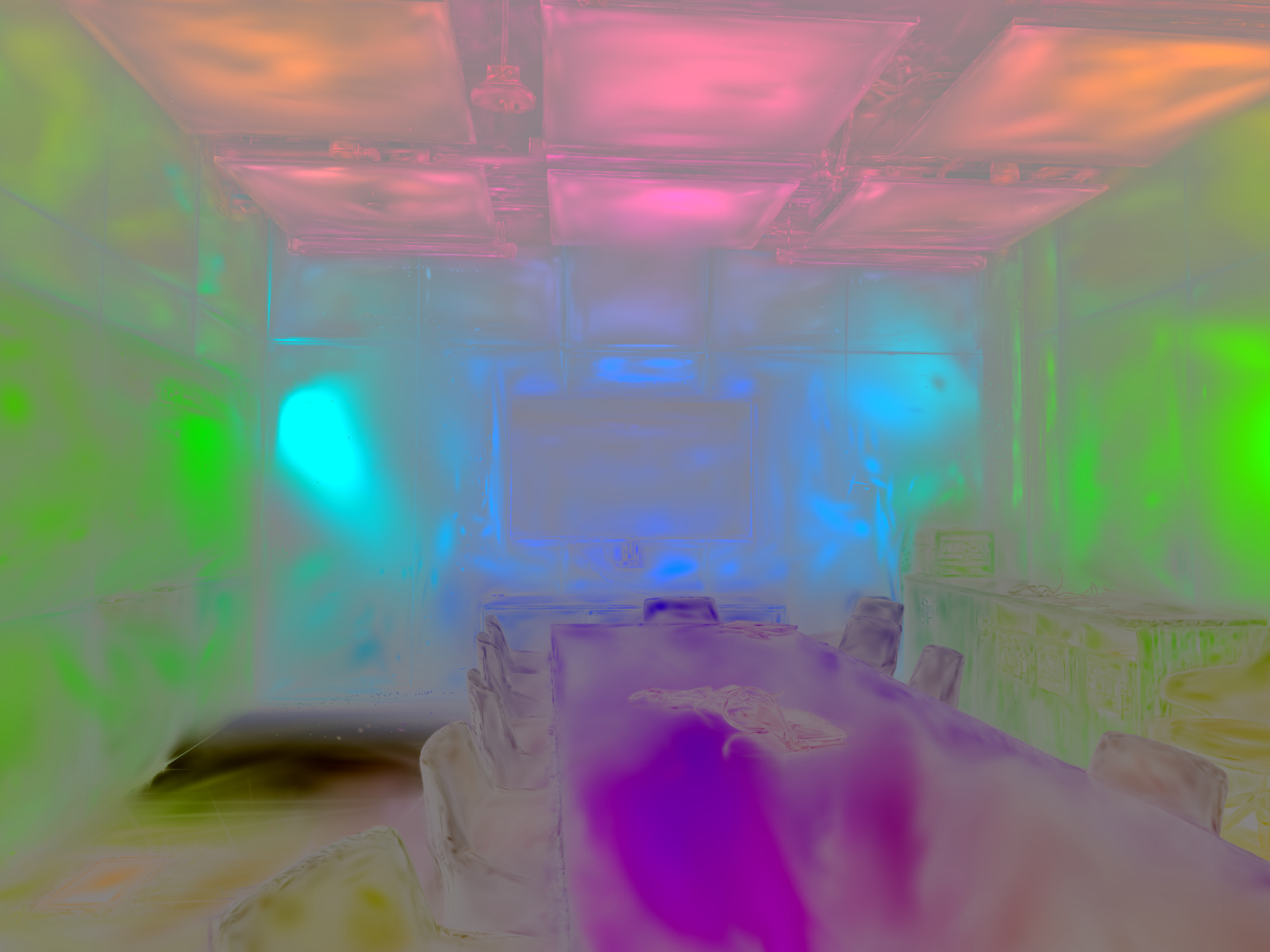}
        \caption{GaussianEditor}
    \end{subfigure}
    \begin{subfigure}{.24\linewidth}
        \includegraphics[width=\linewidth]{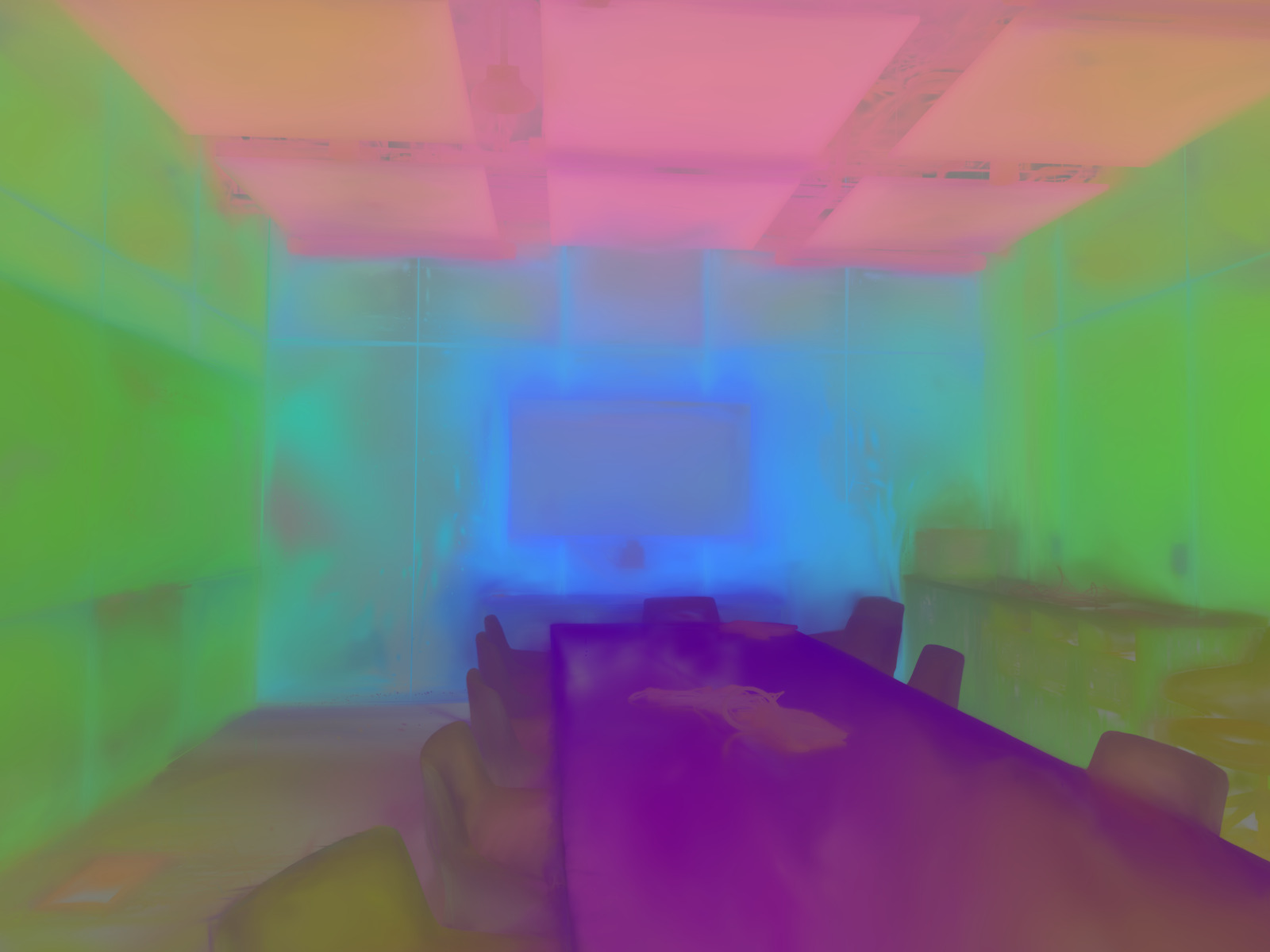}
        \caption{LUDVIG}
    \end{subfigure}
    \begin{subfigure}{.24\linewidth}
        \includegraphics[width=\linewidth]{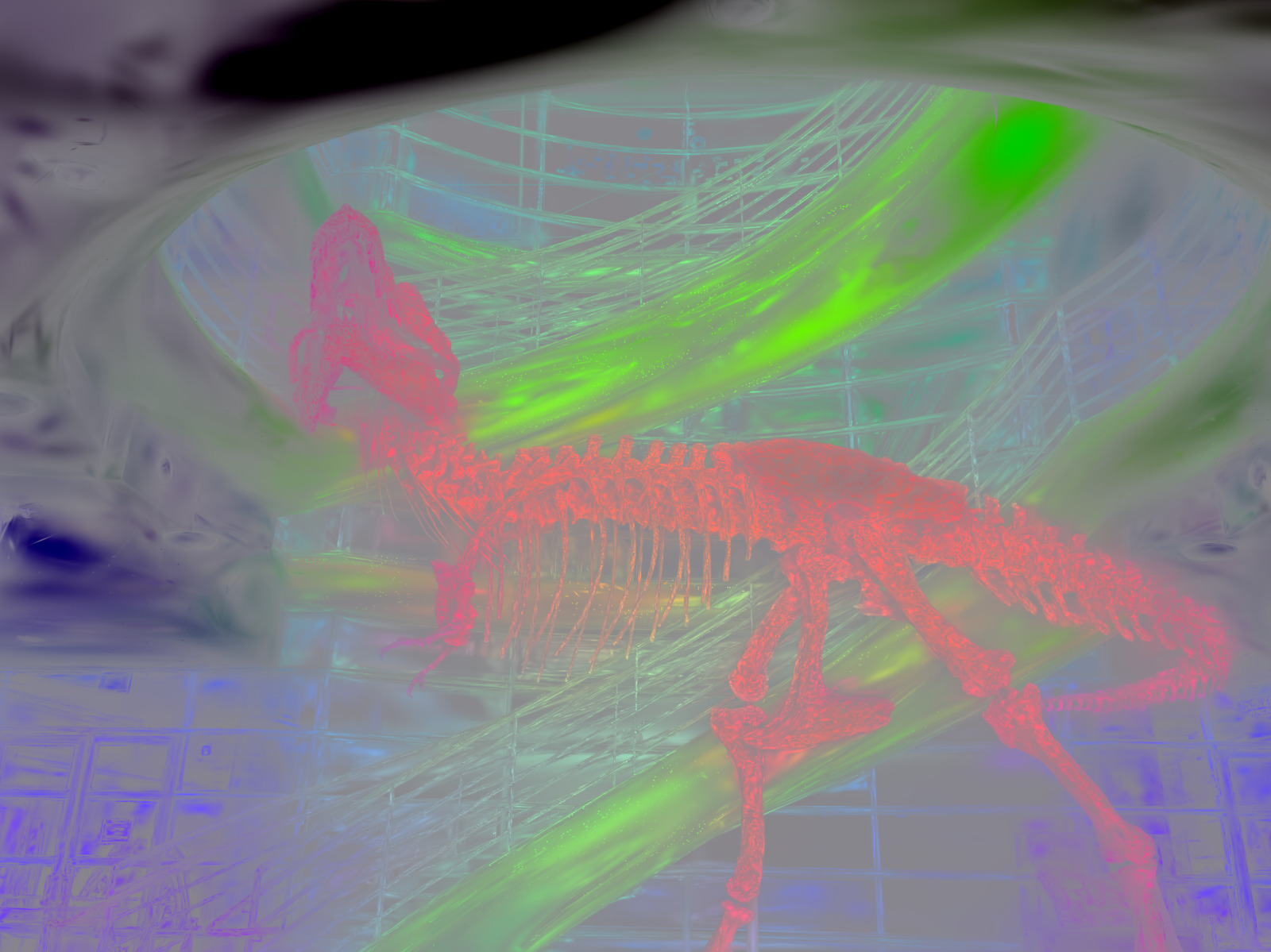}
        \caption{GaussianEditor}
    \end{subfigure}
    \begin{subfigure}{.24\linewidth}
        \includegraphics[width=\linewidth]{figures/pcas/trex_pca.jpg}
        \caption{LUDVIG}
    \end{subfigure}
    \caption{\textbf{Comparison to GaussianEditors's 
    uplifting.} Comparison of PCA visualization of uplifted features between LUDVIG's and GaussianEditor's aggregation~\citep{chen2024gaussianeditor}.}
    \label{fig:appendix-gaussian-editor}
\end{figure*}

Our aggregation procedure in Eq. (\ref{eq:uplifting}) from the main paper, illustrated in Fig.~\ref{fig:uplift_render}, 
bears similarity with the one from \cite{chen2024gaussianeditor} for uplifting 2D binary masks to a 3D Gaussian splatting scene. 
In their method, uplifted masks are thresholded to create 3D binary masks that are used for semantic tracing. Specifically, they rely on rough 3D segmentation masks to selectively optimize Gaussians that are relevant for an editing task. 
Unlike in Eq.~(\ref{eq:uplifting}) and (\ref{eq:uplifting_matrix}) from the main paper,
\cite{chen2024gaussianeditor} propose to normalize their uplifted masks based on the total count of view/pixel pairs $(d,p)$ contributing to the mask of a Gaussian $i$, i.e. $\sum_{d,p\in \mathcal{S}_i} 1$, without taking the rendering weight $w_i(d,p)$ into account. Consequently, the uplifted features tend to have larger values for large, opaque Gaussians.
Fig.~\ref{fig:appendix-gaussian-editor} shows a qualitative comparison between 3D DINOv2 features obtained using the aggregation proposed by \cite{chen2024gaussianeditor} and our approach. The aggregation by \cite{chen2024gaussianeditor} fails to assign the right semantics to large gaussians, which is particularly visible in scenes with high specularity such as Room. This showcases the importance of defining 3D features as \emph{convex combinations} of 2D pixel features.

\subsection{Visualization of CLIP segmentation results}
\label{sec:appendx-clip-visualization}

In this section, we present illustrations of the impact of the diffusion process (Fig.~\ref{fig:appendix-ablation-relevancy}), and comparative visualizations of localization heatmaps with LangSplat and LERF (Fig.~\ref{fig:appendix-clip-heatmaps}).

\subsubsection{DINOv2-guided graph diffusion}

Fig.~\ref{fig:appendix-ablation-relevancy} shows 2D segmentation masks colored by CLIP relevancy scores, obtained with and without leveraging SAM and/or DINOv2-guided graph diffusion for refining 3D relevancy scores. 

\myparagraph{Direct segmentation from raw 3D relevancy scores} Isolating a specific object in the scene directly based on CLIP relevancy scores is challenging: the segmentation masks obtained by automatic thresholding include parts of other objects with similar features, like for the \emph{sheep} (segmentation of the bear nose) and the \emph{spoon}. The segmentation might also cover surroundings of the object of interest simply due to the low resolution of CLIP visual features, such as in the \emph{knife} example. 

\myparagraph{2D segmentation with SAM} SAM delivers a precise 2D segmentation of the object covered by points with the highest relevancy scores. However, point prompts may not span the entire object, resulting in undersegmentation, like for the \emph{sheep}. In some cases, point prompts with the highest relevancy may even be located on the wrong object, resulting in an entirely wrong segmentation (\emph{e.g.}, the bowl segmented instead of the \emph{spoon}).

\myparagraph{Relevancy score refinement with graph diffusion based on 3D DINOv2 features} The graph diffusion process starts with positive weights for Gaussians with the highest relevancy scores, and propagates their weight to neighbors with similar DINOv2 features. However in cases where the object of interest consists of multiple subparts (e.g. for the \emph{sheep}), the final distribution of weights may be inhomogeneous and the automatic thresholding may select only one subpart. Also, if multiple close objects are to be segmented (e.g. with the \emph{knife}), the final weights may cover surrounding Gaussians and the final thresholding might not clearly isolate the objects.

\myparagraph{3D graph diffusion with 2D SAM segmentation}
Combining 3D graph diffusion and 2D SAM segmentation helps solving the aforementioned problems observed when using either of the two approaches individually. The diffusion process allows selecting a large set of point prompts for SAM spanning the object of interest without covering other object with similar features, resulting in an accurate segmentation.

\subsubsection{Comparisons for open-vocabulary localization} Fig.~\ref{fig:appendix-clip-heatmaps} illustrates open-vocabulary object localization with LERF~\citep{kerr2023lerf}, LangSplat~\citep{qin2023langsplat} and LUDVIG. Both LangSplat and LUDVIG correctly localize all four example objects. For queries such as the chopsticks, LangSplat's localization is more precise, as the CLIP features are constructed by generating full image segmentation masks with SAM. This process is computationally expensive, as constructing a full segmentation mask requires querying SAM over a grid of points on the image and takes about 23s for a single image (on a GPU RTX 6000 ADA), which amounts to an average of 80 minutes for a scene from the LERF dataset. However, it yields coherent instance-level CLIP representations, which is desirable for downstream segmentation tasks.

\newcommand{\plotclip}[2]{%
  \begin{tikzpicture}

    \begin{scope}[on background layer]
      \fill[blue!40, rounded corners, opacity=.5] (-1.5*#2, -1.4*#2) rectangle (1.5*#2, -.5*#2);
      \node[anchor=north east, blue!80] at (1.1*#2, -0.7*#2) {\scriptsize Graph diffusion};

      \fill[green!40, rounded corners, opacity=.5] (0.5*#2, -1.4*#2) rectangle (1.5*#2, .4*#2);
      \node[anchor=north west, green!80] at (0.8*#2, 0.2*#2) {\scriptsize SAM};
    \end{scope}
    
    \node[anchor=center] at (-2.03*#2, 0) {\includegraphics[width=3cm]{#1_rgb.jpg}};
    \node[below=12mm] at (-2.03*#2, 0) {\scriptsize RGB image};

    \node[anchor=center] at (-2.03*#2, -.9*#2) {\includegraphics[width=3cm]{#1_gt.jpg}};
    \node[below=12mm] at (-2.03*#2, -.9*#2) {\scriptsize Ground truth};

    \node[anchor=center] at (0, 0) {\includegraphics[width=3cm]{#1_clip.jpg}};
    \node[below=12mm] at (0, 0) {\scriptsize CLIP features};

    \node[anchor=center] at (0, -.9*#2) {\includegraphics[width=3cm]{#1_dino.jpg}};
    \node[below=12mm] at (0, -.9*#2) {\scriptsize DINOv2 features};

    \node[anchor=center] at (-#2, 0) {\includegraphics[width=3cm]{#1.jpg}};
    \node[below=12mm] at (-#2, 0) {\scriptsize No diffusion, no SAM};

    \node[anchor=center] at (-#2, -.9*#2) {\includegraphics[width=3cm]{#1_dif.jpg}};
    \node[below=12mm] at (-#2, -.9*#2) {\scriptsize Diffusion, no SAM};

    \node[anchor=center] at (#2, 0) {\includegraphics[width=3cm]{#1_sam.jpg}};
    \node[below=12mm] at (#2, 0) {\scriptsize No diffusion, SAM};

    \node[anchor=center] at (#2, -.9*#2) {\includegraphics[width=3cm]{#1_dif_sam.jpg}};
    \node[below=12mm] at (#2, -.9*#2) {\scriptsize Diffusion, SAM};

    \draw[->,line width=.7mm] (-.45*#2, 0) -- (-.55*#2, 0);
    \draw[->,line width=.7mm] (-.45*#2, -.35*#2) -- (-.55*#2, -.55*#2);
    \draw[->,line width=.7mm] (.45*#2, 0) -- (.55*#2, 0);
    \draw[->,line width=.7mm] (.45*#2, -.35*#2) -- (.55*#2, -.55*#2);
    \draw[->,line width=.7mm] (-.45*#2, -.9*#2) -- (-.55*#2, -.9*#2);
    \draw[->,line width=.7mm] (.45*#2, -.9*#2) -- (.55*#2, -.9*#2);

  \end{tikzpicture}%
}
\begin{figure*}
    \centering
    \plotclip{figures/ablation_relevancy/spoon}{3.5} \\ \vspace{.5cm}
    \plotclip{figures/ablation_relevancy/knife}{3.5} \\ \vspace{.5cm}
    \plotclip{figures/ablation_relevancy/sheep}{3.5}
    \caption{\textbf{Open-vocabulary object segmentation with and without using 3D graph diffusion (blue) and/or 2D SAM segmentation (green).} Projections of 3D CLIP and DINOv2 features colored by three main PCA components and 2D segmentation masks colored by relevancy scores.}
    \label{fig:appendix-ablation-relevancy}
\end{figure*}

\begin{figure*}
    \centering
    \includegraphics[width=\linewidth]{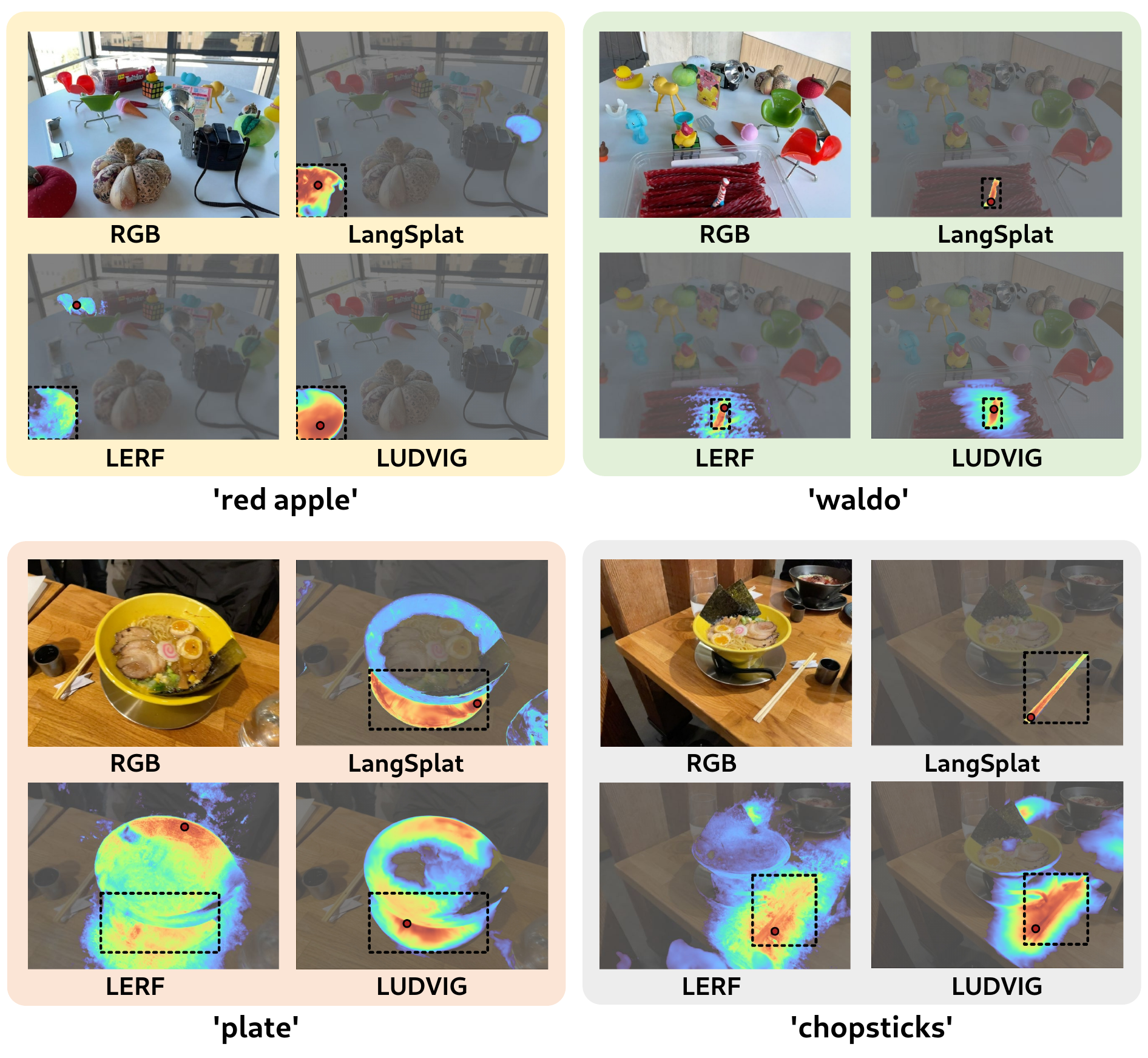}
    \caption{\textbf{Qualitative comparisons of open-vocabulary 3D object localization on the LERF dataset.} The red points are the model predictions and the black dashed bounding boxes denote the annotations. This figure is taken and adapted from LangSplat's website (https://langsplat.github.io/), licensed under a Creative Commons Attribution-ShareAlike 4.0 International License.}
    \label{fig:appendix-clip-heatmaps}
\end{figure*}

\end{document}